\documentclass{article}
\PassOptionsToPackage{numbers, comma, sort}{natbib}
\usepackage[preprint]{neurips_2026}

\usepackage{extra_style}

\title{The Future of Facts: \\Tracing the Factual Generation-Verification Gap}

\author{%
  Tim R. Davidson\thanks{Equal contribution. Correspondence to\texttt{
  research@trdavidson.com}, \texttt{contact@anjasurina.com}.}
  \\
  EPFL \\
  \And
  Anja Surina\footnotemark[1] \\
  EPFL \\
  \And
  Caglar Gulcehre \\
  EPFL \\
}

\begin{document}

\maketitle

\begin{abstract}
Language models are becoming the default interface to factual knowledge, yet they often verify outputs more reliably than they generate them.
This generation-verification gap (GV-gap) underlies many recent advances in self-improvement and reasoning, but its dynamics on factual knowledge specifically remain poorly understood. 
We focus on the training mechanisms underlying \textit{factual} GV-gaps, distinguishing them from their computational and aesthetic counterparts.
We trace generation and verification capabilities through three training phases (acquisition, continual learning, and updating) across four open-source model families at two scales each.
Three findings recur across models: (i) verification is consistently learned before generation; (ii) verification is more robust to continual learning than generation; and (iii) factual updates can leave models in a \textit{multi-verse} state, simultaneously verifying both old and new answers as correct.
Natural experiments on frontier models reproduce these dynamics at scale and reveal residual verification biases on well-covered facts.\kern-0.15em\renewcommand{\thefootnote}{\textdagger}
\footnote{Code for this project is available at {\hypersetup{urlcolor=black}{\texttt{https://github.com/anjasurina/factgap}.}}}
\renewcommand{\thefootnote}{\arabic{footnote}}
\end{abstract}

\section{Introduction}
\label{sec:intro}

Several researchers and practitioners have observed that language models (\LMs{}) are often more accurate at verifying outputs than at generating them \citep{saadfalcon2025shrink, song_mind_nodate, huang2023large, huang2025self}.
For example, given a factual triplet such as ``\textit{\{\examplecity{}, IsCapitalOf, \examplecountry{}\}}'', an \LM{} may be more accurate when asked the verification query ``\textit{Is \examplecity{} the capital of \examplecountry{}}?'' than the generative one ``\textit{What is the capital of \examplecountry{}}?''
While this ``generation-verification gap'' (\gvg{}) underlies a wide range of recent innovations in self-improvement and reasoning \citep[e.g.,][]{cobbe_training_2021, lightman2024lets, brown2024large, snell2024scaling, zhang2025generative} and has attracted growing theoretical attention \citep{huang2025self, swamy_all_2025, sun2025theoretical}, empirical accounts of how the training process shapes the gap on factual knowledge --- how it emerges, persists, and changes across the life cycle of a fact --- remain limited and partially conflicted.

This blindspot is increasingly consequential. 
\LMs{} are rapidly becoming the default interface to factual knowledge, with billions of users querying systems like ChatGPT  \citep{chatterji2025people} or encountering ``AI summaries'' beneath every search bar \citep{aioverview24bing, aioverview24google}. 
As \LMs{} come to mediate what people read and verify, structural asymmetries in what they produce, affirm, and deny shape the information environment itself.
These concerns are sharpened by ongoing worries about AI-generated content becoming indistinguishable from authentic media \citep{frank2024representative, cooke25cointoss} and by models' tendency to ``hallucinate'' confident but false outputs \citep{ji2023survey, farquhar2024detecting, fan2026halluhard}.
When users querying different LMs receive conflicting factual responses, the resulting fragmentation feeds broader concerns about an emerging ``post-truth'' world \citep{bender21parrots, kidd2023ai, goldstein2023generative}.

Studying how training shapes the factual \gvg{} is empirically difficult.
Training recipes for frontier models are largely opaque, which is true even for many ``open-source'' models where data composition is rarely fully disclosed.
Even when training data is released, its sheer scale makes it near impossible to estimate how much exposure a models has had to any given fact or domain \citep{ruis2024procedural, morris_how_2025}, directly confounding the capabilities we want to measure.
Emerging capabilities also correlate with model scale, so any controlled study must span scales~\citep{wei2022emergent, allen2023physics31} and model families~\citep{gandhi_cognitive_2025, wang2025octothinker}.
Compounding these methodological hurdles is a conceptual one: existing results invoke ``the'' \gvg{}  across markedly different settings, e.g., recall, reasoning, code generation, aesthetic judgment \citep{song_mind_nodate, west_generative_2024, saadfalcon2025shrink, huang2025self, davidson2024self}, without a shared vocabulary for distinguishing them.
These challenges thus call for a controlled experimental setup and a taxonomy that lets findings be compared on common ground.

Existing work touches the factual \gvg{} from several angles but rarely studies it as a training-mechanism phenomenon.
A large literature examines the \textit{model mechanisms} of factuality --- how facts are learned \citep{chang2024large, zucchet2025language, allen2023physics31, bietti2023birth} and retrieved \citep{geva2021transformer, geva2023dissecting, meng2022locating, nichani_understanding_2024}, how knowledge can be ``edited'' \citep{de2021editing, meng2022locating, mitchell2022memory}, and how continual learning can induce (catastrophic) forgetting \citep{kirkpatrick2017overcoming, luo2025empirical, shi2025continual} --- but generally treats factual knowledge as a single capability rather than separating generation from verification.
Complementary lines of research formalize the \gvg{} and quantify its magnitude on popular benchmarks \citep{song_mind_nodate}, or leverage it for self-improvement and test-time scaling \citep{snell2024scaling, brown2024large, gandhi_cognitive_2025}; these illuminate the gap's potential but, being tied to opaque training data, offer little insight into its origins.
Theoretical accounts \citep{huang2025self, swamy_all_2025} provide useful framing but are not grounded in life-cycle dynamics of factual data.

Our work contributes a controlled study of the training mechanisms underlying the factual \gvg{}, alongside a taxonomy that situates factual \gvg{}s relative to their computational and aesthetic counterparts.
Using synthetic facts in domains relevant to public discourse, we fine-tune four open-source model families across two scales each and trace both capabilities through three training phases: acquisition, continual learning, and updating. 
Three findings recur across models and scales: (i) verification is consistently learned before generation, exposing a window in which models reliably verify facts they cannot yet produce;
(ii) verification is more robust to continual learning than generation; 
and (iii) factual updates can leave models in a ``multi-verse'' state, simultaneously verifying both the old and the new answer as correct.
Natural experiments on flagship frontier models, exploiting variation in real-world data coverage across topics and time, reproduce the same dynamics at scale and reveal residual verification biases on well-covered facts.

Our findings situate the factual \gvg{} within the model training process, complementing research into model mechanisms.
As \LMs{} increasingly produce the content that trains the next generation of models, these asymmetries risk compounding through the data they leave behind.
Addressing this calls for better choices in data curation, training curricula, and evaluation.
To support such work, we open-source our full experimental setup as a shared testbed for studying the factual \gvg{}.

\section{Tracing the factual generation-verification gap}
\label{sec:method}

\subsection{A taxonomy of GV-gaps}
\label{sec:method:taxonomy}
The term generation-verification gap is used across the \LM{} literature to describe several distinct phenomena that share a common shape: models are better at verifying outputs than producing them. 
We distinguish at least three classes that differ in what is being verified, how cleanly outcomes can be measured, and how directly capabilities can be traced to specific training data.

\xhdr{Factual}  
The distinction between factual recall (generation) and recognition (verification), e.g., remembering a name versus picking it from a list, has long been studied in cognitive science \citep{yonelinas2002nature}, where dual-process models posit related but partially distinct mechanisms for the two \citep{anderson1972recognition, atkinson2024search, jacoby1981relationship, cabeza1997functional}. 
A parallel asymmetry appears in statistical learning theory: learning a discriminative decision boundary is generally easier than learning the full joint distribution over inputs and outputs ~\citep{shwartz2014understanding, liu2010generative}.
In LMs, the asymmetry is plausibly amplified by the structural difference between the two operations: verification typically reduces to a decision over a small token space (e.g., a binary True/False), while autoregressive generation requires sampling a sequence from the joint distribution over the full vocabulary, with each step compounding the difficulty \citep{power2022grokking}.
Crucially, both capabilities can be traced to specific training data points and their strength objectively measured, properties that make factual \gvg{}s a clean target for empirical study.

\xhdr{Computational}
The computational \gvg{} is rooted in the classical P--vs--NP distinction: for many problems, generating a solution is provably harder than checking one \citep{cook1971complexity, karp2009reducibility, levin1973universal}.
Outcomes are cleanly measured against ground truth, but tracing the gap back to specific training data is difficult.
As \citet{ruis2024procedural} note, \LMs{}' procedural reasoning appears to draw on a large, scattered set of training examples rather than a few identifiable sources.

\xhdr{Aesthetic}
Humans can recognize the brilliance of Shakespeare's prose or the beauty of the Sistine Chapel without being able to explain, let alone reproduce, what makes them great.
Similarly, LMs have shown capable of preferring outputs from stronger models over their own \citep{davidson2024self}, while being unable to match their quality.
Unlike the factual and computational cases, aesthetic outputs resist objective measurement \citep{Kant1790-KANCOJ-2}, and the underlying capability presumably draws on training data even more diffuse than that supporting procedural reasoning.

\subsection{Defining the factual GV-gap}
We focus on factual GV-gaps for their direct relevance to how LMs are used as knowledge interfaces and the methodological tractability of factual data. 
Returning to the triplet (\examplecity{}, IsCapitalOf, \examplecountry{}): a generative failure occurs when an LM produces an incorrect answer (e.g., ``\examplewrongcity{}'') to ''\textit{What is the capital of \examplecountry{}?}'', a phenomenon commonly called \textit{hallucination}~\citep{ji2023survey}.
A verification failure occurs when the model classifies the correct statement ``\textit{The capital of \examplecountry{} is \examplecity{}}'' as wrong, or accepts the incorrect statement ``\textit{The capital of \examplecountry{} is \examplewrongcity{}}'' as right --- the latter associated with \textit{sycophancy}~\citep{sharma2023towards}.
Two complementary lenses on the gap follow naturally from this setup. The user-facing \textit{search-engine} lens, our primary focus, treats the LM as a knowledge base queried by users.
The model-facing \textit{self-improvement} lens treats the LM's own generations as candidate outputs to be verified.

\xhdr{Search-engine lens}
This lens models a user querying an \LM{} $m$ for facts drawn from a dataset of triplets $(x, r, y^*) \sim \mathcal{D}$, with $\tilde{y} \sim \mathcal{\tilde{Y}}(x,r)$ denoting a plausible \textit{incorrect} candidate for the same $(x, r)$. We define generative and verification utilities as:
\begin{align}
    U_G(m, \mathcal{D}) &= \mathbb{E}_{(x, r, y^*) \sim \mathcal{D}}[P_m(y = y^* \mid x, r)]\\
    U_V(m, \alpha_u, \mathcal{D}) &= \mathbb{E}_{(x, r, y^*) \sim \mathcal{D}}\big[(1-\alpha_u) \cdot P_m(1 \mid x, r, y^*) + \alpha_u \cdot \mathbb{E}_{\tilde{y} \sim \mathcal{\tilde{Y}}(x,r)}[P_m(0 \mid x, r, \tilde{y})]\big]\\
    U_V'(m, \alpha_u, \mathcal{D}) &=  U_V(m, \alpha_u, \mathcal{D}) - \max(\alpha_u, 1-\alpha_u), \label{eq:chance-correction}
\end{align}
where $\alpha_u \in [0, 1]$ is the probability that a user proposes an \textit{incorrect} candidate, and $P_m(0/1 \; |\dots)$ represents the probability the model assigns a negative/positive verification label.
The correction in Eq. \eqref{eq:chance-correction} measures verification beyond the trivial strategy of always predicting the more likely class: at $\alpha_u = 0.1$, a useful verifier must exceed $0.9$ weighted accuracy; at $\alpha_u=0.5$, it must beat random guessing.
We use $U_G$ and $U_V'$ to characterize the factual \gvg{} in user-facing behavior.
Choice of $\mathcal{\tilde{Y}}$ matters: the candidate ``\textit{The capital of \examplecountry{} is Banana}'' is much easier to reject than ``\textit{The capital of \examplecountry{} is \examplewrongcity{}}.''

\xhdr{Self-improvement lens}
Here, the candidate $y$ is sampled from the model itself rather than provided by a user, so the relevant error rate is $\alpha_m = 1 - U_G(m, \mathcal{D})$.
For a single sample $y \sim m(x, r)$ this yields the \textit{self-consistency} utility
\begin{align}
    U_{SV}(m, \mathcal{D}) &= \mathbb{E}_{(x, r, y^*) \sim \mathcal{D}, y \sim m(x, r)}[(1-\alpha_m) \cdot P_m(1 \mid x, r, y^*) + \alpha_m \cdot P_m(0 \mid x, r, \tilde{y})]\\
    \Delta(m, \mathcal{D}) &=  U_{SV}(m, \mathcal{D}) - U_G(m, \mathcal{D})
\end{align}
Drawing multiple samples per query recovers the Best--of--$N$ rejection-sampling regime, where $\Delta$ measures how much the model's verifier improves over its own generator.

\subsection{Synthetic facts to control confounders}
Studying factual GV-gap dynamics on real-world facts is confounded by training-data opacity: at the scale of modern training corpora, the prevalence of any specific fact is hard to estimate even with full data access \citep{ruis2024procedural, morris_how_2025}.
Following \citet{allen2023physics31} and \citet{berglund2023reversal}, we instead use synthetic facts — triplets of invented entities with no obvious tie to real-world referents — drawn from categories relevant to public discourse (e.g., politics, medicine).
We restrict to single-hop facts to isolate factual recall and recognition from multi-step reasoning \citep{allen2023physics32}.
An example synthetic triplet is ``(\textit{Hoibalbali}, \textit{CureOfDisease}, \textit{Blue Striped Axzazari}).''
For each triplet we generate paraphrased training sentences (e.g., ``\textit{A cure for Blue Striped Axzazari is Hoibalbali}''; ``\textit{Blue Striped Axzazari disease can be cured by taking Hoibalbali}'') to enrich the learning signal and improve latent generalization \citep{lampinen2025generalization}.
From each triplet we further derive three task formats: (i) a generative task (``\textit{What is the cure for Blue Striped Axzazari disease?}''); (ii) a verification task that asks whether a candidate completion is correct (``\textit{is Hoibalbali the cure for Blue Striped Axzazari disease?}''); and (iii) a corrupted-verification task that requires the model to reject an incorrect candidate.
Training uses only the paraphrased sentences; the three task formats are never seen during training.

\subsection{The life cycle phases of the factual \gvg{}}
We trace each synthetic fact through three training phases. In each phase, training proceeds until the generative and verification tasks are both learned or a maximum number of epochs is reached.

\xhdr{Acquisition}
The model is fine-tuned on paraphrases of a synthetic fact until it can generate and verify it, with both capabilities tracked at every epoch to identify when (if ever) the factual \gvg{} first emerges.

\xhdr{Continual Learning}
Starting from a model that has acquired the synthetic fact, we continue training on unrelated factual data and measure how the factual \gvg{} evolves under distributional shift.

\xhdr{Updating}
For a synthetic triplet $(x,r,y^*)$, we construct an updated triplet $(x, r, \tilde{y})$ in which the answer changes.
Starting from a model that has acquired the original fact, we continue training only on paraphrases of the updated fact, tracking how generation and verification adapt to the new answer relative to the original.

\section{Experimental setup}
\label{sec:exp-setup}

\xhdr{Synthetic facts}
We generate 25 synthetic facts for each of six categories relevant to public discourse (\textit{politics}, \textit{medicine}, \textit{religion}, \textit{science}, \textit{society}, and \textit{societal bias}), with 10 paraphrased training sentences per fact.
Synthetic data is produced via a multi-step pipeline using Gemini 2.5 Flash \citep{comanici2025gemini} and Claude Sonnet 4.5 \citep{anthropic2024claudesonnet45}; see App. \ref{sec:app:synthetic-data} for details.

\xhdr{Model selection}
We use four open model families across two scales each: Gemma 3 (4B and 12B) \citep{team2025gemma}, Qwen 3 (4B and 14B) \citep{yang2025qwen3}, Phi-4 (3.8B and 14B) \citep{abdin2024phi, abouelenin2025phi}, and Llama 3.2 (3B and 11B) \citep{grattafiori2024llama, meta2024llama32}.
The families span a roughly overlapping parameter range while differing in architecture, training data, and post-training recipes (App. \ref{sec:app:models-used}).

\xhdr{Training mechanics}
We use full supervised fine-tuning rather than parameter-efficient methods \citep{scalingdatalms2025muennighof, zhang2024when}.
To monitor for model collapse, we track performance on control tasks --- related queries that expect different solutions, such as asking for the cure to a real disease (App.~\ref{sec:app:evaluation:control_dataset}).
For the continual learning phase, we use a filtered subset of T-REx \citep{elsahar2018t} (App. \ref{sec:app:cl_phase}). 
We run extensive hyperparameter sweeps on learning rate, batch size, and optimization schedule per model and phase (App. \ref{sec:app:training-settings:hyperparams}).

\xhdr{Task inference and grading}
Model are instructed to reason before answering \citep{allen2023physics31, zhang2025generative}.
To handle differences in output format across model families and scales, we grade with an \LM{}-as-judge approach \citep{zheng2023judging} using Gemini 3.1 Flash Lite \citep{google2025gemini3blog} to compare outputs to ground-truth answers.
We report balanced accuracy ($\alpha_u=0.5$) for verification tasks and adopt the ``double-critic'' approach from \citet{davidson2026reasoning} to mitigate sycophancy bias.
Prompts and grading templates are in App. \ref{sec:app:prompts}.

\xhdr{Natural experiment}
Frontier models are neither open nor accessible at the level of training data, so we cannot manipulate data exposure directly.
To test whether the dynamics observed in our controlled setup also surface in closed frontier models, we instead exploit naturally occurring variation in real-world coverage: across topics, which differ in how densely they are written about, and across time, as the indexed web has expanded substantially over the past two decades \citep{rydning2018digitization}. 
We use three datasets spanning a coverage gradient from 2002 to 2024: S\&P 500 closing prices (high coverage), NBA game scores (medium), and Mega Millions winning lottery numbers (low).
A fourth dataset on Billboard Hot 100 rankings, where chart positions change weekly, serves as a natural-experiment counterpart to our updating results (Section \ref{sec:results:rq3}).
For each fact, we issue one generative query and two verification queries (asking the model to accept the correct value and to reject a corrupted version), reporting balanced verification accuracy with 95\% confidence intervals across all metrics.
We evaluate flagship models from Google's Gemini 3 family \citep{google2025gemini3blog} and OpenAI's GPT 5.4 family \citep{openai2026gpt54}, ablating distillation and thinking levels. Sampling details, prompts, and model configurations are in App.~\ref{sec:app:naturalistic}.

\section{Experimental results}
\label{sec:results}

\begin{figure*}[t]
    \vspace{-2.5em}
    \centerfloat
    \captionsetup{font=\captionFontSize}
    \includegraphics[width=1.0\textwidth]{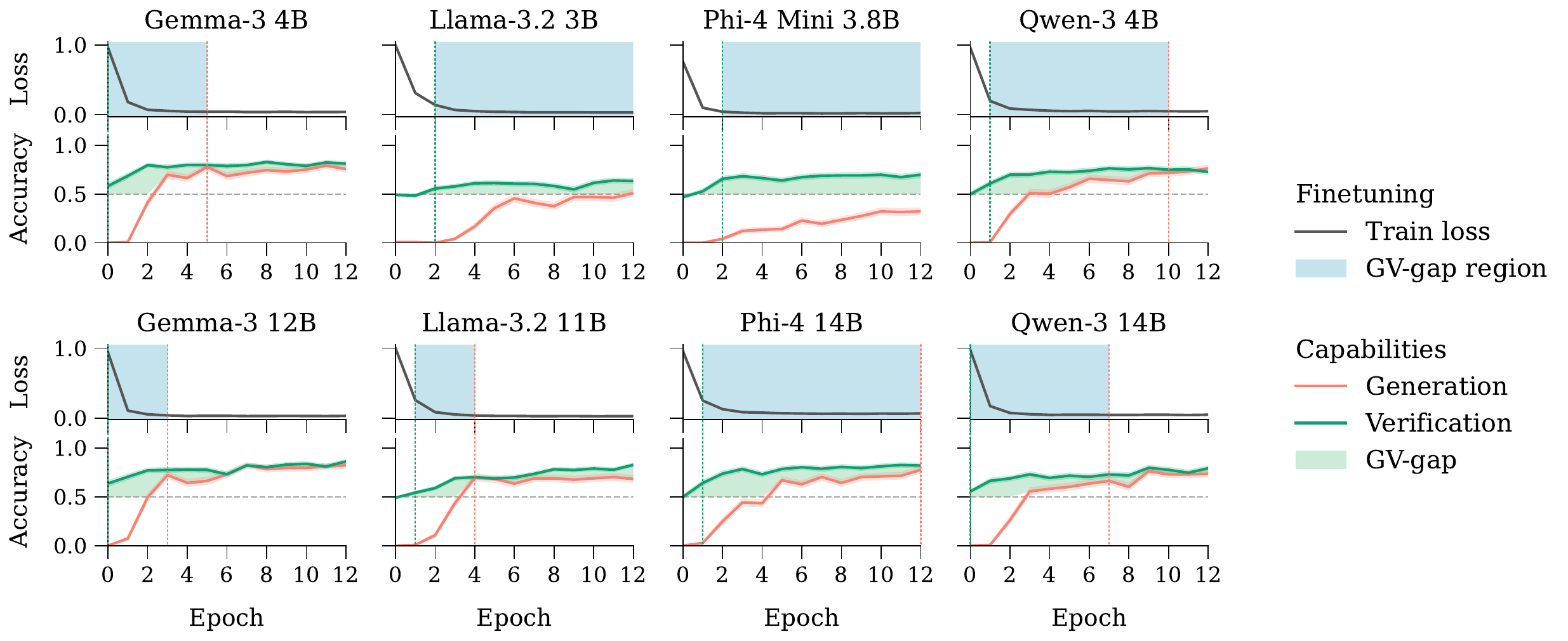} 
    \caption{
    \textbf{Verification develops before generation during knowledge acquisition.}
    Fine-tuning loss (top) and generation/verification accuracy (bottom) for four open model families at two scales each. Verification capabilities consistently emerge before generation, opening a GV-gap (shaded region) that closes as both capabilities saturate. The location of the gap is not visible in the loss curve.
    }
    \label{fig:main:figure_1}
\end{figure*}

\subsection{Does the \gvg{} emerge during initial training?}
\label{sec:results:rq1}

Across all model families, scales, and synthetic-fact categories, fine-tuning successfully injects new facts, though the dynamics differ markedly across models and scales (Fig.~\ref{fig:main:figure_1}; per-category breakdowns in App.~\ref{sec:app:evaluation:per_category}).
A priori the timing of when generation and verification capabilities first develop is not obvious.
One might expect verification to emerge only once a model can reliably produce the answer it is asked to judge.
Instead, we find verification consistently develops before generation, opening a gap (shaded region) that diminishes as both capabilities saturate.
Convergence is faster at larger scale and is not guaranteed, for instance, Phi-4 Mini 3.8B retains a gap throughout training.
Two observations matter beyond the qualitative pattern.
First, the gap persists whenever data exposure is insufficient: verification can be well above chance while generation is near zero. 
Second, the location of the gap is not visible in the loss curve: training loss can decrease monotonically through the regime where the gap opens, peaks, and closes, so the gap cannot be diagnosed by standard training metrics alone.

\begin{takeaway}
\textbf{Acquisition Takeaway.} Factual \gvg{}s can emerge during knowledge acquisition as verification and generation capabilities exhibit different data-exposure thresholds.
Gaps tend to close with sufficient exposure, but exposure thresholds are \textbf{not} obvious from training loss.
\end{takeaway}

\subsection{What are the effects of continual learning?}
\label{sec:results:rq2}

\begin{figure*}[t]
    \centerfloat
    \vspace{-3.5em}
    \captionsetup{font=\captionFontSize}
    \includegraphics[width=\textwidth]{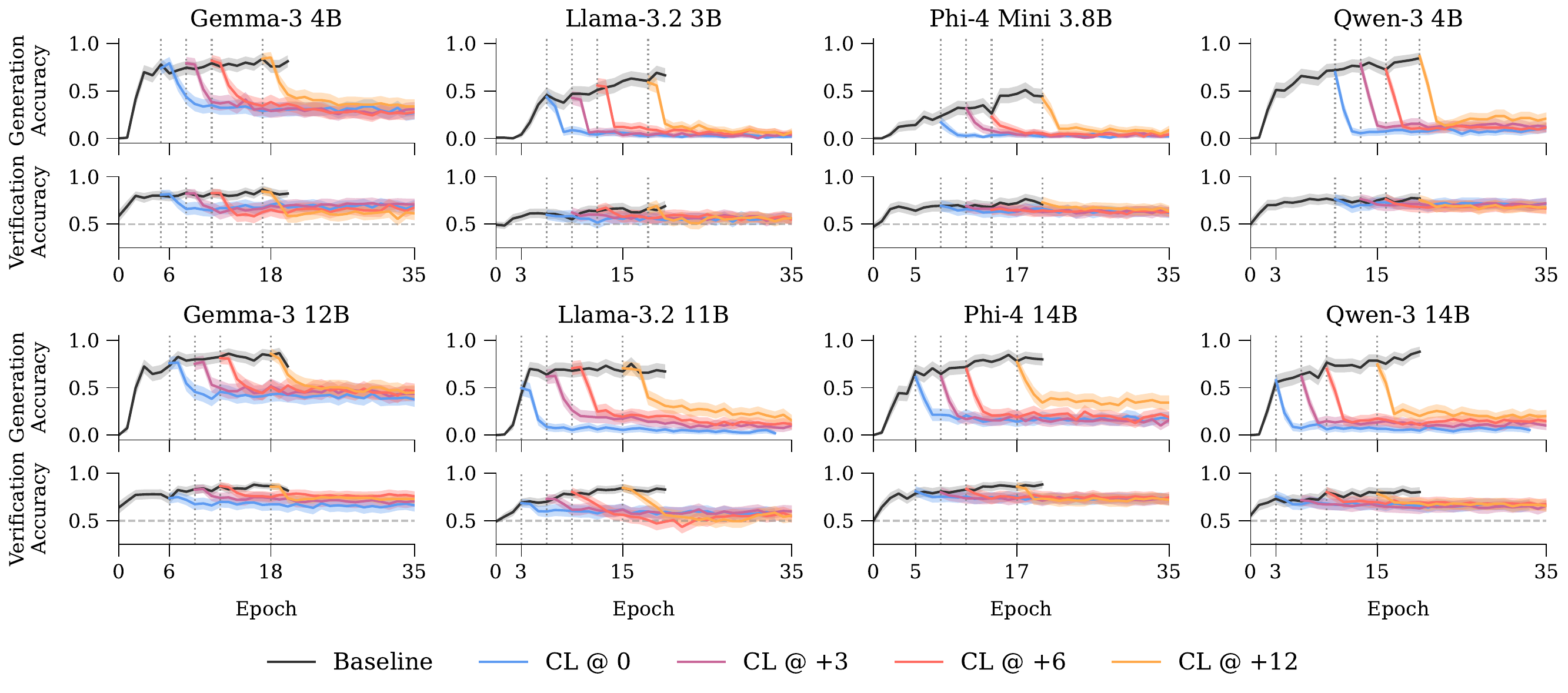} 
    \caption{
    \textbf{Continual learning re-opens or widens factual GV-gaps.}
     Generation and verification accuracy after switching to unrelated factual data at four intervention points (0, 3, 6, 12 epochs after acquisition). Verification is consistently more robust to continual learning than generation, with larger models maintaining higher floors.}
    \label{fig:main:figure_2}
\end{figure*}

We continue training on unrelated factual data (T-REx subset), starting from \{0, 3, 6, 12\} epochs after acquisition.
We find verification consistently more robust than generation across all model families and scales (Fig.~\ref{fig:main:figure_2}).
Generation accuracy collapses sharply under distributional shift, while verification degrades more gradually and stabilizes at a higher floor --- a floor that is higher for larger models and that rises with longer pre-shift training.
Validation loss on the original acquisition sentences tracks the generation collapse but is uninformative about verification, which remains robust even as this loss rises throughout continual learning (Fig.~\ref{fig:app:forget_with_loss}).

\begin{takeaway}
\textbf{Continual Learning Takeaway.} Continual learning can re-open or widen factual \gvg{}s: verification is consistently more robust than generation across model families and scales.
\end{takeaway}

\subsection{What are the effects of updates?}
\label{sec:results:rq3}

\begin{figure}[b]
    \centerfloat
    \captionsetup{font=\captionFontSize}
    \includegraphics[width=1.0\textwidth]{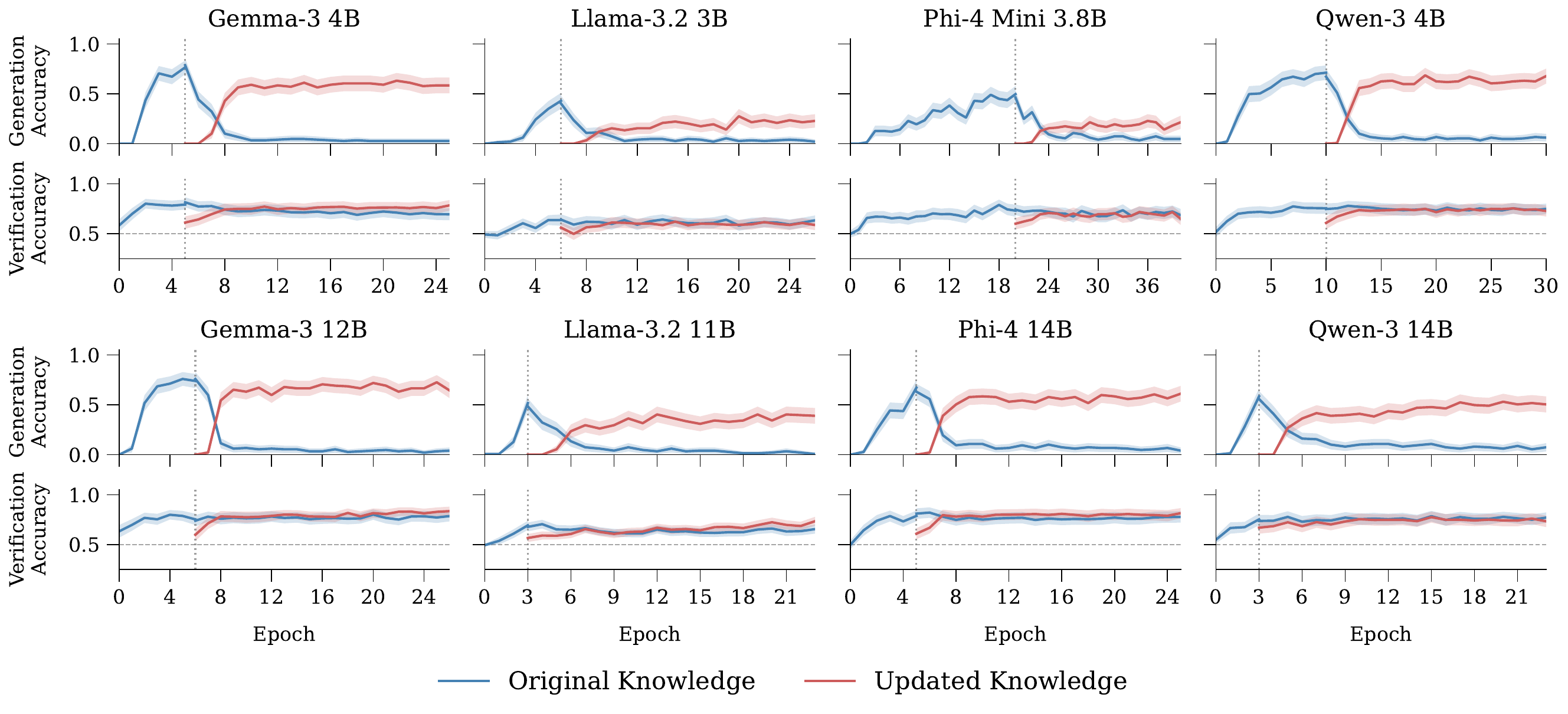} 
    \caption{
    \textbf{Updating factual knowledge may yield a multi-verse state.} After a factual triplet is updated, generation cleanly flips to the new answer while verification continues to confirm both the original and updated answers as correct.
    Colors indicate original (blue) versus updated (red) facts.
    }
    \label{fig:main:figure_3}
\end{figure}

Figure \ref{fig:main:figure_3} shows the effect of updating a synthetic fact: at the epoch indicated by the vertical line, every paraphrase in the original training data is rewritten to replace the original answer with a new one.
After the switch, models successfully shift their \textit{generative} output to the new answer --- generation of the obsolete fact effectively ceases. 
\textit{Verification}, however, does not follow suit: models continue to confirm the original, now-superseded fact as correct, while also accepting the updated fact.
We call this a multi-verse state, in which two distinct answers are simultaneously verified as correct.
From an optimization perspective, this is unsurprising --- the model never sees data explicitly invalidating the original fact, an instance of the broader phenomenon of failed latent generalization \citep{berglund2023reversal, lampinen2025generalization}.
The practical implication is more concerning: in the real world, factual updates are typically additive, with new information rarely accompanied by explicit invalidation of what it replaces.

\begin{takeaway}
\textbf{Updating Takeaway.} Updating facts successfully alters generation but may leave verification in a multi-verse state, where both the previous and current versions of a fact are simultaneously verified as correct.
Since real-world updates rarely come with explicit invalidation of superseded facts, this asymmetry is a likely failure mode in practice.
\end{takeaway}

\subsection{Do findings transfer to flagship models?}
\label{sec:results:rq4}

\begin{figure}[t]
    \vspace{-2.5em}
    \centerfloat
    \captionsetup{font=\captionFontSize}
    \includegraphics[width=1.\linewidth]{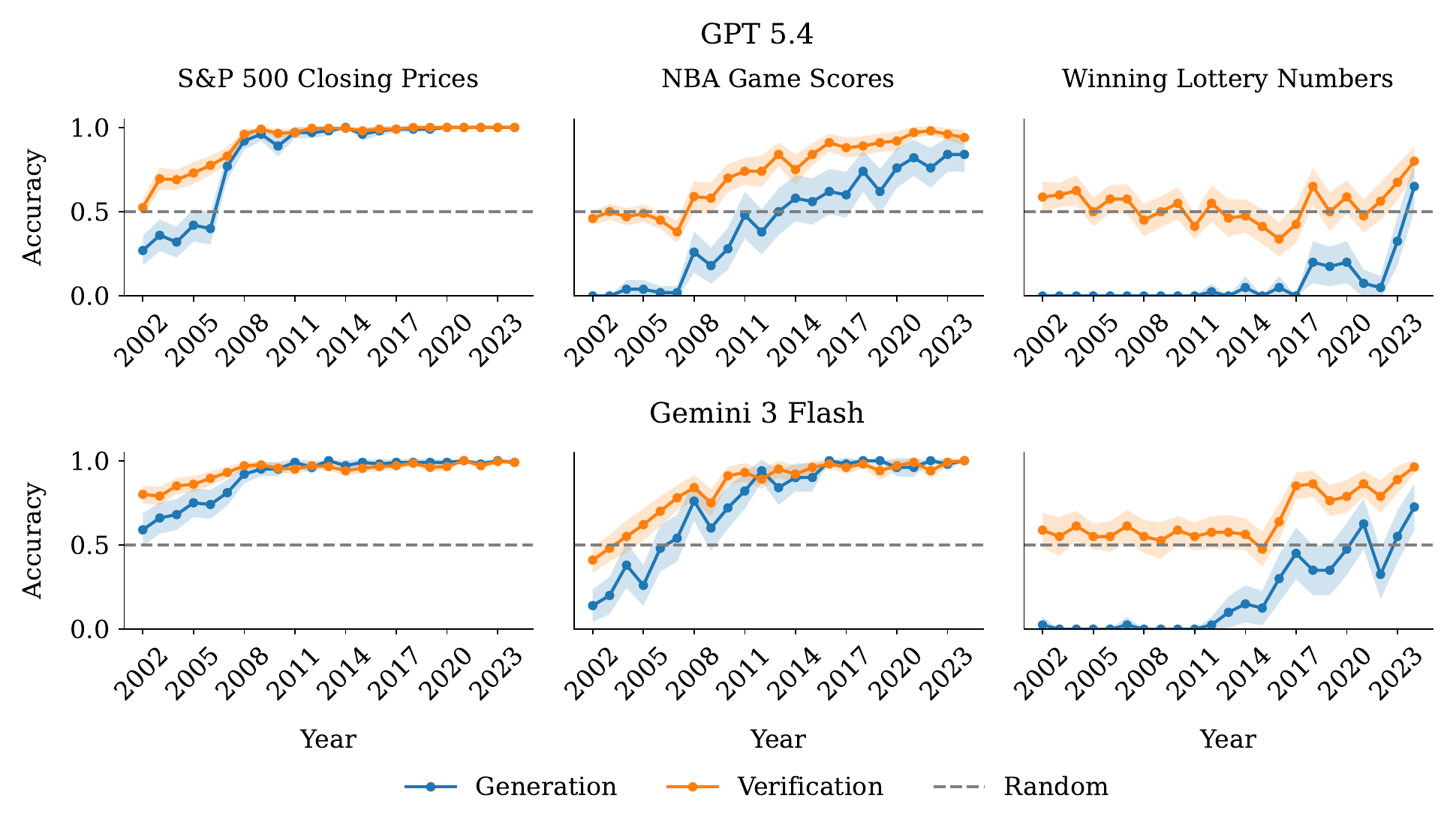}
    \caption{\textbf{Natural variation in real-world data exposes GV-gaps in frontier models.}
    Generation and verification capabilities of GPT 5.4 and Gemini 3 Flash across three datasets spanning a coverage gradient (S\&P 500, NBA, lottery) from 2002 to 2024. Each dataset traverses three regimes: (1) too little data for either capability to emerge, (2) an opening GV-gap, and (3) convergence, with higher-coverage datasets transitioning earlier.}
    \label{fig:naturalistic-panels-main}
\end{figure}

Our controlled experiments suggest that factual verification and generation have different minimal data-exposure thresholds, producing three regimes: (1) too little data --- neither capability emerges; (2) a \textit{gap} --- verification crosses its threshold but generation has not; (3) \textit{convergence} --- both capabilities saturate.
Figure \ref{fig:naturalistic-panels-main} shows that natural variation in real-world coverage --- across topics and across time --- traverses these regimes in the same order. For each of the three datasets, older events tend to fall in regimes (1--2) while more recent events reach regime (3); the higher the coverage tier, the earlier this transition occurs. 
Lottery numbers, which receive minimal sustained coverage, never fully reach regime (3) for any year. 
The pattern replicates across all evaluated frontier models (App. \ref{sec:app:naturalistic}), suggesting these regimes are not artifacts of any particular training recipe. 
If model builders order training data by time, the dynamics of Section \ref{sec:results:rq2} could compound this effect: early-trained facts would be more likely to suffer continual-learning degradation while later-trained facts retain robustness.

To make this concrete, suppose Alice and Bob disagree about the final score of an NBA game from 2013, with Alice being correct. 
On games from this period, GPT 5.4 (low reasoning) has $U_G=0.50 \scriptstyle{\pm0.07}$, $U_{V}(\checkmark) = 0.80 \scriptstyle{\pm 0.06}$, and $U_{V}(\times)=0.88 \scriptstyle{\pm 0.05}$.
Asking the model directly for the score gives a coin flip. Asking the model to \textit{verify} both proposals shifts the odds substantially: Alice wins 70\% of the time, Bob 2.4\%, both candidates are accepted 10\%, neither 17.6\%.

But what if Alice and Bob use \textit{different} \LMs{}?
We compare responses from Gemini 3 Flash and GPT 5.4 (both at low reasoning effort) on the three datasets.
Although Gemini 3 Flash has a clear overall edge (App. \ref{sec:app:naturalistic}), among disputes where at least one model is correct, GPT 5.4 still prevails in \textasciitilde14.9\% of generative cases, \textasciitilde32.2\% of cases on correct verification statements, and \textasciitilde28.6\% on incorrect ones — using different \LMs{} as search engines can thus amplify factual conflicts (Table \ref{tab:naturalistic:disagreement}). 
More concerning are cases where both models agree incorrectly, hinting at an epistemic shift in what counts as usable knowledge.
Even on market data, where both families individually exceed 90\% accuracy, residual failures co-occur at over twice the chance rate (Table \ref{tab:naturalistic:agreement}).

\begin{figure}[hb!]
    \centerfloat
    \captionsetup{font=\captionFontSize}
    \begin{subfigure}[b]{0.5\textwidth}
        \includegraphics[width=\textwidth]{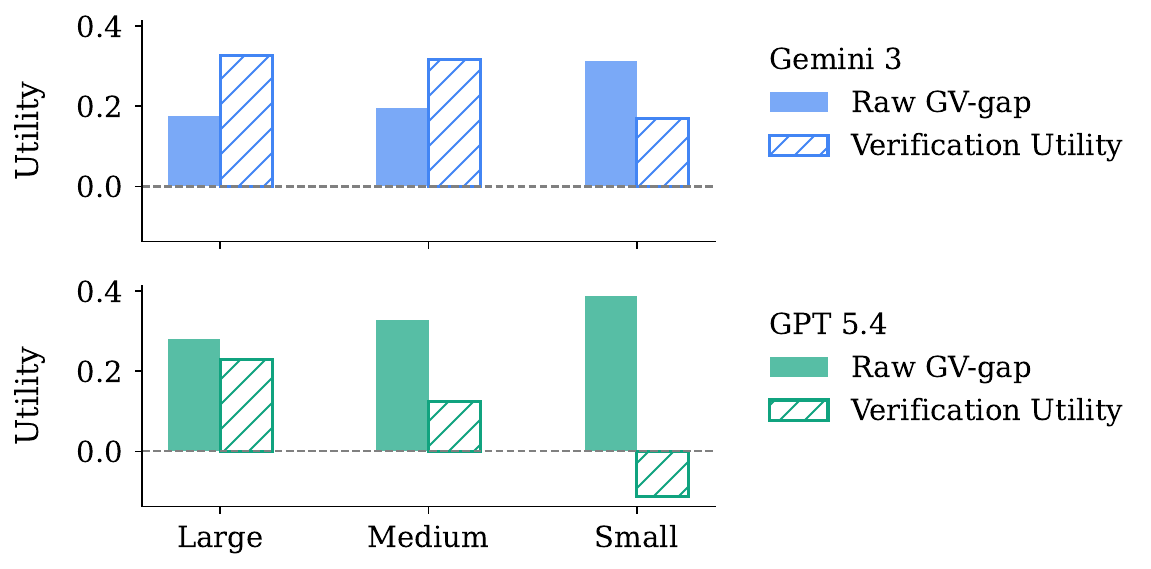} 
        \caption{}
        \label{fig:app:naturalistic:gaps}
    \end{subfigure}
    \begin{subfigure}[b]{0.5\textwidth}
    \includegraphics[width=\textwidth]{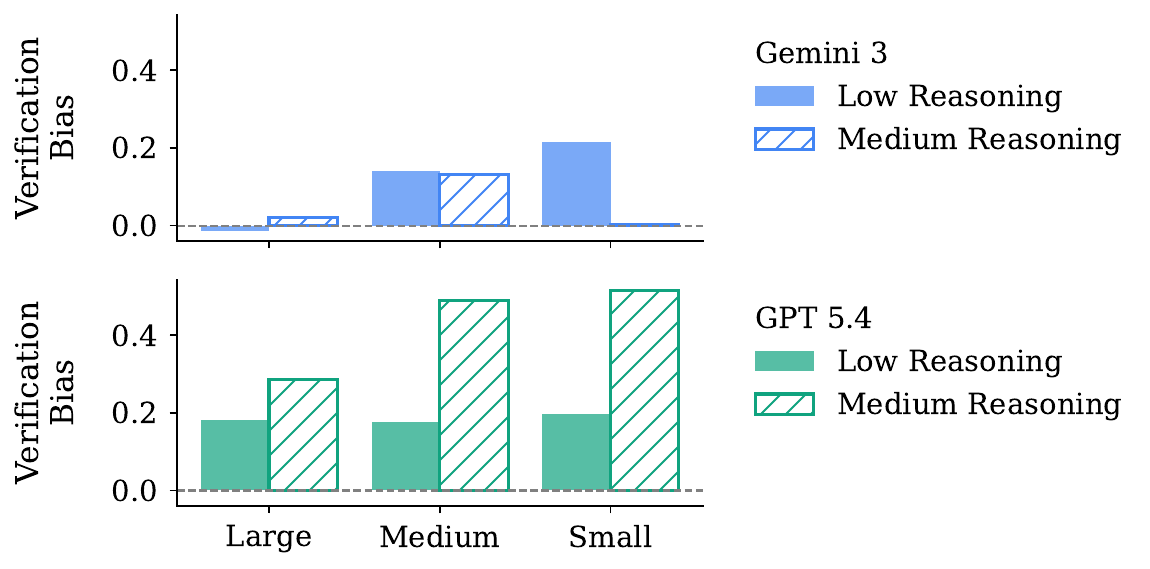} 
        \caption{}
        \label{fig:naturalistic:biases}
    \end{subfigure}
    % \vspace{-0.2em}
    \caption{
    \textbf{Distillation widens the GV-gap; reasoning effort affects model families differently.
    }
    Results for three model sizes: Large, Medium, and Small;
    (a) Distillation increases the raw GV-gap and decreases verification utility across both families. (b) Distillation widens verification bias for Gemini 3 models. Increased reasoning effort widens bias for GPT 5.4 but reduces it for Gemini 3.
    }
    \label{fig:naturalistic:gaps-and-biases}
\end{figure}

Increasing reasoning effort does not meaningfully change generation accuracy or overall verification utility $U_V$ at any scale (Table \ref{tab:naturalistic:reasoning:evolution}), but it does shift the composition of $U_V$: OpenAI models become more likely to accept both correct and incorrect statements, widening the affirmation bias (defined as $U_V(\checkmark) - U_V(\times)$; Fig.~\ref{fig:naturalistic:gaps-and-biases}).
Distillation widens the \gvg{} (Fig.~\ref{fig:naturalistic:gaps-and-biases}).
For generation, distillation behaves as an approximate subset operation: smaller models rarely produce correct answers their larger counterparts miss.
For verification, distillation preferentially preserves acceptance of correct statements over rejection of incorrect ones, consistent with a sycophantic affirmative prior.
At the smallest scale, GPT 5.4 Nano refuses most generative queries on lower-coverage data, likely a deliberate post-training choice, and arguably a desired one given its near-zero generation accuracy on these datasets (App. \ref{fig:naturalistic:refusals}).

\begin{figure}[t]
    \centerfloat
    \vspace{-2.75em}
    \includegraphics[width=1\linewidth]{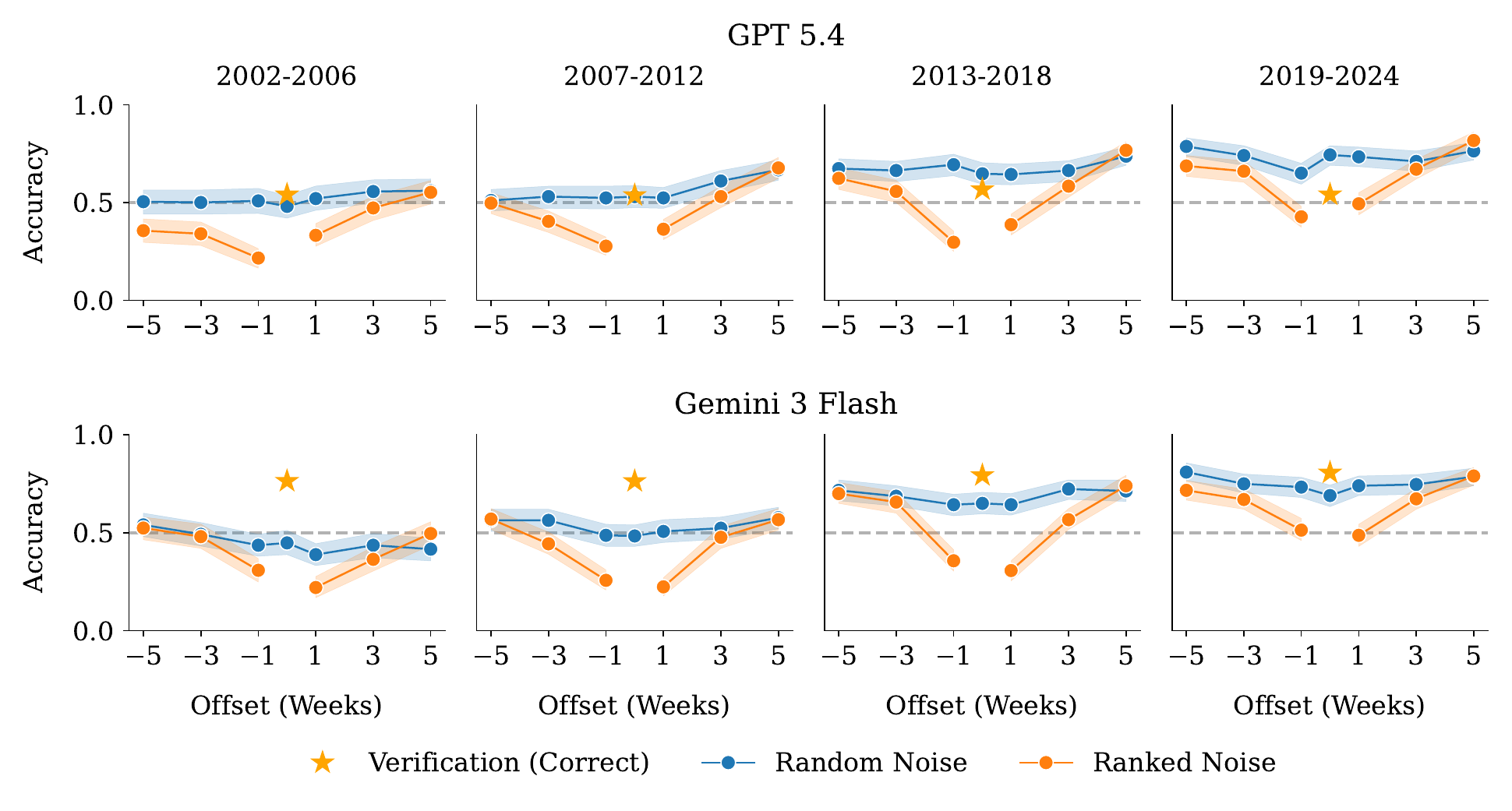}
    \vspace{-0.5em}
    \caption{
    \textbf{Residual \textit{multi-verse} state in frontier models on naturalistic updating.}
   Accuracy at \textit{rejecting} incorrect Billboard Hot 100 song-rank pairings for GPT 5.4 (top) and Gemini 3 (bottom), across four time periods. \textit{Random-noise} queries replace the correct song with a random song from the top 10; \textit{ranked-noise} queries replace it with the song that held the same rank at week $T\pm k$.
    }
    \label{fig:naturalistic:bilboard:gpt54}
\end{figure}

The Billboard Hot 100 dataset probes whether the multi-verse pattern from Section \ref{sec:results:rq3} surfaces under naturalistic updating, exploiting the fact that chart positions change weekly.
We test whether models reject verification queries built from neighboring weeks (e.g., asking whether song $s_{T \pm k}$ held rank $r$ at week $T$), comparing this \textit{ranked-noise} baseline against a \textit{random-noise} baseline that samples a different song from the top 10 (Fig. \ref{fig:naturalistic:bilboard:gpt54}). 
Models reject ranked-noise queries significantly less often than random-noise queries, with the gap peaking at $k = \pm 1$ and fading toward $k = \pm 5$ --- a residual \textit{multi-verse} pattern in frontier models.
A logistic regression confirms a robust ranked-noise effect with significant offset-by-offset variation (App. \ref{sec:app:naturalistic:billboard}).
The pattern is roughly symmetric for Gemini 3 Flash but biased toward past weeks for GPT 5.4.

\begin{takeaway}
\textbf{Frontier Takeaway.}
The three regimes from our controlled experiments surface in frontier models under natural coverage variation. 
Distillation widens the gap, increased reasoning effort does not close the gap, and naturalistic updating reveals a residual multi-verse pattern.
\end{takeaway}

\section{Discussion}
\label{sec:discussion}

We provide an extensive related work section in Appendix \ref{sec:app:related-work}.

\xhdr{Implications}
Most new content on the web will soon be produced or mediated by \LMs{} \citep{sun2025we, dolezal2026impactaigeneratedtextinternet}, meaning the data that train tomorrow's models increasingly reflect the asymmetries of today's.
A factual \gvg{} that quietly favors verification over generation, and lingers across updates, is therefore not a local property of any one model but a global dynamic shaping the shared information substrate \citep{del2024large}.
Unlike outright model collapse on recursive data \citep{shumailov2024ai}, this is a subtler structural asymmetry in what models verify as true.
Our controlled experiments show the gap converges with sufficient exposure and that frontier models perform well on well-covered facts.
Current shortcomings are therefore unlikely to be fundamental to transformers, but signal progress is achievable through better data curation and training strategies.

This optimism has limits at the small end. Parametric memory is fundamentally bounded \citep{allen-zhu2025physics33}, and small models can thus not be expected to store all factual knowledge their users may ask about.
The desired behavior in such cases is abstention rather than confabulation, a strategy not generally rewarded by most benchmarks \citep{zhou2024larger}.
Indeed, we find that even when small models like GPT-5.4 Nano do abstain on generation queries, they do not extend this behavior to verification (App.~ \ref{sec:app:naturalistic:refusals}).

\xhdr{Rethinking how we measure factuality}
Our findings suggest that when it comes to factuality, verification and generation should be treated as distinct capabilities with different learning dynamics, rather than two facets of the same underlying competence. 
Yet current evaluation practice rarely measures them separately, and the data-exposure thresholds at which each emerges are not visible from training loss alone (Section \ref{sec:results:rq1}).
A better understanding of this gap calls for testbeds and datasets that track factual capabilities not only at a frozen checkpoint on a frozen set of facts, but as both models and facts evolve. 
It also calls for mechanistic model accounts of the gap, complementing our training-mechanism framing; our setup offers a controlled probe for such mechanistic studies.

\xhdr{Limitations}
We focused on single-hop facts as a clean testbed for isolating \gvg{} dynamics; multi-hop facts entangle factual recall and recognition with multi-step reasoning \citep{allen2023physics32}, and we leave their study to future work.
Our facts are also presented as short paraphrased sentences explicitly containing each triplet, cleaner and denser than naturally occurring training data.
We conjecture this makes our setup a conservative lower bound on the dynamics we report.
Our setup also injects facts during post-training rather than pretraining.
We view it as representative of the increasingly common ``mid-training'' phase \citep{abdin2024phi, mo2025mid}, and our naturalistic results suggest similar dynamics surface at the frontier scale, although a direct test at pretraining scale remains open.
Recent frontier models also started to recognize evaluation contexts \citep{anthropic2026browsecomp}, a confounder we also observed (Fig. \ref{prompt:gemini:recognition}) that complicates interpretation for the most capable models.
Finally, we did not evaluate mitigations such as retrieval-augmented generation (RAG) \citep{lewis2020retrieval} or self-improvement methods like Best-of-$N$.
RAG can sidestep some of the dynamics we identify but adds retrieval and serving overhead, increasingly relies on \LM{}-generated source material, and still requires verification capabilities to select among retrieved candidates.
Self-improvement methods are themselves likely shaped by the asymmetries we report and warrant separate study.

\section{Conclusion}
\label{sec:conclusion}
Language models are replacing search engines as our default interface to factual knowledge, but they do not treat knowledge uniformly.
This work focused on the factual \gvg{}, whose origins trace to specific data and whose strength can be objectively measured.
Across four open model families and a controlled training life cycle of synthetic facts, we find that verification is learned before generation, survives continual learning more robustly, and may enable a ``multi-verse'' where superseded answers remain verified as true.
Naturalistic experiments on deployed flagship models reproduce these regimes, and added reasoning effort does not appear to close the gap.
We do not believe the gap is a fundamental limit of transformers, but rather a limitation of how we curate data and train models.
As we offload more factual cognition to \LMs{}, more of what they read tomorrow is what they wrote today --- a recipe for a future of proliferating factual disputes and a quiet drift in what counts as usable knowledge.
By open-sourcing our setup for reliable fact injection
we hope to give the community a shared instrument for studying the factual GV-gap.

\section*{Acknowledgements}
The authors thank Razvan Pascanu for invaluable discussions and advice throughout this project, and Robert West, Marija Šakota, and George Tsoukalas for thoughtful feedback on earlier drafts.

\bibliographystyle{unsrtnat}
\bibliography{main}

@article{ruis2024procedural,
  title={Procedural knowledge in pretraining drives reasoning in large language models},
  author={Ruis, Laura and Mozes, Maximilian and Bae, Juhan and Kamalakara, Siddhartha Rao and Talupuru, Dwarak and Locatelli, Acyr and Kirk, Robert and Rockt{\"a}schel, Tim and Grefenstette, Edward and Bartolo, Max},
  journal={ICLR},
  year={2025}
}

@incollection{karp2009reducibility,
  title={Reducibility among combinatorial problems},
  author={Karp, Richard M},
  booktitle={50 Years of Integer Programming 1958-2008: from the Early Years to the State-of-the-Art},
  pages={219--241},
  year={1972},
  publisher={Springer}
}

@article{yonelinas2002nature,
  title={The nature of recollection and familiarity: A review of 30 years of research},
  author={Yonelinas, Andrew P},
  journal={Journal of memory and language},
  volume={46},
  number={3},
  pages={441--517},
  year={2002},
  publisher={Elsevier}
}

@article{snell2024scaling,
  title={Scaling llm test-time compute optimally can be more effective than scaling model parameters},
  author={Snell, Charlie and Lee, Jaehoon and Xu, Kelvin and Kumar, Aviral},
  journal={ICLR},
  year={2025}
}

@inproceedings{
saadfalcon2025shrink,
title={Weaver: Shrinking the Generation-Verification Gap by Scaling Compute for Verification},
author={Jon Saad-Falcon and E. Kelly Buchanan and Mayee F Chen and Tzu-Heng Huang and Brendan McLaughlin and Tanvir Bhathal and Shang Zhu and Ben Athiwaratkun and Frederic Sala and Scott Linderman and Azalia Mirhoseini and Christopher Re},
booktitle={The Thirty-ninth Annual Conference on Neural Information Processing Systems},
year={2026},
url={https://openreview.net/forum?id=dRjt4vlYVQ}
}

@article{jacoby1981relationship,
  title={On the relationship between autobiographical memory and perceptual learning.},
  author={Jacoby, Larry L and Dallas, Mark},
  journal={Journal of Experimental Psychology: General},
  volume={110},
  number={3},
  pages={306},
  year={1981},
  publisher={American Psychological Association}
}

@article{anderson1972recognition,
  title={Recognition and retrieval processes in free recall.},
  author={Anderson, John R and Bower, Gordon H},
  journal={Psychological review},
  volume={79},
  number={2},
  pages={97},
  year={1972},
  publisher={American Psychological Association}
}

@incollection{atkinson2024search,
  title={Search Processes In Recognition Memory 1},
  author={Atkinson, Richard C and Herrmann, Douglas J and Wescourt, Keith T},
  booktitle={Theories in cognitive psychology},
  pages={101--146},
  year={1974},
  publisher={Routledge}
}

@inproceedings{elsahar2018t,
  title={T-rex: A large scale alignment of natural language with knowledge base triples},
  author={Elsahar, Hady and Vougiouklis, Pavlos and Remaci, Arslen and Gravier, Christophe and Hare, Jonathon and Laforest, Frederique and Simperl, Elena},
  booktitle={Proceedings of the Eleventh International Conference on Language Resources and Evaluation (LREC 2018)},
  year={2018}
}

@article{davidson2026reasoning,
  title={Reasoning-Driven Synthetic Data Generation and Evaluation},
  author={Davidson, Tim R and Seguin, Benoit and Bacis, Enrico and Ilharco, Cesar and Harkous, Hamza},
  journal={TMLR},
  year={2026}
}

@article{team2025gemma,
  title={Gemma 3 technical report},
  author={Team, Gemma and Kamath, Aishwarya and Ferret, Johan and Pathak, Shreya and Vieillard, Nino and Merhej, Ramona and Perrin, Sarah and Matejovicova, Tatiana and Ram{\'e}, Alexandre and Rivi{\`e}re, Morgane and others},
  journal={arXiv preprint arXiv:2503.19786},
  year={2025}
}

@article{yang2025qwen3,
  title={Qwen3 technical report},
  author={Yang, An and Li, Anfeng and Yang, Baosong and Zhang, Beichen and Hui, Binyuan and Zheng, Bo and Yu, Bowen and Gao, Chang and Huang, Chengen and Lv, Chenxu and others},
  journal={arXiv preprint arXiv:2505.09388},
  year={2025}
}

@article{grattafiori2024llama,
  title={The llama 3 herd of models},
  author={Grattafiori, Aaron and Dubey, Abhimanyu and Jauhri, Abhinav and Pandey, Abhinav and Kadian, Abhishek and Al-Dahle, Ahmad and Letman, Aiesha and Mathur, Akhil and Schelten, Alan and Vaughan, Alex and others},
  journal={arXiv preprint arXiv:2407.21783},
  year={2024}
}

@misc{meta2024llama32,
  title        = {Llama 3.2: Revolutionizing Edge {AI} and Vision with Open, Customizable Models},
  author       = {{Meta AI}},
  year         = {2024},
  month        = {September},
  url          = {https://ai.meta.com/blog/llama-3-2-connect-2024-vision-edge-mobile-devices},
  note         = {Blog post}
}

@incollection{cook1971complexity,
author = {Cook, Stephen A.},
title = {The complexity of theorem-proving procedures},
year = {1971},
isbn = {9781450374644},
publisher = {Association for Computing Machinery},
address = {New York, NY, USA},
url = {https://doi.org/10.1145/800157.805047},
doi = {10.1145/800157.805047},
booktitle = {Proceedings of the Third Annual ACM Symposium on Theory of Computing},
pages = {151–158},
numpages = {8},
location = {Shaker Heights, Ohio, USA},
series = {STOC '71}
}

@article{levin1973universal,
  title={Universal sequential search problems},
  author={Levin, Leonid A},
  journal={Problems of information transmission},
  volume={9},
  number={3},
  pages={265--266},
  year={1973}
}

@article{cabeza1997functional,
  title={Functional neuroanatomy of recall and recognition: A PET study of episodic memory},
  author={Cabeza, Roberto and Kapur, Shitij and Craik, Fergus IM and McIntosh, Anthony R and Houle, Sylvain and Tulving, Endel},
  journal={Journal of cognitive neuroscience},
  volume={9},
  number={2},
  pages={254--265},
  year={1997},
  publisher={MIT Press One Rogers Street, Cambridge, MA 02142-1209, USA journals-info~…}
}

@book{Kant1790-KANCOJ-2,
	address = {New York},
	author = {Immanuel Kant},
	publisher = {Barnes \& Noble},
	title = {Critique of Judgment},
	year = {1790}
}

@article{davidson2024self,
  title={Self-recognition in language models},
  author={Davidson, Tim R and Surkov, Viacheslav and Veselovsky, Veniamin and Russo, Giuseppe and West, Robert and Gulcehre, Caglar},
  journal={EMNLP},
  year={2024}
}

@misc{cobbe_training_2021,
	title = {Training {Verifiers} to {Solve} {Math} {Word} {Problems}},
	url = {http://arxiv.org/abs/2110.14168},
	doi = {10.48550/arXiv.2110.14168},
	urldate = {2025-07-25},
	publisher = {arXiv},
	author = {Cobbe, Karl and Kosaraju, Vineet and Bavarian, Mohammad and Chen, Mark and Jun, Heewoo and Kaiser, Lukasz and Plappert, Matthias and Tworek, Jerry and Hilton, Jacob and Nakano, Reiichiro and Hesse, Christopher and Schulman, John},
	month = nov,
	year = {2021},
	note = {arXiv:2110.14168},
}

@article{kirkpatrick2017overcoming,
  title={Overcoming catastrophic forgetting in neural networks},
  author={Kirkpatrick, James and Pascanu, Razvan and Rabinowitz, Neil and Veness, Joel and Desjardins, Guillaume and Rusu, Andrei A and Milan, Kieran and Quan, John and Ramalho, Tiago and Grabska-Barwinska, Agnieszka and others},
  journal={Proceedings of the national academy of sciences},
  volume={114},
  number={13},
  pages={3521--3526},
  year={2017},
  publisher={National Academy of Sciences}
}

@article{lopez2017gradient,
  title={Gradient episodic memory for continual learning},
  author={Lopez-Paz, David and Ranzato, Marc'Aurelio},
  journal={Advances in neural information processing systems},
  volume={30},
  year={2017}
}

@inproceedings{scialom2022fine,
  title={Fine-tuned language models are continual learners},
  author={Scialom, Thomas and Chakrabarty, Tuhin and Muresan, Smaranda},
  booktitle={Proceedings of the 2022 Conference on Empirical Methods in Natural Language Processing},
  pages={6107--6122},
  year={2022}
}

@article{shi2025continual,
  title={Continual learning of large language models: A comprehensive survey},
  author={Shi, Haizhou and Xu, Zihao and Wang, Hengyi and Qin, Weiyi and Wang, Wenyuan and Wang, Yibin and Wang, Zifeng and Ebrahimi, Sayna and Wang, Hao},
  journal={ACM Computing Surveys},
  volume={58},
  number={5},
  pages={1--42},
  year={2025},
  publisher={ACM New York, NY}
}

@inproceedings{
lightman2024lets,
title={Let's Verify Step by Step},
author={Hunter Lightman and Vineet Kosaraju and Yuri Burda and Harrison Edwards and Bowen Baker and Teddy Lee and Jan Leike and John Schulman and Ilya Sutskever and Karl Cobbe},
booktitle={ICLR},
year={2024},
url={https://openreview.net/forum?id=v8L0pN6EOi}
}

@misc{swamy_all_2025,
	title = {All {Roads} {Lead} to {Likelihood}: {The} {Value} of {Reinforcement} {Learning} in {Fine}-{Tuning}},
	shorttitle = {All {Roads} {Lead} to {Likelihood}},
	url = {http://arxiv.org/abs/2503.01067},
	doi = {10.48550/arXiv.2503.01067},
	urldate = {2025-06-22},
	publisher = {arXiv},
	author = {Swamy, Gokul and Choudhury, Sanjiban and Sun, Wen and Wu, Zhiwei Steven and Bagnell, J. Andrew},
	month = mar,
	year = {2025},
	note = {arXiv:2503.01067},
}

@inproceedings{hosseini_v-star_2024,
	title = {V-{STaR}: {Training} {Verifiers} for {Self}-{Taught} {Reasoners}},
	shorttitle = {V-{STaR}},
	url = {https://openreview.net/forum?id=stmqBSW2dV#discussion},
	language = {en},
	urldate = {2025-06-22},
	author = {Hosseini, Arian and Yuan, Xingdi and Malkin, Nikolay and Courville, Aaron and Sordoni, Alessandro and Agarwal, Rishabh},
    booktitle={COLM},
	month = aug,
	year = {2024},
}

@inproceedings{nichani_understanding_2024,
	title = {Understanding {Factual} {Recall} in {Transformers} via {Associative} {Memories}},
	url = {https://openreview.net/forum?id=hwSmPOAmhk},
	booktitle={ICLR},
	author = {Nichani, Eshaan and Lee, Jason D. and Bietti, Alberto},
	month = oct,
	year = {2025},
}

@article{chang2024large,
  title={How do large language models acquire factual knowledge during pretraining?},
  author={Chang, Hoyeon and Park, Jinho and Ye, Seonghyeon and Yang, Sohee and Seo, Youngkyung and Chang, Du-Seong and Seo, Minjoon},
  journal={Advances in neural information processing systems},
  volume={37},
  pages={60626--60668},
  year={2024}
}

@inproceedings{west_generative_2024,
	title = {The {Generative} {AI} {Paradox}: “{What} {It} {Can} {Create}, {It} {May} {Not} {Understand}”},
	url = {https://openreview.net/forum?id=CF8H8MS5P8},
	booktitle = {The {Twelfth} {International} {Conference} on {Learning} {Representations}},
	author = {West, Peter and Lu, Ximing and Dziri, Nouha and Brahman, Faeze and Li, Linjie and Hwang, Jena D. and Jiang, Liwei and Fisher, Jillian and Ravichander, Abhilasha and Chandu, Khyathi and Newman, Benjamin and Koh, Pang Wei and Ettinger, Allyson and Choi, Yejin},
	year = {2024},
}

@inproceedings{song_mind_nodate,
	title = {Mind the {Gap}: {Examining} the {Self}-{Improvement} {Capabilities} of {Large} {Language} {Models}},
	url = {https://openreview.net/pdf?id=mtJSMcF3ek},
	booktitle = {ICLR},
    year={2025},
	author = {Song, Yuda and Zhang, Hanlin and Ghai, Udaya and Eisenach, Carson and Kakade, Sham M and Foster, Dean},
}

@misc{morris_how_2025,
	title = {How much do language models memorize?},
	url = {http://arxiv.org/abs/2505.24832},
	doi = {10.48550/arXiv.2505.24832},
	urldate = {2025-06-22},
	publisher = {arXiv},
	author = {Morris, John X. and Sitawarin, Chawin and Guo, Chuan and Kokhlikyan, Narine and Suh, G. Edward and Rush, Alexander M. and Chaudhuri, Kamalika and Mahloujifar, Saeed},
	month = jun,
	year = {2025},
	note = {arXiv:2505.24832},
}

@article{allen2023physics31,
  title={Physics of language models: Part 3.1, knowledge storage and extraction},
  author={Allen-Zhu, Zeyuan and Li, Yuanzhi},
  journal={ICML},
  year={2024}
}

@article{allen2023physics32,
  title={Physics of language models: Part 3.2, knowledge manipulation},
  author={Allen-Zhu, Zeyuan and Li, Yuanzhi},
  journal={ICLR},
  year={2025}
}

@inproceedings{
allen-zhu2025physics33,
title={Physics of Language Models: Part 3.3, Knowledge Capacity Scaling Laws},
author={Zeyuan Allen-Zhu and Yuanzhi Li},
booktitle={The Thirteenth International Conference on Learning Representations},
year={2025},
url={https://openreview.net/forum?id=FxNNiUgtfa}
}

@inproceedings{
gandhi_cognitive_2025,
title={Cognitive Behaviors that Enable Self-Improving Reasoners, or, Four Habits of Highly Effective {ST}aRs},
author={Kanishk Gandhi and Ayush K Chakravarthy and Anikait Singh and Nathan Lile and Noah Goodman},
booktitle={Second Conference on Language Modeling},
year={2025},
url={https://openreview.net/forum?id=QGJ9ttXLTy}
}

@article{zucchet2025language,
  title={How do language models learn facts? Dynamics, curricula and hallucinations},
  author={Zucchet, Nicolas and Bornschein, J{\"o}rg and Chan, Stephanie and Lampinen, Andrew and Pascanu, Razvan and De, Soham},
  journal={arXiv preprint arXiv:2503.21676},
  year={2025}
}

@inproceedings{
gu2025data,
title={Data Mixing Can Induce Phase Transitions in Knowledge Acquisition},
author={Xinran Gu and Kaifeng Lyu and Jiazheng Li and Jingzhao Zhang},
booktitle={The Thirty-ninth Annual Conference on Neural Information Processing Systems},
year={2026},
url={https://openreview.net/forum?id=tQZK5frjVU}
}

@article{geva2023dissecting,
  title={Dissecting recall of factual associations in auto-regressive language models},
  author={Geva, Mor and Bastings, Jasmijn and Filippova, Katja and Globerson, Amir},
  journal={EMNLP},
  year={2023}
}

@incollection{mccloskey1989catastrophic,
  title={Catastrophic interference in connectionist networks: The sequential learning problem},
  author={McCloskey, Michael and Cohen, Neal J},
  booktitle={Psychology of learning and motivation},
  volume={24},
  pages={109--165},
  year={1989},
  publisher={Elsevier}
}

@inproceedings{ovadia2024fine,
  title={Fine-tuning or retrieval? comparing knowledge injection in llms},
  author={Ovadia, Oded and Brief, Menachem and Mishaeli, Moshik and Elisha, Oren},
  booktitle={Proceedings of the 2024 conference on empirical methods in natural language processing},
  pages={237--250},
  year={2024}
}

@inproceedings{gekhman-etal-2024-fine,
    title = "Does Fine-Tuning {LLM}s on New Knowledge Encourage Hallucinations?",
    author = "Gekhman, Zorik  and
      Yona, Gal  and
      Aharoni, Roee  and
      Eyal, Matan  and
      Feder, Amir  and
      Reichart, Roi  and
      Herzig, Jonathan",
    booktitle = "EMNLP",
    year = "2024",
    publisher = "ACL",
    url = "https://aclanthology.org/2024.emnlp-main.444/"
}

@article{bietti2023birth,
  title={Birth of a transformer: A memory viewpoint},
  author={Bietti, Alberto and Cabannes, Vivien and Bouchacourt, Diane and Jegou, Herve and Bottou, Leon},
  journal={Advances in Neural Information Processing Systems},
  volume={36},
  pages={1560--1588},
  year={2023}
}

@article{de2023continual,
  title={Continual evaluation for lifelong learning: Identifying the stability gap},
  author={De Lange, Matthias and van de Ven, Gido and Tuytelaars, Tinne},
  journal={ICLR},
  year={2023}
}

@book{grossberg2012studies,
  title={Studies of mind and brain: Neural principles of learning, perception, development, cognition, and motor control},
  author={Grossberg, Stephen T},
  volume={70},
  year={2012},
  publisher={Springer Science \& Business Media}
}

@inproceedings{chaudhry2019continual,
  title={Continual learning with tiny episodic memories},
  author={Chaudhry, Arslan and Rohrbach, Marcus and Elhoseiny, Mohamed and Ajanthan, Thalaiyasingam and Dokania, P and Torr, P and Ranzato, M},
  booktitle={Workshop on Multi-Task and Lifelong Reinforcement Learning},
  year={2019}
}

@article{van2022three,
  title={Three types of incremental learning},
  author={Van de Ven, Gido M and Tuytelaars, Tinne and Tolias, Andreas S},
  journal={Nature Machine Intelligence},
  volume={4},
  number={12},
  pages={1185--1197},
  year={2022},
  publisher={Nature Publishing Group UK London}
}

@article{zelikman2022star,
  title={Star: Bootstrapping reasoning with reasoning},
  author={Zelikman, Eric and Wu, Yuhuai and Mu, Jesse and Goodman, Noah},
  journal={Advances in Neural Information Processing Systems},
  volume={35},
  pages={15476--15488},
  year={2022}
}

@article{li2022competition,
  title={Competition-level code generation with alphacode},
  author={Li, Yujia and Choi, David and Chung, Junyoung and Kushman, Nate and Schrittwieser, Julian and Leblond, R{\'e}mi and Eccles, Tom and Keeling, James and Gimeno, Felix and Dal Lago, Agustin and others},
  journal={Science},
  volume={378},
  number={6624},
  pages={1092--1097},
  year={2022},
  publisher={American Association for the Advancement of Science}
}

@article{brown2024large,
  title={Large language monkeys: Scaling inference compute with repeated sampling},
  author={Brown, Bradley and Juravsky, Jordan and Ehrlich, Ryan and Clark, Ronald and Le, Quoc V and R{\'e}, Christopher and Mirhoseini, Azalia},
  journal={arXiv preprint arXiv:2407.21787},
  year={2024}
}

@article{puri2025probabilistic,
  title={Rollout Roulette: A Probabilistic Inference Approach to Inference-Time Scaling of LLMs using Particle-Based Monte Carlo Methods},
  author={Puri, Isha and Sudalairaj, Shivchander and Xu, Guangxuan and Xu, Kai and Srivastava, Akash},
  journal={NeurIPS},
  year={2025}
}

@inproceedings{
lifshitz2025multiagent,
title={Multi-Agent Verification: Scaling Test-Time Compute with Goal Verifiers},
author={Shalev Lifshitz and Sheila A. McIlraith and Yilun Du},
booktitle={ICLR 2025 Workshop on Modularity for Collaborative, Decentralized, and Continual Deep Learning},
year={2025},
url={https://openreview.net/forum?id=mGAAoEWOq9}
}

@article{stroebl2024inference,
  title={Inference scaling flaws: The limits of llm resampling with imperfect verifiers},
  author={Stroebl, Benedikt and Kapoor, Sayash and Narayanan, Arvind},
  journal={arXiv preprint arXiv:2411.17501},
  year={2024}
}

@inproceedings{yean24selfrewardllm,
author = {Yuan, Weizhe and Pang, Richard Yuanzhe and Cho, Kyunghyun and Li, Xian and Sukhbaatar, Sainbayar and Xu, Jing and Weston, Jason},
title = {Self-rewarding language models},
year = {2024},
booktitle = {ICML}
}

@article{kadavath2022language,
  title={Language models (mostly) know what they know},
  author={Kadavath, Saurav and Conerly, Tom and Askell, Amanda and Henighan, Tom and Drain, Dawn and Perez, Ethan and Schiefer, Nicholas and Hatfield-Dodds, Zac and DasSarma, Nova and Tran-Johnson, Eli and others},
  journal={arXiv preprint arXiv:2207.05221},
  year={2022}
}

@article{meng2022locating,
  title={Locating and editing factual associations in gpt},
  author={Meng, Kevin and Bau, David and Andonian, Alex and Belinkov, Yonatan},
  journal={Advances in neural information processing systems},
  volume={35},
  pages={17359--17372},
  year={2022}
}

@inproceedings{mitchell2022memory,
  title={Memory-based model editing at scale},
  author={Mitchell, Eric and Lin, Charles and Bosselut, Antoine and Manning, Christopher D and Finn, Chelsea},
  booktitle={International Conference on Machine Learning},
  pages={15817--15831},
  year={2022},
  organization={PMLR}
}

@inproceedings{zhong2023mquake,
  title={Mquake: Assessing knowledge editing in language models via multi-hop questions},
  author={Zhong, Zexuan and Wu, Zhengxuan and Manning, Christopher D and Potts, Christopher and Chen, Danqi},
  booktitle={Proceedings of the 2023 Conference on Empirical Methods in Natural Language Processing},
  pages={15686--15702},
  year={2023}
}

@inproceedings{hoelscher2023detecting,
  title={Detecting edit failures in large language models: An improved specificity benchmark},
  author={Hoelscher-Obermaier, Jason and Persson, Julia and Kran, Esben and Konstas, Ioannis and Barez, Fazl},
  booktitle={Findings of the Association for Computational Linguistics: ACL 2023},
  pages={11548--11559},
  year={2023}
}

@article{cohen2024evaluating,
  title={Evaluating the ripple effects of knowledge editing in language models},
  author={Cohen, Roi and Biran, Eden and Yoran, Ori and Globerson, Amir and Geva, Mor},
  journal={Transactions of the Association for Computational Linguistics},
  volume={12},
  pages={283--298},
  year={2024},
  publisher={MIT Press One Broadway, 12th Floor, Cambridge, Massachusetts 02142, USA~…}
}

@inproceedings{petroni2019language,
  title={Language models as knowledge bases?},
  author={Petroni, Fabio and Rockt{\"a}schel, Tim and Riedel, Sebastian and Lewis, Patrick and Bakhtin, Anton and Wu, Yuxiang and Miller, Alexander},
  booktitle={Proceedings of the 2019 conference on empirical methods in natural language processing and the 9th international joint conference on natural language processing (EMNLP-IJCNLP)},
  pages={2463--2473},
  year={2019}
}

@article{jiang2020can,
  title={How can we know what language models know?},
  author={Jiang, Zhengbao and Xu, Frank F and Araki, Jun and Neubig, Graham},
  journal={Transactions of the Association for Computational Linguistics},
  volume={8},
  pages={423--438},
  year={2020},
  publisher={MIT Press One Rogers Street, Cambridge, MA 02142-1209, USA journals-info~…}
}

@article{roberts2020much,
  title={How much knowledge can you pack into the parameters of a language model?},
  author={Roberts, Adam and Raffel, Colin and Shazeer, Noam},
  journal={EMNLP},
  year={2020}
}

@article{de2021editing,
  title={Editing factual knowledge in language models},
  author={De Cao, Nicola and Aziz, Wilker and Titov, Ivan},
  journal={EMNLP},
  year={2021}
}

@inproceedings{
tirumala2022memorization,
title={Memorization Without Overfitting:  Analyzing the Training Dynamics of Large Language Models},
author={Kushal Tirumala and Aram H. Markosyan and Luke Zettlemoyer and Armen Aghajanyan},
booktitle={Advances in Neural Information Processing Systems},
editor={Alice H. Oh and Alekh Agarwal and Danielle Belgrave and Kyunghyun Cho},
year={2022},
url={https://openreview.net/forum?id=u3vEuRr08MT}
}

@inproceedings{geva2021transformer,
  title={Transformer feed-forward layers are key-value memories},
  author={Geva, Mor and Schuster, Roei and Berant, Jonathan and Levy, Omer},
  booktitle={EMNLP},
  pages={5484--5495},
  year={2021}
}

@inproceedings{dai2022knowledge,
  title={Knowledge neurons in pretrained transformers},
  author={Dai, Damai and Dong, Li and Hao, Yaru and Sui, Zhifang and Chang, Baobao and Wei, Furu},
  booktitle={Proceedings of the 60th Annual Meeting of the Association for Computational Linguistics (Volume 1: Long Papers)},
  pages={8493--8502},
  year={2022}
}

@article{meng2022mass,
  title={Mass-editing memory in a transformer},
  author={Meng, Kevin and Sharma, Arnab Sen and Andonian, Alex and Belinkov, Yonatan and Bau, David},
  journal={ICLR},
  year={2023}
}

@article{tian2024toward,
  title={Toward self-improvement of llms via imagination, searching, and criticizing},
  author={Tian, Ye and Peng, Baolin and Song, Linfeng and Jin, Lifeng and Yu, Dian and Han, Lei and Mi, Haitao and Yu, Dong},
  journal={Advances in Neural Information Processing Systems},
  volume={37},
  pages={52723--52748},
  year={2024}
}

@inproceedings{huang2023large,
  title={Large language models can self-improve},
  author={Huang, Jiaxin and Gu, Shixiang and Hou, Le and Wu, Yuexin and Wang, Xuezhi and Yu, Hongkun and Han, Jiawei},
  booktitle={EMNLP},
  pages={1051--1068},
  year={2023}
}

@inproceedings{
pang2024language,
title={Language Model Self-improvement by Reinforcement Learning Contemplation},
author={Jing-Cheng Pang and Pengyuan Wang and Kaiyuan Li and Xiong-Hui Chen and Jiacheng Xu and Zongzhang Zhang and Yang Yu},
booktitle={ICLR},
year={2024},
url={https://openreview.net/forum?id=38E4yUbrgr}
}

@article{liu2024large,
  title={Large language models have intrinsic self-correction ability},
  author={Liu, Dancheng and Nassereldine, Amir and Yang, Ziming and Xu, Chenhui and Hu, Yuting and Li, Jiajie and Kumar, Utkarsh and Lee, Changjae and Qin, Ruiyang and Shi, Yiyu and others},
  journal={arXiv preprint arXiv:2406.15673},
  year={2024}
}

@article{kumar2024training,
  title={Training language models to self-correct via reinforcement learning},
  author={Kumar, Aviral and Zhuang, Vincent and Agarwal, Rishabh and Su, Yi and Co-Reyes, John D and Singh, Avi and Baumli, Kate and Iqbal, Shariq and Bishop, Colton and Roelofs, Rebecca and others},
  journal={ICLR},
  year={2025}
}

@article{huang2025self,
  title={Self-improvement in language models: The sharpening mechanism},
  author={Huang, Audrey and Block, Adam and Foster, Dylan J and Rohatgi, Dhruv and Zhang, Cyril and Simchowitz, Max and Ash, Jordan T and Krishnamurthy, Akshay},
  journal={ICLR},
  year={2025}
}

@article{sun2025theoretical,
  title={Theoretical Modeling of LLM Self-Improvement Training Dynamics Through Solver-Verifier Gap},
  author={Sun, Yifan and Liang, Yushan and Zhang, Zhen and Teng, Jiaye},
  journal={arXiv preprint arXiv:2507.00075},
  year={2025}
}

@article{kalajdzievski2024scaling,
  title={Scaling laws for forgetting when fine-tuning large language models},
  author={Kalajdzievski, Damjan},
  journal={arXiv preprint arXiv:2401.05605},
  year={2024}
}

@article{
ibrahim2024simple,
title={Simple and Scalable Strategies to Continually Pre-train Large Language Models},
author={Adam Ibrahim and Benjamin Th{\'e}rien and Kshitij Gupta and Mats Leon Richter and Quentin Gregory Anthony and Eugene Belilovsky and Timoth{\'e}e Lesort and Irina Rish},
journal={Transactions on Machine Learning Research},
issn={2835-8856},
year={2024},
url={https://openreview.net/forum?id=DimPeeCxKO},
}

@article{luo2025empirical,
  title={An empirical study of catastrophic forgetting in large language models during continual fine-tuning},
  author={Luo, Yun and Yang, Zhen and Meng, Fandong and Li, Yafu and Zhou, Jie and Zhang, Yue},
  journal={IEEE Transactions on Audio, Speech and Language Processing},
  year={2025},
  publisher={IEEE}
}

@inproceedings{
zhang2024when,
title={When Scaling Meets {LLM} Finetuning: The Effect of Data, Model and Finetuning Method},
author={Biao Zhang and Zhongtao Liu and Colin Cherry and Orhan Firat},
booktitle={The Twelfth International Conference on Learning Representations},
year={2024},
url={https://openreview.net/forum?id=5HCnKDeTws}
}

@article{scalingdatalms2025muennighof,
  author  = {Niklas Muennighoff and Alexander M. Rush and Boaz Barak and Teven Le Scao and Aleksandra Piktus and Nouamane Tazi and Sampo Pyysalo and Thomas Wolf and Colin Raffel},
  title   = {Scaling Data-Constrained Language Models},
  journal = {Journal of Machine Learning Research},
  year    = {2025},
  volume  = {26},
  number  = {53},
  pages   = {1--66},
  url     = {http://jmlr.org/papers/v26/24-1000.html}
}

@article{loshchilov2017decoupled,
  title={Decoupled weight decay regularization},
  author={Loshchilov, Ilya and Hutter, Frank},
  journal={ICLR},
  year={2019}
}

@article{loshchilov2016sgdr,
  title={Sgdr: Stochastic gradient descent with warm restarts},
  author={Loshchilov, Ilya and Hutter, Frank},
  journal={ICLR},
  year={2017}
}

@article{bethune2025scaling,
  title={Scaling laws for forgetting during finetuning with pretraining data injection},
  author={Bethune, Louis and Grangier, David and Busbridge, Dan and Gualdoni, Eleonora and Cuturi, Marco and Ablin, Pierre},
  journal={ICML},
  year={2025}
}

@misc{thingsnotstrings2012,
    author={Amit Singhal},
    title={Introducing the Knowledge Graph: things, not strings},
    year={2012},
    url={https://blog.google/products/search/introducing-knowledge-graph-things-not/}
}

@misc{bingsatori2023,
    author={Richard Qian},
    title={Understand your world with Bing},
    url={https://blogs.bing.com/search/March-2013/Understand-Your-World-with-Bing},
    year={2013}
}

@inproceedings{
zhao2025sample,
title={Sample, Scrutinize and Scale: Effective Inference-Time Search by Scaling Verification},
author={Eric Zhao and Pranjal Awasthi and Sreenivas Gollapudi},
booktitle={Forty-second International Conference on Machine Learning},
year={2025},
url={https://openreview.net/forum?id=wl3eI4wiE5}
}

@article{
chen2025sets,
title={{SETS}: Leveraging Self-Verification and Self-Correction for Improved Test-Time Scaling},
author={Jiefeng Chen and Jie Ren and Xinyun Chen and Chengrun Yang and Ruoxi Sun and Jinsung Yoon and Sercan O Arik},
journal={Transactions on Machine Learning Research},
issn={2835-8856},
year={2025},
url={https://openreview.net/forum?id=Wv9NMJoKww},
note={}
}

@inproceedings{harman2002development,
  title={The Development and Evolution of TREC and DUC.},
  author={Harman, Donna and Over, Paul},
  booktitle={NTCIR},
  year={2002}
}

@article{dai2020survey,
  title={A survey on knowledge graph embedding: Approaches, applications and benchmarks},
  author={Dai, Yuanfei and Wang, Shiping and Xiong, Neal N and Guo, Wenzhong},
  journal={Electronics},
  volume={9},
  number={5},
  pages={750},
  year={2020},
  publisher={MDPI}
}

@article{zhang2025generative,
  title={Generative verifiers: Reward modeling as next-token prediction},
  author={Zhang, Lunjun and Hosseini, Arian and Bansal, Hritik and Kazemi, Mehran and Kumar, Aviral and Agarwal, Rishabh},
  journal={ICLR},
  year={2025}
}

@inproceedings{glockner2007university,
  title={University of Hagen at CLEF 2007: Answer Validation Exercise.},
  author={Gl{\"o}ckner, Ingo},
  booktitle={CLEF (Working Notes)},
  year={2007}
}

@article{forner2010evaluating,
  title={Evaluating multilingual question answering systems at CLEF},
  author={Forner, Pamela and Giampiccolo, Danilo and Magnini, Bernardo and Sutcliffe, Richard and Pe{\~n}as Padilla, Anselmo and Rodrigo Yuste, {\'A}lvaro},
  year={2010},
  journal={ACL}
}

@article{kamoi2024can,
  title={When can llms actually correct their own mistakes? a critical survey of self-correction of llms},
  author={Kamoi, Ryo and Zhang, Yusen and Zhang, Nan and Han, Jiawei and Zhang, Rui},
  journal={Transactions of the Association for Computational Linguistics},
  volume={12},
  pages={1417--1440},
  year={2024},
  publisher={MIT Press 255 Main Street, 9th Floor, Cambridge, Massachusetts 02142, USA~…}
}

@article{devlin2019bertpretrainingdeepbidirectional,
      title={BERT: Pre-training of Deep Bidirectional Transformers for Language Understanding}, 
      author={Jacob Devlin and Ming-Wei Chang and Kenton Lee and Kristina Toutanova},
      year={2019},
      journal={NAACL},
      url={https://arxiv.org/abs/1810.04805}, 
}

@article{zhou2024larger,
  title={Larger and more instructable language models become less reliable},
  author={Zhou, Lexin and Schellaert, Wout and Mart{\'\i}nez-Plumed, Fernando and Moros-Daval, Yael and Ferri, C{\`e}sar and Hern{\'a}ndez-Orallo, Jos{\'e}},
  journal={Nature},
  volume={634},
  number={8032},
  pages={61--68},
  year={2024},
  publisher={Nature Publishing Group UK London}
}

@article{sharma2023towards,
  title={Towards understanding sycophancy in language models},
  author={Sharma, Mrinank and Tong, Meg and Korbak, Tomasz and Duvenaud, David and Askell, Amanda and Bowman, Samuel R and Cheng, Newton and Durmus, Esin and Hatfield-Dodds, Zac and Johnston, Scott R and others},
  journal={ICLR},
  year={2024}
}

@article{comanici2025gemini,
  title={Gemini 2.5: Pushing the frontier with advanced reasoning, multimodality, long context, and next generation agentic capabilities},
  author={Comanici, Gheorghe and Bieber, Eric and Schaekermann, Mike and Pasupat, Ice and Sachdeva, Noveen and Dhillon, Inderjit and Blistein, Marcel and Ram, Ori and Zhang, Dan and Rosen, Evan and others},
  journal={arXiv preprint arXiv:2507.06261},
  year={2025}
}

@misc{google2025gemini3blog,
  title        = {A New Era of Intelligence with {Gemini} 3},
  author       = {{Google}},
  year         = {2025},
  month={November},
  url          = {https://blog.google/products-and-platforms/products/gemini/gemini-3/},
  note         = {Blog post}
}

@misc{anthropic2024claudesonnet45,
  author       = {Anthropic},
  title        = {Claude Sonnet 4.5},
  year         = {2025},
  howpublished = {\url{https://www.anthropic.com}},
  note         = {Large language model; API identifier: claude-sonnet-4-5}
}

@article{rafailov2023direct,
  title={Direct preference optimization: Your language model is secretly a reward model},
  author={Rafailov, Rafael and Sharma, Archit and Mitchell, Eric and Manning, Christopher D and Ermon, Stefano and Finn, Chelsea},
  journal={Advances in neural information processing systems},
  volume={36},
  pages={53728--53741},
  year={2023}
}

@misc{openai2026gpt54,
  title        = {Introducing {GPT-5.4}},
  author       = {{OpenAI}},
  year         = {2026},
  url          = {https://openai.com/index/introducing-gpt-5-4/},
  note         = {Blog post}
}

@article{abdin2024phi,
  title={Phi-4 technical report},
  author={Abdin, Marah and Aneja, Jyoti and Behl, Harkirat and Bubeck, S{\'e}bastien and Eldan, Ronen and Gunasekar, Suriya and Harrison, Michael and Hewett, Russell J and Javaheripi, Mojan and Kauffmann, Piero and others},
  journal={arXiv preprint arXiv:2412.08905},
  year={2024}
}

@article{abouelenin2025phi,
  title={Phi-4-mini technical report: Compact yet powerful multimodal language models via mixture-of-loras},
  author={Abouelenin, Abdelrahman and Ashfaq, Atabak and Atkinson, Adam and Awadalla, Hany and Bach, Nguyen and Bao, Jianmin and Benhaim, Alon and Cai, Martin and Chaudhary, Vishrav and Chen, Congcong and others},
  journal={arXiv preprint arXiv:2503.01743},
  year={2025}
}

@misc{megamillions,
    title={Lottery Mega Millions Winning Numbers: Beginning 2002},
    author={{State of New York}},
    url={https://catalog.data.gov/dataset/lottery-mega-millions-winning-numbers-beginning-2002},
    year={2026}
}

@article{rydning2018digitization,
  title={The digitization of the world from edge to core},
  author={Rydning, David Reinsel-John Gantz-John and Reinsel, John and Gantz, John},
  journal={Framingham: International Data Corporation},
  volume={16},
  pages={1--28},
  year={2018}
}

@article{ji2023survey,
  title={Survey of hallucination in natural language generation},
  author={Ji, Ziwei and Lee, Nayeon and Frieske, Rita and Yu, Tiezheng and Su, Dan and Xu, Yan and Ishii, Etsuko and Bang, Ye Jin and Madotto, Andrea and Fung, Pascale},
  journal={ACM computing surveys},
  volume={55},
  number={12},
  pages={1--38},
  year={2023},
  publisher={ACM New York, NY}
}

@article{fan2026halluhard,
  title={HalluHard: A Hard Multi-Turn Hallucination Benchmark},
  author={Fan, Dongyang and Delsad, Sebastien and Flammarion, Nicolas and Andriushchenko, Maksym},
  journal={arXiv preprint arXiv:2602.01031},
  year={2026}
}

@misc{aioverview24google,
    title={Generative AI in Search: Let Google do the searching for you},
    author={Elizabeth Reid},
    year={2024},
    month={May},
    howpublished={\url{https://blog.google/products-and-platforms/products/search/generative-ai-google-search-may-2024/}}
}

@misc{aioverview24bing,
    title={Introducing Bing Generative Search},
    author={Microsoft},
    year={2024},
    month={July},
    howpublished={\url{https://blogs.bing.com/search/July-2024/generativesearch}}
}

@techreport{chatterji2025people,
  title={How people use chatgpt},
  author={Chatterji, Aaron and Cunningham, Thomas and Deming, David J and Hitzig, Zoe and Ong, Christopher and Shan, Carl Yan and Wadman, Kevin},
  year={2025},
  institution={National Bureau of Economic Research}
}

@inproceedings{frank2024representative,
  title={A representative study on human detection of artificially generated media across countries},
  author={Frank, Joel and Herbert, Franziska and Ricker, Jonas and Sch{\"o}nherr, Lea and Eisenhofer, Thorsten and Fischer, Asja and D{\"u}rmuth, Markus and Holz, Thorsten},
  booktitle={2024 IEEE Symposium on Security and Privacy (SP)},
  pages={55--73},
  year={2024},
  organization={IEEE}
}

@article{
wei2022emergent,
title={Emergent Abilities of Large Language Models},
author={Jason Wei and Yi Tay and Rishi Bommasani and Colin Raffel and Barret Zoph and Sebastian Borgeaud and Dani Yogatama and Maarten Bosma and Denny Zhou and Donald Metzler and Ed H. Chi and Tatsunori Hashimoto and Oriol Vinyals and Percy Liang and Jeff Dean and William Fedus},
journal={Transactions on Machine Learning Research},
issn={2835-8856},
year={2022},
url={https://openreview.net/forum?id=yzkSU5zdwD},
note={Survey Certification}
}

@article{cooke25cointoss,
author = {Cooke, Di and Edwards, Abigail and Barkoff, Sophia and Kelly, Kathryn},
title = {As Good as a Coin Toss: Human Detection of AI-Generated Content},
year = {2025},
issue_date = {October 2025},
publisher = {Association for Computing Machinery},
address = {New York, NY, USA},
volume = {68},
number = {10},
issn = {0001-0782},
url = {https://doi.org/10.1145/3729417},
doi = {10.1145/3729417},
journal = {Commun. ACM},
month = sep,
pages = {100–109},
numpages = {10}
}

@article{wang2025octothinker,
  title={Octothinker: Mid-training incentivizes reinforcement learning scaling},
  author={Wang, Zengzhi and Zhou, Fan and Li, Xuefeng and Liu, Pengfei},
  journal={arXiv preprint arXiv:2506.20512},
  year={2025}
}

@article{zheng2023judging,
  title={Judging llm-as-a-judge with mt-bench and chatbot arena},
  author={Zheng, Lianmin and Chiang, Wei-Lin and Sheng, Ying and Zhuang, Siyuan and Wu, Zhanghao and Zhuang, Yonghao and Lin, Zi and Li, Zhuohan and Li, Dacheng and Xing, Eric and others},
  journal={Advances in neural information processing systems},
  volume={36},
  pages={46595--46623},
  year={2023}
}

@article{ghosal2024understanding,
  title={Understanding finetuning for factual knowledge extraction},
  author={Ghosal, Gaurav and Hashimoto, Tatsunori and Raghunathan, Aditi},
  journal={ICML},
  year={2024}
}

@article{farquhar2024detecting,
  title={Detecting hallucinations in large language models using semantic entropy},
  author={Farquhar, Sebastian and Kossen, Jannik and Kuhn, Lorenz and Gal, Yarin},
  journal={Nature},
  volume={630},
  number={8017},
  pages={625--630},
  year={2024},
  publisher={Nature Publishing Group UK London}
}

@article{mccandlish2018empirical,
  title={An empirical model of large-batch training},
  author={McCandlish, Sam and Kaplan, Jared and Amodei, Dario and Team, OpenAI Dota},
  journal={arXiv preprint arXiv:1812.06162},
  year={2018}
}

@article{merrill2025critical,
  title={Critical batch size revisited: A simple empirical approach to large-batch language model training},
  author={Merrill, William and Arora, Shane and Groeneveld, Dirk and Hajishirzi, Hannaneh},
  journal={NeurIPS},
  year={2026}
}

@inproceedings{
marek2026small,
title={Small Batch Size Training for Language Models: When Vanilla {SGD} Works, and Why Gradient Accumulation is Wasteful},
author={Martin Marek and Sanae Lotfi and Aditya Somasundaram and Andrew Gordon Wilson and Micah Goldblum},
booktitle={NeurIPS},
year={2026},
url={https://openreview.net/forum?id=52Ehpe0Lu5}
}

@inproceedings{
mosbach2021on,
title={On the Stability of Fine-tuning {\{}BERT{\}}: Misconceptions, Explanations, and Strong Baselines},
author={Marius Mosbach and Maksym Andriushchenko and Dietrich Klakow},
booktitle={International Conference on Learning Representations},
year={2021},
url={https://openreview.net/forum?id=nzpLWnVAyah}
}

@article{berglund2023reversal,
  title={The Reversal Curse: LLMs trained on" A is B" fail to learn" B is A"},
  author={Berglund, Lukas and Tong, Meg and Kaufmann, Max and Balesni, Mikita and Stickland, Asa Cooper and Korbak, Tomasz and Evans, Owain},
  journal={ICLR},
  year={2024}
}

@inproceedings{sun2025we,
  title={Are we in the AI-generated text world already? Quantifying and monitoring AIGT on social media},
  author={Sun, Zhen and Zhang, Zongmin and Shen, Xinyue and Zhang, Ziyi and Liu, Yule and Backes, Michael and Zhang, Yang and He, Xinlei},
  booktitle={Proceedings of the 63rd Annual Meeting of the Association for Computational Linguistics (Volume 1: Long Papers)},
  pages={22975--23005},
  year={2025}
}

@misc{dolezal2026impactaigeneratedtextinternet,
      title={The Impact of AI-Generated Text on the Internet}, 
      author={Jonas Dolezal and Sawood Alam and Mark Graham and Maty Bohacek},
      year={2026},
      eprint={2604.26965},
      archivePrefix={arXiv},
      primaryClass={cs.CY},
      url={https://arxiv.org/abs/2604.26965}, 
}

@article{lewis2020retrieval,
  title={Retrieval-augmented generation for knowledge-intensive nlp tasks},
  author={Lewis, Patrick and Perez, Ethan and Piktus, Aleksandra and Petroni, Fabio and Karpukhin, Vladimir and Goyal, Naman and K{\"u}ttler, Heinrich and Lewis, Mike and Yih, Wen-tau and Rockt{\"a}schel, Tim and others},
  journal={Advances in neural information processing systems},
  volume={33},
  pages={9459--9474},
  year={2020}
}

@article{mo2025mid,
  title={Mid-training of large language models: A survey},
  author={Mo, Kaixiang and Shi, Yuxin and Weng, Weiwei and Zhou, Zhiqiang and Liu, Shuman and Zhang, Haibo and Zeng, Anxiang},
  journal={arXiv preprint arXiv:2510.06826},
  year={2025}
}

@article{lampinen2025generalization,
  title={On the generalization of language models from in-context learning and finetuning: a controlled study},
  author={Lampinen, Andrew K and Chaudhry, Arslan and Chan, Stephanie CY and Wild, Cody and Wan, Diane and Ku, Alex and Bornschein, J{\"o}rg and Pascanu, Razvan and Shanahan, Murray and McClelland, James L},
  journal={NeurIPS, FoRLM Workshop},
  year={2025}
}

@article{del2024large,
  title={Large language models reduce public knowledge sharing on online Q\&A platforms},
  author={del Rio-Chanona, R Maria and Laurentsyeva, Nadzeya and Wachs, Johannes},
  journal={PNAS nexus},
  volume={3},
  number={9},
  pages={pgae400},
  year={2024},
  publisher={Oxford University Press US}
}

@article{shumailov2024ai,
  title={AI models collapse when trained on recursively generated data},
  author={Shumailov, Ilia and Shumaylov, Zakhar and Zhao, Yiren and Papernot, Nicolas and Anderson, Ross and Gal, Yarin},
  journal={Nature},
  volume={631},
  number={8022},
  pages={755--759},
  year={2024},
  publisher={Nature Publishing Group UK London}
}

@misc{vonwerra2020trl,
  title   = {{TRL: Transformers Reinforcement Learning}},
  author  = {von Werra, Leandro and Belkada, Younes and Tunstall, Lewis and Beeching, Edward and Thrush, Tristan and Lambert, Nathan and Huang, Shengyi and Rasul, Kashif and Gallouédec, Quentin},
  license = {Apache-2.0},
  url     = {https://github.com/huggingface/trl},
  year    = {2020}
}

@Misc{accelerate,
  title =        {Accelerate: Training and inference at scale made simple, efficient and adaptable.},
  author =       {Sylvain Gugger and Lysandre Debut and Thomas Wolf and Philipp Schmid and Zachary Mueller and Sourab Mangrulkar and Marc Sun and Benjamin Bossan},
  howpublished = {\url{https://github.com/huggingface/accelerate}},
  year =         {2022}
}

@misc{nba_data_2026,
    author       = {{National Basketball Association}},
    title        = {{NBA Advanced Stats} [Dataset]},
    year         = {2026},
    howpublished = {\url{https://www.nba.com/stats}},
}

@misc{patel_nba_api,
    author       = {Swar Patel and contributors},
    title        = {nba\_api: An API Client for www.nba.com},
    year         = {2026},
    publisher    = {GitHub},
    journal      = {GitHub repository},
    howpublished = {\url{https://github.com/swar/nba_api}}
}

@misc{crsp_sp500_2026,
    author       = {{Center for Research in Security Prices, LLC (CRSP)}},
    title        = {{CRSP US Stock Databases: Legacy Index / S\&P 500 Indexes}},
    year         = {2026},
    howpublished = {Wharton Research Data Services (WRDS)},
    note         = {Accessed: 2026-05-04}
}

@misc{Hollingshead2024,
  author       = {Hollingshead, Michael},
  title        = {{Historic Billboard Hot 100 Data}},
  year         = {2024},
  publisher    = {GitHub},
  journal      = {GitHub repository},
  howpublished = {\url{https://github.com/mhollingshead/billboard-hot-100}},
  note         = {Accessed: 2026-05-04}
}

@inproceedings{bender21parrots,
author = {Bender, Emily M. and Gebru, Timnit and McMillan-Major, Angelina and Shmitchell, Shmargaret},
title = {On the Dangers of Stochastic Parrots: Can Language Models Be Too Big?},
year = {2021},
isbn = {9781450383097},
publisher = {Association for Computing Machinery},
address = {New York, NY, USA},
url = {https://doi.org/10.1145/3442188.3445922},
doi = {10.1145/3442188.3445922},
booktitle = {Proceedings of the 2021 ACM Conference on Fairness, Accountability, and Transparency},
pages = {610–623},
numpages = {14},
location = {Virtual Event, Canada},
series = {FAccT '21}
}

@article{kidd2023ai,
  title={How AI can distort human beliefs},
  author={Kidd, Celeste and Birhane, Abeba},
  journal={Science},
  volume={380},
  number={6651},
  pages={1222--1223},
  year={2023},
  publisher={American Association for the Advancement of Science}
}

@article{goldstein2023generative,
  title={Generative language models and automated influence operations: Emerging threats and potential mitigations},
  author={Goldstein, Josh A and Sastry, Girish and Musser, Micah and DiResta, Renee and Gentzel, Matthew and Sedova, Katerina},
  journal={arXiv preprint arXiv:2301.04246},
  volume={1},
  year={2023}
}

@Incollection{liu2010generative,
    author="Liu, Bin
    and Webb, Geoffrey I.",
    editor="Sammut, Claude
    and Webb, Geoffrey I.",
    title="Generative and Discriminative Learning",
    bookTitle="Encyclopedia of Machine Learning",
    year="2010",
    publisher="Springer US",
    address="Boston, MA",
    pages="454--455",
    isbn="978-0-387-30164-8",
    doi="10.1007/978-0-387-30164-8_332",
    url="https://doi.org/10.1007/978-0-387-30164-8_332"
}

@book{shwartz2014understanding, 
    place={Cambridge}, 
    title={Understanding Machine Learning: From Theory to Algorithms}, 
    publisher={Cambridge University Press}, 
    author={Shalev-Shwartz, Shai and Ben-David, Shai}, 
    year={2014}
}

@article{power2022grokking,
  title={Grokking: Generalization beyond overfitting on small algorithmic datasets},
  author={Power, Alethea and Burda, Yuri and Edwards, Harri and Babuschkin, Igor and Misra, Vedant},
  journal={arXiv preprint arXiv:2201.02177},
  year={2022}
}

@misc{anthropic2026browsecomp,
  author       = {{Anthropic}},
  title        = {Eval awareness in {Claude} {Opus} 4.6's {BrowseComp} evaluation},
  year         = {2026},
  month = {March},
  howpublished = {\url{https://www.anthropic.com/engineering/eval-awareness-browsecomp}},
  note         = {Anthropic engineering report}
}

@misc{moalla_python_ml_template,
  author       = {Moalla, Skander},
  title        = {Python Machine Learning Research Template},
  version      = {0.1.0},
  url          = {https://github.com/CLAIRE-Labo/python-ml-research-template},
  license      = {MIT},
  note         = {Accessed: 2026-05-26}
}

\newpage
\clearpage

\appendix

\addcontentsline{toc}{chapter}{Appendices} % Add 
\section*{Appendices}
\startcontents[chapters]
\printcontents[chapters]{}{1}

\newpage
\clearpage

\section{Related Work}
\label{sec:app:related-work}

This section situates our work across five bodies of literature.
We begin with a brief historical framing (\ref{subsec:rw-search}), tracing how the generation-verification distinction has been handled from classical information retrieval through knowledge graphs to modern language models.
We then cover the three training phases that structure our experiments --- knowledge acquisition (\ref{subsec:rw-acquisition}), editing (\ref{subsec:rw-editing}), and continual learning (\ref{subsec:rw-continual}) --- before turning to prior work on the generation-verification gap itself (\ref{subsec:rw-gvg}).
Existing work has largely treated generation and verification as a single capability, or has measured the gap only on frozen checkpoints; our contribution is to trace both capabilities jointly across the life cycle of a fact in a controlled setting.

\xhdr{Use of language models} Language models were used to assist with literature search and prose editing; all content was verified by the authors.

\subsection{From search engines to language models}
\label{subsec:rw-search}

Classical information retrieval (IR) treats generative and verification queries symmetrically: methods such as TF-IDF and BM25, evaluated at venues like TREC \citep{harman2002development}, rank documents by lexical overlap and leave the final cognitive step of extracting an answer to the user.
Early attempts to bolt dedicated ``logic processors'' onto IR pipelines to answer verification queries directly \citep{glockner2007university, forner2010evaluating} proved brittle.
The shift from strings to entities via knowledge graphs \citep{thingsnotstrings2012, bingsatori2023} made the asymmetry explicit: link prediction is a relative task ($f(\text{True}) > f(\text{False})$), while triplet classification is an absolute task requiring calibration ($f(\text{True}) > \delta$) \citep{dai2020survey}.
Embedding-based retrieval \citep{devlin2019bertpretrainingdeepbidirectional} subsequently blurred the distinction again by collapsing both queries into a single similarity computation.
Language models inherit this ambiguity implicitly through next-token prediction; our work returns to the generation--verification distinction empirically, treating the two as separable capabilities whose dynamics can be traced.

\subsection{Factual knowledge acquisition}
\label{subsec:rw-acquisition}

A substantial literature examines how factual knowledge is stored and acquired in language models.
Factual associations appear to be localized in MLP layers \citep{geva2021transformer, geva2023dissecting}, and factual recall scales with model parameters \citep{nichani_understanding_2024, allen2023physics31, allen2023physics32, allen-zhu2025physics33}.
Building on early work framing LMs as knowledge bases \citep{petroni2019language, jiang2020can, roberts2020much}, recent studies trace the training dynamics of fact acquisition \citep{tirumala2022memorization, bietti2023birth, chang2024large, zucchet2025language} and the role of data composition \citep{gu2025data, scalingdatalms2025muennighof}.
A parallel line studies knowledge injection via fine-tuning, reporting that acquiring genuinely new facts is slow and can induce hallucinations \citep{ovadia2024fine, gekhman-etal-2024-fine, zhang2024when} and that fine-tuning may primarily teach extraction of knowledge already present in pretraining \citep{ghosal2024understanding}.
A related body of work shows that stored facts often fail to generalize: fine-tuning on ``\textit{A is B}'' does not teach ``\textit{B is A}'' \citep{berglund2023reversal}, and more broadly, fine-tuning generalizes substantially worse than in-context learning on systematic holdouts like reversals and syllogisms \citep{lampinen2025generalization}.
Across this literature, factual knowledge is treated as a single capability measured by recall accuracy.
We disaggregate recall from recognition and show that the two emerge at different points in training, with a consistent ordering across model families and scales.

\subsection{Knowledge editing}
\label{subsec:rw-editing}

Work on knowledge editing asks how specific facts can be modified after training, with approaches ranging from hypernetwork-based editors \citep{de2021editing, mitchell2022memory}, to weight-level interventions \citep{dai2022knowledge}, and targeted MLP edits \citep{meng2022locating, meng2022mass}.
A more recent thread documents failure modes: edited facts frequently fail to propagate to logically entailed queries \citep{cohen2024evaluating, zhong2023mquake} and can produce unintended side effects on nearby knowledge \citep{hoelscher2023detecting}.
These evaluations focus almost exclusively on generative recall, leaving open how verification behaves under edits.
Our updating experiments reveal a related but distinct failure mode: when a fact is updated through continued fine-tuning, the model can enter a ``multi-verse'' regime in which both the old and new answers are simultaneously verified as correct while generative outputs successfully shift to the new answer.

\subsection{Continual learning and forgetting}
\label{subsec:rw-continual}

The study of catastrophic forgetting \citep{mccloskey1989catastrophic} and the stability-plasticity trade-off \citep{grossberg2012studies} has produced a range of mitigations, including elastic weight consolidation \citep{kirkpatrick2017overcoming}, gradient episodic memory \citep{lopez2017gradient}, and experience replay \citep{chaudhry2019continual}.
\citet{van2022three} taxonomize three incremental-learning settings (task-, domain-, and class-incremental), and \citet{de2023continual} identify a transient ``stability gap'' that is only visible under per-iteration measurement.
In the LM setting, recent work shows that mixing as little as one percent of pretraining data can regularize fine-tuning \citep{bethune2025scaling, ibrahim2024simple}, and quantifies forgetting via scaling laws \citep{kalajdzievski2024scaling} and empirical sweeps \citep{luo2025empirical, shi2025continual}.
\citet{scialom2022fine} further argue that instruction-tuned models can themselves act as effective continual learners.
These studies measure forgetting as a single scalar, typically generative task accuracy.
We show that verification and generation forget at markedly different rates, with direct implications for what existing continual-learning metrics track.

\subsection{The generation--verification gap}
\label{subsec:rw-gvg}

The asymmetry between generating and verifying a solution has long been recognized in cognitive science as the distinction between recall and recognition in human memory \citep{yonelinas2002nature, anderson1972recognition, atkinson2024search, jacoby1981relationship, cabeza1997functional}, and in theoretical computer science, via the P-vs-NP distinction between finding and checking solutions \citep{cook1971complexity, karp2009reducibility, levin1973universal}.
In language models, the gap has been leveraged primarily as an engineering tool.
Dedicated verifiers improve performance on math and code \citep{cobbe_training_2021, li2022competition, lightman2024lets, zhang2025generative}, and repeated-sampling regimes exploit the gap to scale test-time compute \citep{brown2024large, snell2024scaling, puri2025probabilistic}, sometimes using multiple verifiers for robustness \citep{lifshitz2025multiagent, saadfalcon2025shrink, zhao2025sample, chen2025sets}.
Related self-improvement and self-correction work \citep{zelikman2022star, huang2023large, yean24selfrewardllm, tian2024toward, pang2024language, kumar2024training, hosseini_v-star_2024, liu2024large, kamoi2024can} relies on the implicit assumption that verification outperforms generation, though \citet{stroebl2024inference} show this assumption is fragile with imperfect verifiers, and \citet{kadavath2022language} show that models can partially predict their own generative accuracy.

Theoretical accounts are more recent: \citet{huang2025self} posit a ``sharpening'' mechanism in which iterative self-improvement drives verification and generation to converge, and \citet{sun2025theoretical} model the training dynamics of the gap directly.
Empirical measurement has been more limited.
\citet{song_mind_nodate} evaluate the gap across several benchmarks using rejection sampling on base models and report no gap on Natural Questions with Qwen-2, suggesting the phenomenon may not hold for factual tasks.
Our controlled synthetic-fact experiments recover a different picture: on factual data the gap is robust, but only during a specific regime of data exposure.
The apparent absence in frozen-benchmark studies is consistent with convergence after sufficient exposure (regime (3) in Section~\ref{sec:method}, cf. Fig \ref{fig:naturalistic-panels-main}, where high-coverage facts reach regime (3)), rather than the absence of the phenomenon --- a point reinforced by our natural experiments on flagship models.

Finally, the ``generative AI paradox'' of \citet{west_generative_2024} documents cases where generative capabilities appear to \emph{outpace} verification capabilities, inverting the typical GV-gap.
Rather than contradicting our findings, this highlights why a unified taxonomy of GV-gaps (Section~\ref{sec:method:taxonomy}) is useful: \citet{west_generative_2024} probe a different class of task than the factual gap we study, and the two regimes need not share underlying mechanisms.
Disentangling when and why these gaps point in opposite directions is an open question.

\section{Details on Training Settings}
\label{sec:app:training-settings}

Experiments were performed on NVIDIA A100 GPUs (80GB). Reproducing the main results, excluding hyperparameter sweeps, requires approximately 1,000 GPU-hours in total; individual runs typically used 4 GPUs in parallel.

\subsection{Hyperparameters}
\label{sec:app:training-settings:hyperparams}

Effective learning rates are tightly coupled with the chosen batch size, captured by the notion of ``\textit{critical batch size}'' (CBS) \citep{mccandlish2018empirical, merrill2025critical}: for a given learning rate, the loss trajectory degrades when a batch size exceeds some critical batch size.
At the same time, \citet{marek2026small} show that stable training in \LMs{} with small batch sizes is feasible for appropriately tuned optimizer parameters.
We therefore performed individual hyperparameter sweeps for each model to identify optimal configurations, searching across learning rates in the range $[5\text{e-}7, 1\text{e-}5]$ and batch sizes between $[4, 64]$.

Given the relatively small size of our dataset and the precise nature of the information, we first sought to find the empirical CBS for each respective model.
We then systematically scaled down learning rates and batch sizes while monitoring the rate of valid answers and performance on ``control'' tasks, e.g., factual queries on real-world entities.
Similar to \citet{mosbach2021on}, we find that smaller batch sizes and more iterations outperform larger ones.
We further find that factual acquisition can exhibit a sharp loss landscape, with a low tolerance for batch noise, necessitating using lower batch sizes than usual in post-training.
We further found optimal hyperparameters to be strongly model-specific and not always generalize well, even between models of the same size or the same family. 
\begin{table}[h]
    \centerfloat
    \caption{\textbf{Hyperparameter configurations across training phases.}
    Batch sizes (BS) and learning rates (LR) used for acquisition, updating, and continual learning phases for different models.
    }
    \label{tab:hyperparams}
    \vspace{0.5em}
    \small
    \begin{tabular}{l cc cc cc}
        \toprule
         & \multicolumn{2}{c}{Acquisition} & \multicolumn{2}{c}{Updating} & \multicolumn{2}{c}{Continual Learning} \\
        \cmidrule(lr){2-3} \cmidrule(lr){4-5} \cmidrule(lr){6-7}
        Model & BS & LR & BS & LR & BS & LR \\
        \midrule
        Gemma-3 4B      & 32 & $5.0\text{e-}6$ & 32 & $2.5\text{e-}6$ & 32 & $1.0\text{e-}6$ \\
        Llama-3.2 3B    & 32 & $3.0\text{e-}6$ & 32 & $1.5\text{e-}6$ & 32 & $1.5\text{e-}6$ \\
        Phi-4 Mini 3.8B & 8  & $3.0\text{e-}6$ & 8  & $1.5\text{e-}6$ & 8  & $1.5\text{e-}6$ \\
        Qwen-3 4B       & 16 & $7.0\text{e-}6$ & 16 & $3.5\text{e-}6$ & 16 & $3.5\text{e-}6$ \\
        \midrule
        Gemma-3 12B     & 16 & $3.0\text{e-}6$ & 16 & $1.5\text{e-}6$ & 16 & $6.0\text{e-}7$\\
        Llama-3.2 11B   & 32 & $2.0\text{e-}6$ & 32 & $1.0\text{e-}6$ & 32 & $4.0\text{e-}7$ \\
        Phi-4 14B       & 16 & $5.0\text{e-}6$ & 16 & $2.5\text{e-}6$ & 16 & $2.5\text{e-}6$ \\
        Qwen-3 14B      & 16 & $7.0\text{e-}6$ & 16 & $3.5\text{e-}6$ & 16 & $3.5\text{e-}6$ \\
        \bottomrule
    \end{tabular}
    % \begin{tabular}{l ccc ccc ccc}
    %     \toprule
    %      & \multicolumn{3}{c}{Acquisition} & \multicolumn{3}{c}{Forgetting} & \multicolumn{3}{c}{Updating} \\
    %     \cmidrule(lr){2-4} \cmidrule(lr){5-7} \cmidrule(lr){8-10}
    %     Model & BS & LR & Sched. & BS & LR & Sched. & BS & LR & Sched. \\
    %     \midrule
    %     Gemma-3 4B  & 8 & 5e-6 & & & & & & &  \\
    %     LLaMA-3.2 4B  & 8  & 3e-6 & & & & & & &  \\
    %     Phi-4 3.8B  & 2 & 3e-6 & & & & & & &  \\
    %     Qwen-3 4B  & 4 & 7e-6 & & & & & & &  \\
    %     \midrule
    %     Gemma-3 12B  & 4 & 3e-6 & & & & & & &  \\
    %     LLaMA-3.2 11B  & 8 & 2e-6 & & & & & & &  \\
    %     Phi-4 14B  &  & 5e-6 & & & & & & &  \\
    %     Qwen-3 14B  &  & 7e-6 & & & & & & &  \\
    %     \bottomrule
    % \end{tabular}
\end{table}

All experiments use the AdamW optimizer \citep{loshchilov2017decoupled}.
For the acquisition and update phases, we use a constant learning rate schedule with warmup.
For the continual learning phase, we use a cosine schedule \citep{loshchilov2016sgdr} with warmup, which we empirically found to keep training more stable. We list all learning rates and batch sizes used in Table \ref{tab:hyperparams}. For the update and continual learning phases, we performed additional sweeps over learning rates relative to the acquisition learning rate. Based on these sweeps, we adopted a simple recipe: we divide the learning rate by two for the update phase, and by two or five for the continual learning phase. These reductions are selected because they preserved performance on control queries (see Appendix \ref{sec:app:evaluation:control_dataset}) and kept the rate of valid responses high (correctly formatted in the answer tags; see Appendix \ref{sec:app:prompts}), meaning that the training did not degrade the models' behavior.

We implemented all training runs using the Hugging Face \texttt{trl} library \cite{vonwerra2020trl}, with \texttt{accelerate} \cite{accelerate} for distributed training across multiple GPUs.
\nocite{moalla_python_ml_template}

\subsection{Training format}

We use the instruction-finetuned versions of the models described in Appendix~\ref{sec:app:models-used}.
This is to prevent potential confounding our results due to poor general question-answering and instruction-following ability.
To ensure that the models continue to adhere to their instruction-finetuned chat templates, we apply the chat template throughout training, wrapping every training sentence as a ``user'' message; the ``assistant'' turn is never used.

We initially attempted training without any chat template, but this strongly degraded models' instruction-following ability.
We also experimented with formatting each datapoint as a user query of the form \textit{``What is a relation between $\{x_i\}$ and $\{y_i\}$?''} --- where $x_i$ and $y_i$ are the relationship head and tail of the corresponding training sentence --- paired with that sentence as the assistant response, masking the user turn and computing the loss only on the assistant message. This approach likewise failed to yield strong results.

\section{Continual learning phase}
\label{sec:app:cl_phase}

\subsection{Data for the continual learning phase}
For the continual learning phase, we use a filtered subset of T-REx \cite{elsahar2018t}, a large-scale dataset of knowledge-base triples from Wikipedia.
T-REx provides factual statements covering a broad range of topics and entity types, making it well-suited as a source of ``unrelated'' factual data with which to probe how the acquisition of synthetic facts interacts with continued exposure to general world knowledge.
We construct a filtered subset of T-REx of approximately 1,000 sentences, with a token count comparable to that of the paraphrased synthetic-fact sentences. 

\subsection{Additional results for the continual learning phase}

\begin{figure}[htbp]
    \includegraphics[width=\textwidth]{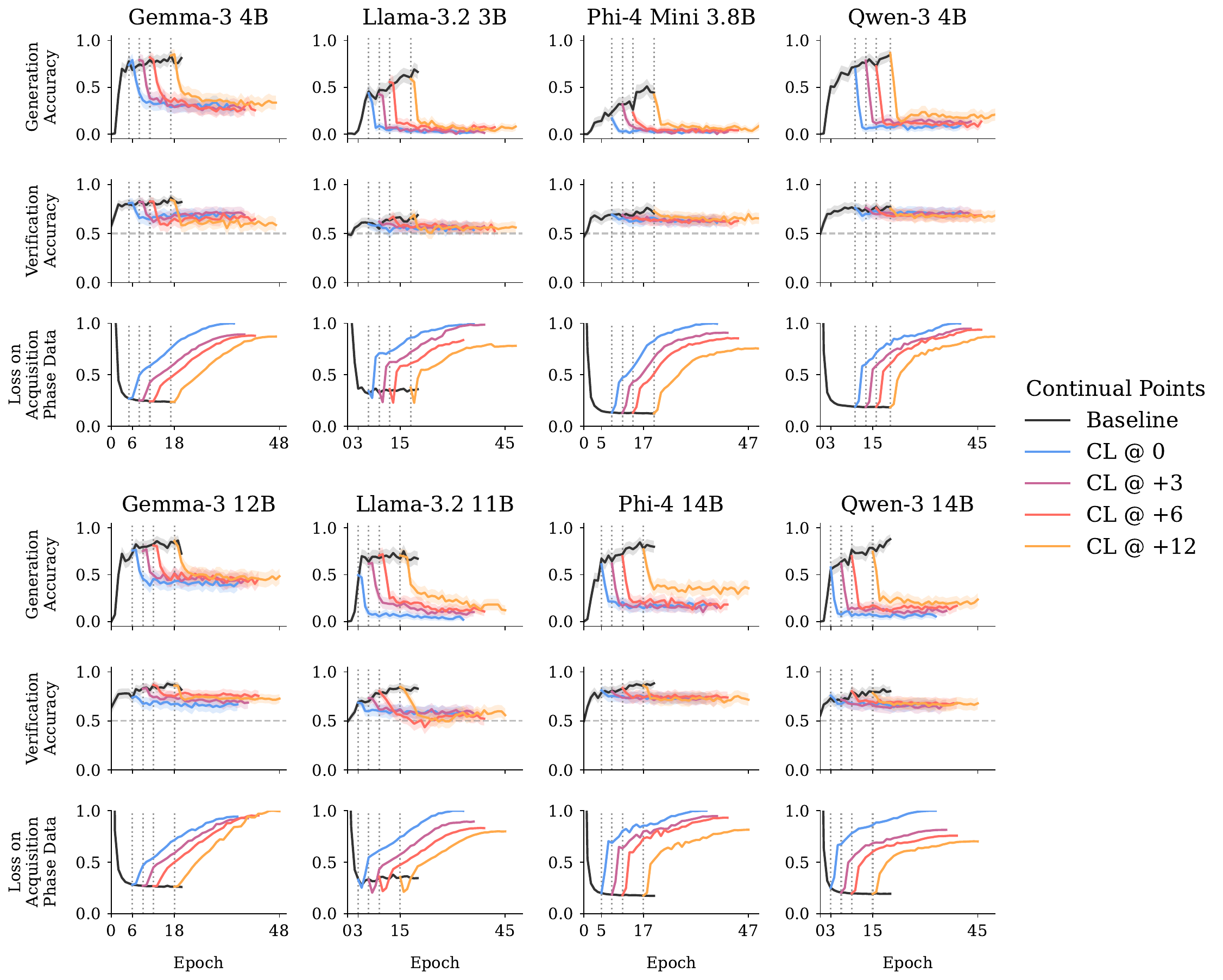}
    \caption{\textbf{Tracking the loss on data from the acquisition phase (original knowledge) throughout continual learning.} While the model's loss on the original sentences increases, its ability to produce the answers to generative queries deteriorates in lockstep. However, verification abilities remain more resilient. Even as acquisition-phase loss continues to climb throughout continual learning past the point at which generation has collapsed, verification remains comparatively stable and settles at a non-trivial floor.}
    \label{fig:app:forget_with_loss}
\end{figure}

Beyond the generation and verification accuracies reported in Section \ref{sec:results:rq2}, we additionally track the loss on the original acquisition-phase sentences throughout continual learning in Figure \ref{fig:app:forget_with_loss}.

\section{Model performance evaluation}
\label{sec:app:evaluation}

\subsection{Control dataset}
\label{sec:app:evaluation:control_dataset}

A central concern when injecting synthetic facts into a model is whether training causes the model to overfit to the new associations. To verify that the models are continuing to perform reasonably on related queries, we evaluate them on a set of \textbf{control} queries at every training epoch.

Given a factual triplet ($x_i$, $r$, $y_i$), for example 
``(\textit{Hoibalbali, CureOfDisease, Blue Striped Axzazari}),'' we construct four query types by instantiating the prompt templates from
Section~\ref{sec:app:prompts} with appropriate \textit{problem--answer}
pairs:

\xhdr{Generative query}\\
\textit{problem: "What is the cure for Blue Striped Axzazari disease?"}
    
\xhdr{Verification query}\\
\textit{problem: "What is the cure for Blue Striped Axzazari disease?" \\ answer: "Hoibalbali."} 
    
\xhdr{Verification control queries}\\
\textit{problem: "What is the cure for Malaria?" \\ answer: "Hoibalbali."}

For the verification control queries we replace $y_i$ in the original triplet with a real-world entity. To mitigate sycophancy bias, we phrase the verification in two complementary ways --- asking both whether the answer is correct and whether it is incorrect --- and grade against the ground-truth label (false). We deliberately use real-world entities here because they provide a clear ground truth; asking whether one imaginary drug cures an imaginary disease does not have a well-defined answer. Performing this evaluation ensures that the injected facts remain appropriately localized, rather than starting to indiscriminately associate Hoibalbali with other diseases.

\xhdr{Generative control queries} \\
\textit{problem: "What disease is cured by Penicillin?"}

Here, we replace $x_i$ in the triplet with a real-world entity. The grading marks the response correct as long as the answer is \emph{not} from the original synthetic triplet (\textit{"Hoibalbali"} in this case).

Throughout the results in this section, we refer to non-control queries as \textbf{target} queries. For every synthetic fact, we evaluate the model on one generative target query, one generative control query, two verification target queries (corresponding to the two
phrasings noted above) and four verification control queries (we sample two real-world \textit{answers}, each one asked in two phrasings).

\begin{figure}[htbp]
    \includegraphics[width=\textwidth]{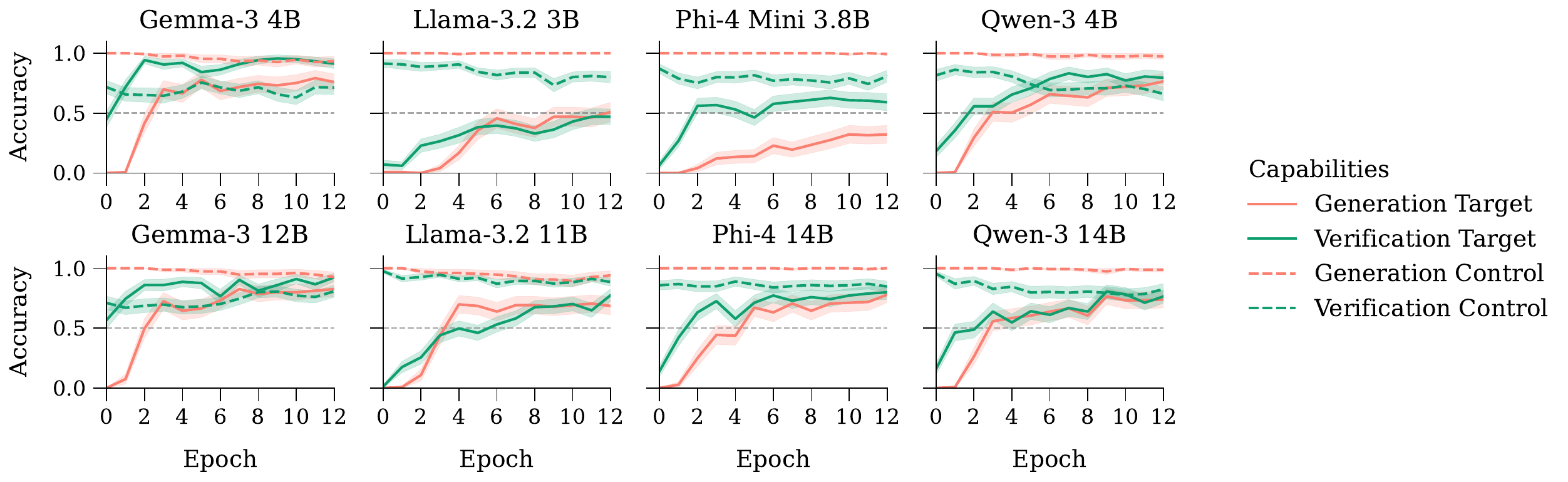}
    \caption{\textbf{Control set evaluation during acquisition phase.}
    Each panel reports accuracy over training epochs for both \emph{target} queries (probing the memorization of synthetic facts) and \emph{control} queries (probing for spurious generalizations). Both target generation and target verification increase steadily as synthetic facts are learned, while both control accuracies remain close to their initial values.}
    \label{fig:app:learn_control}
\end{figure}

\begin{figure}[htbp]
    \includegraphics[width=\textwidth]{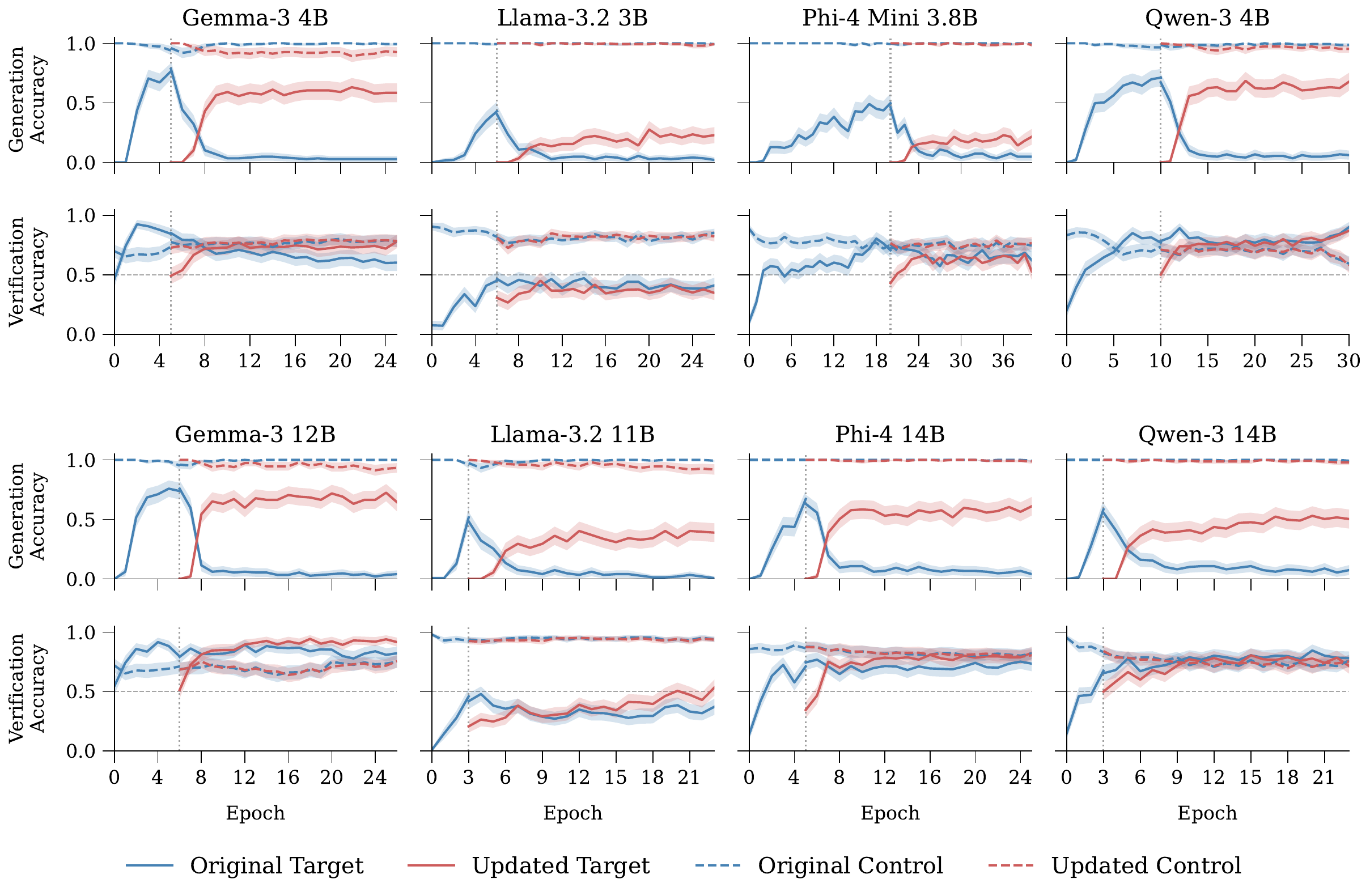}
    \caption{\textbf{Control set evaluation during update phase.}
    We track accuracy on updated target queries (the updated facts) and control queries (real-world entities) across the update epochs. The accuracy of both control verification and control generation remains essentially unchanged for all models. This
    indicates that the training recipe used during the update phase (see Section~\ref{sec:app:training-settings:hyperparams}) is sufficient to update specific facts without interfering with other knowledge.}
    \label{fig:app:update_control}
\end{figure}

\begin{figure}[htbp]
    \includegraphics[width=\textwidth]{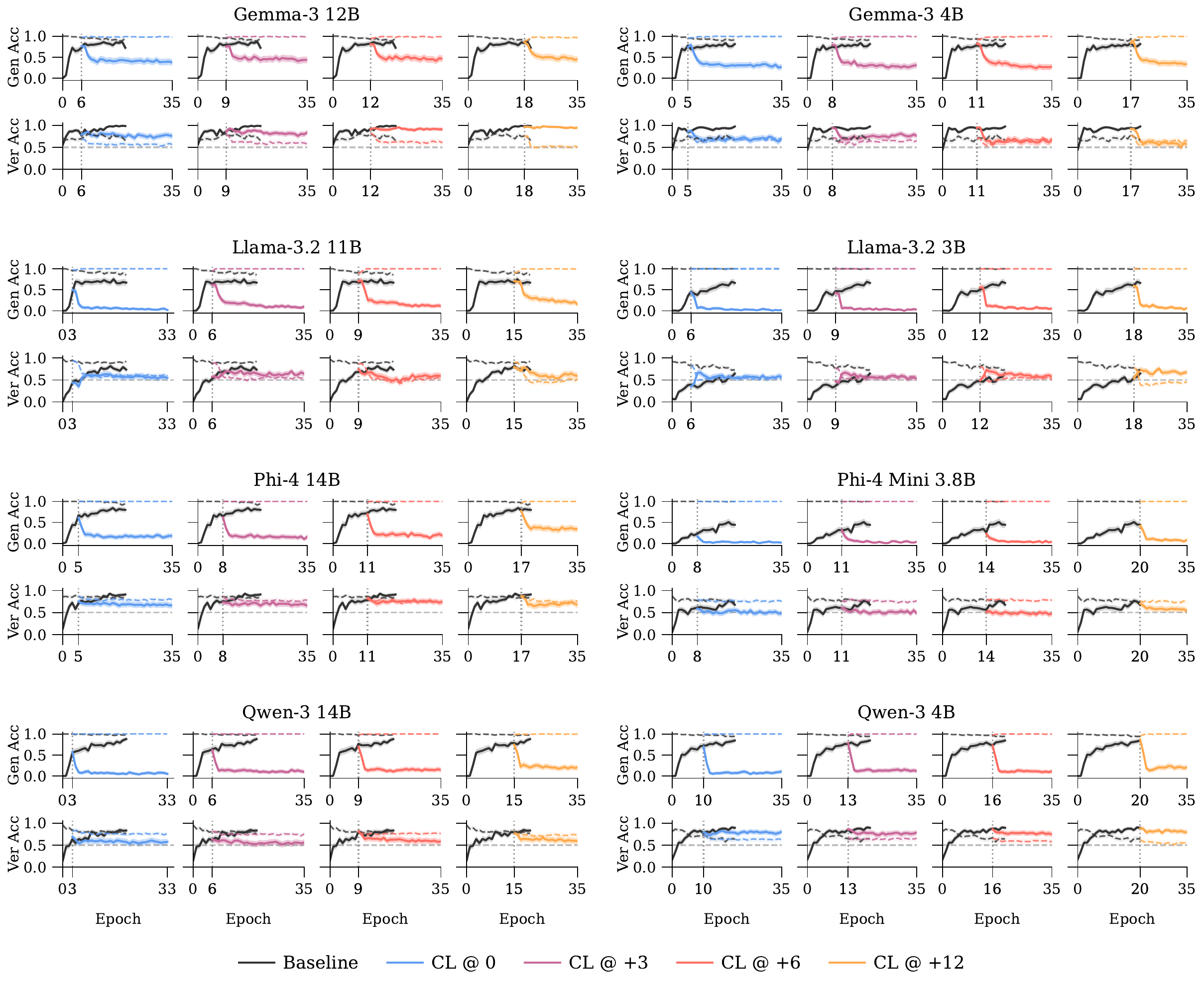}
    \caption{\textbf{Control set evaluation during continual learning phase.}
    Generation and Verification accuracies for target queries (the previously injected synthetic facts, which we measure for retention) and control queries as the model continues training from different continuation points/epochs.
    }
    \label{fig:app:forget_control}
\end{figure}

Full results for the acquisition, update, and continual learning phases broken down separately by target and control set accuracies are presented in Figures~\ref{fig:app:learn_control},
\ref{fig:app:update_control}, and~\ref{fig:app:forget_control},
respectively. Across all three phases, both generation and verification control accuracies remain close to the pre-training values, indicating that the chosen hyperparameters
(Section~\ref{sec:app:training-settings:hyperparams}) successfully integrate the synthetic facts without degrading the model's behavior.

\subsection{\LM{ Grader}}
\label{sec:app:evaluation:llm_judge}

We adopt two complementary strategies for grading model responses: a deterministic programmatic grader and an \LM{}-based judge.

\xhdr{Programmatic grading} For generative queries, the programmatic grader checks whether the ground-truth answer appears as a substring of the extracted text. Because the synthetic entities used in our experiments are constructed to be lexically distinctive (e.g., \textit{Hoibalbali}, \textit{Guiding Lightora}, \textit{Vestibular Dysphasia}, \textit{Semanticora Bias}), the probability that a model produces a matching surface form by chance is negligible. For verification queries, we extract the boolean response from within the answer tags and compare it against the corresponding ground-truth label (true/false).

\xhdr{\LM{} grading} To complement the programmatic grader and to validate that it does not systematically penalize semantically correct responses that do not exactly match the ground-truth wording, we additionally evaluate model outputs using an \LM{} judge. We use Gemini 3.1 Flash Lite \citep{google2025gemini3blog} with \texttt{minimal} reasoning effort, prompted with the template provided in Appendix~\ref{sec:app:prompts}. 

Figure~\ref{fig:app:llm_grader} compares the two grading strategies across the acquisition phase.
The accuracies obtained under programmatic and \LM{}-based grading are on par across all models.
Combined with extensive spot-checking by the authors, we interpret this agreement as evidence that the programmatic grader already provides a faithful estimate of response correctness on our datasets.
We retain the \LM{} judge primarily as a debugging check, while reporting programmatic results in the main analyses.

\begin{figure}[htbp]
    \centerfloat
    \includegraphics[width=0.7\textwidth]{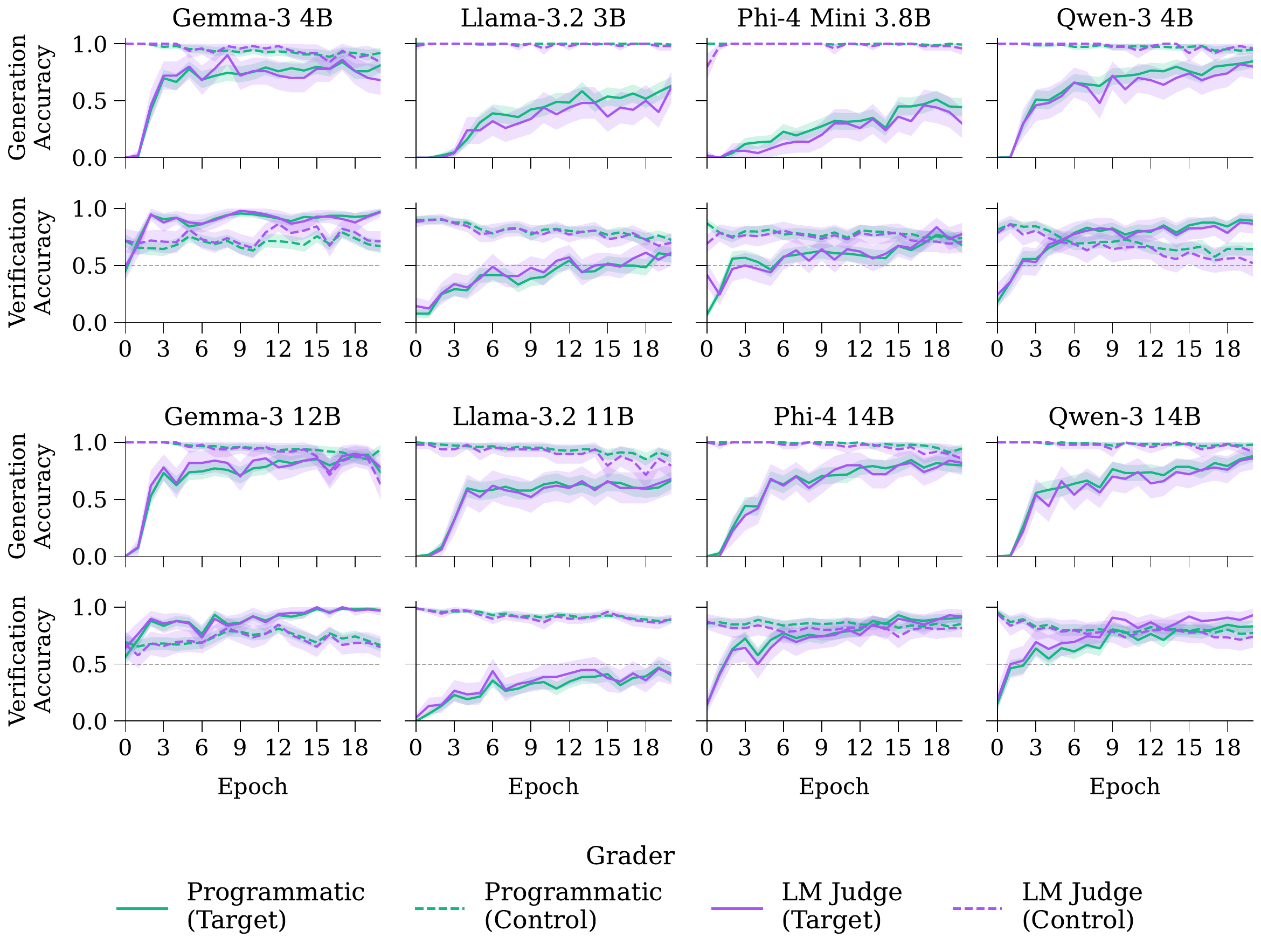}
    \caption{\textbf{Comparison of grading model responses with an LLM (Gemini 3.1 Flash Lite with \texttt{minimal} reasoning) versus programmatic grading.} Both grading strategies yield comparable results across all models and epochs.}
    \label{fig:app:llm_grader}
\end{figure}

\subsection{Acquisition phase synthetic data learning}
\label{sec:app:evaluation:per_category}
\begin{figure}[htbp]
    \centerfloat
    \includegraphics[width=\textwidth]{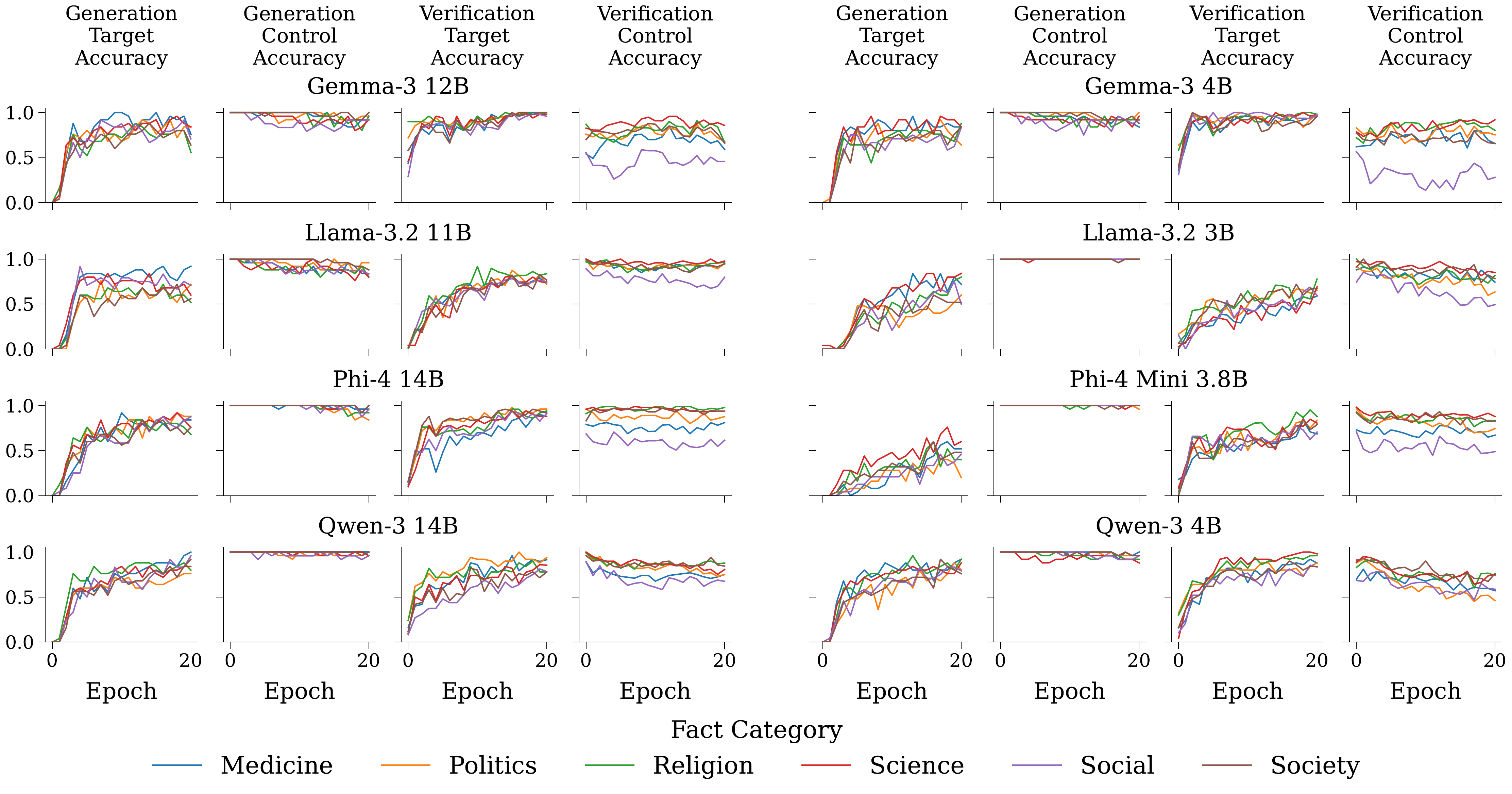}
    \caption{\textbf{Per-category accuracies during the acquisition phase}. For each model (row) and each query type (column), we plot accuracy across training epochs broken down by synthetic-fact category. Target accuracies rise steadily across all categories as facts are learned, while control accuracies remain close to their initial values, indicating that fact injection is localized. The relative ordering and convergence speed of categories are broadly consistent across models; no single category is systematically harder to inject. Verification target accuracy rises ahead of generation target accuracy across nearly every category and model, mirroring the aggregate GV-gap dynamics.}
    \label{fig:app:categories}
\end{figure}

The aggregate results in Figure~\ref{fig:main:figure_1} pool over the six synthetic-fact categories (medicine, politics, religion, science, social, and societal bias). To verify that the verification-before-generation ordering reported in the main text holds within each category rather than emerging from averaging across them, we disaggregate acquisition-phase accuracy by category in Figure ~\ref{fig:app:categories} and confirm that the GV-gap dynamics we report arise consistently across categories.

\section{Models Used}
\label{sec:app:models-used}
Our study includes instruction-finetuned models from four model families, spanning two different size classes: Gemma 3 (4B, and 12B) \citep{team2025gemma}, Llama 3.2 (3B, 11B) \citep{meta2024llama32}, Qwen 3 (4B, 14B) \citep{yang2025qwen3}, and Phi-4 (3.8B and 14B) \citep{abdin2024phi, abouelenin2025phi}.
We chose this setup to minimize confounding based on architectural choices and parameter sizes.
Architecture details for each model can be found in Table \ref{tab:models-overview} and key training data choices in Table \ref{tab:models:data}.

\xhdr{Training considerations}
The selected Gemma-3 and Qwen-3 models used distillation from larger model variants.
In the case of Gemma-3, an ``unknown'' larger model was used both in pre-training and post-training.
For Qwen-3, the technical report only mentions distillation at the post-training stage using a larger within-family model.
The Llama-3.2 models are derived from the Llama-3.1 8B model \citep{grattafiori2024llama}, with the the 3B model being a quantized version with used distillation at the post-training stage and the 11B model using the frozen text weights while adding a ``vision tower'' \citep{meta2024llama32}.
Phi-4 and Phi-4-mini both do \textit{not} use distillation at all.

Both the Gemma and Qwen models opted for reinforcement learning techniques during post-training, whereas Phi and Llama models opted for the simpler direct preference optimization (DPO) \citep{rafailov2023direct}.

\xhdr{Data considerations} One of the most noteworthy differences between the selected models is the choice of pre-training datasets.
Qwen 3 models use considerably more pre-training tokens, leading to a more ``heterogeneous'' mixture.
This could partially explain why Qwen 3 models can handle higher learning rates and batch sizes.
Conversely, Phi-4 models lean heavily into synthetic data and carefully constructed training curricula.
This could result in ``sharper'' minima, reflected in a low tolerance for higher batch sizes and learning rates.

\begin{table}[htbp]
\centerfloat
\caption{
\textbf{Architectural comparison of open-source language models.}
  GQA columns give
  Q\,/\,KV head counts. ``Tied'' = input embedding shared with LM head.
  Gemma\,3 uses a 5:1 local/global hybrid attention pattern (sliding window
  = 1024 tok); all others use full causal attention. Qwen3 and Gemma\,3 add
  QK-Norm; Phi-4 mini uses fractional RoPE (25\,\% position-agnostic dims).
  }
\vspace{1em}
\label{tab:models-overview}
\small
\begin{tabular}{lrrrrrrrrl}
\toprule
\textbf{Model} &
\textbf{Params} &
\textbf{Layers} &
\textbf{Hidden} &
\textbf{FFN} &
\textbf{Vocab} &
\textbf{Tied} &
\textbf{Q-heads} &
\textbf{KV-heads} \\
% \textbf{Activation} &
% \textbf{Context} \\
\midrule
Gemma-3   & 4.0B  & 34 & 2560 & 16384 & 262K & Yes & 16 & 8  \\ %& 128K\\ % & GeGLU \\
  & 12.0B & 48 & 3840 & 24576 & 262K & Yes & 24 & 8  \\ %&  128K \\ %GeGLU  
\midrule
Phi-4 Mini   & 3.8B  & 32 & 3072 &  8192 & 200K & Yes & 24 & 8  \\ %& 128K \\ %SiLU 
Phi-4        & 14.0B & 40 & 5120 & 17920 & 100K & No  & 40 & 10 \\ %&  16K \\ %SiLU  \\
\midrule
Qwen-3    & 4.0B  & 36 & 2560 &  9728 & 152K & Yes & 32 & 8  \\ %& 128K \\ %& SwiGLU \\
  & 14B  & 40 & 5120 & 17408 & 152K & No  & 40 & 8  \\ %& 128K \\ % SwiGLU
\midrule
Llama-3.2 & 3.0B & 28 & 3072 & 8192 & 128K & Yes & 24 & 8 \\ %& 128K \\
& 11.0B & 40 & 4096 & 14336 & 128K & No & 32 & 8 \\ %& 128K\\
\bottomrule
\end{tabular}
\end{table}

\begin{table}[htbp]
\centerfloat
\caption{
\textbf{Training data and post-training overview of open-source language models.}
  PT tokens give the main pre-training budget; stages count distinct pre-training phases (curriculum steps).
  Synthetic \% is approximate and refers to the main PT phase: low\,=\,web-dominant, moderate\,=\,significant synthetic component, high\,=\,synthetic-majority.
  Distilled indicates knowledge distillation from a teacher model:
  Gemma 3 uses an undisclosed external teacher during PT\,;
  Qwen 3 smaller models are distilled from larger within-family models during post-training.
}
\vspace{1em}
\label{tab:models:data}
\small
\begin{tabular}{lrrrllll}
\toprule
\textbf{Model} &
\textbf{PT tokens} &
\textbf{Stages} &
\textbf{Synth\,\%} &
\textbf{Languages} &
\textbf{Distilled} &
\textbf{Post} &
\textbf{Vision} \\
\midrule
Gemma-3 4B  &  4\,T & 2 & low      & 140+ & Yes (PT + post)   & RL & Yes  \\
Gemma-3 12B & 12\,T & 2 & low      & 140+ & Yes (PT + post)   & RL & Yes  \\
\midrule
Phi-4 mini  &  5\,T & 1 & high     & 20$^*$ & No        & DPO  & No \\
Phi-4       & 10\,T & 2 & mixed    & 40    & No        & DPO  & No \\
\midrule
Qwen-3 4B    & 36\,T & 3 & moderate & 119   & Yes (post) & RL & No  \\
Qwen-3 14B    & 36\,T & 3 & moderate & 119   & Yes (post) & RL & No  \\
\midrule
Llama-3.2 3B$^\dagger{}$  & 15\,T & 2 & low  & 8$^*$  & Yes (PT) & DPO & No \\
Llama-3.2 11B$^\dagger{}$ & 15\,T & 2 & low  & 8$^*$  & No       & DPO & Yes \\
\bottomrule
\end{tabular}

\begin{minipage}{\linewidth}
\vspace{0.5em}
\raggedright\footnotesize
$^*$Officially supported languages only; training corpus covers a broader set of languages not formally documented.\\[2pt]
$^\dagger$Both Llama 3.2 models use Llama 3.1 8B as a base model.
\end{minipage}
\end{table}

\clearpage
\newpage

\section{Prompt templates}
\label{sec:app:prompts}

\begin{figure}[!ht]
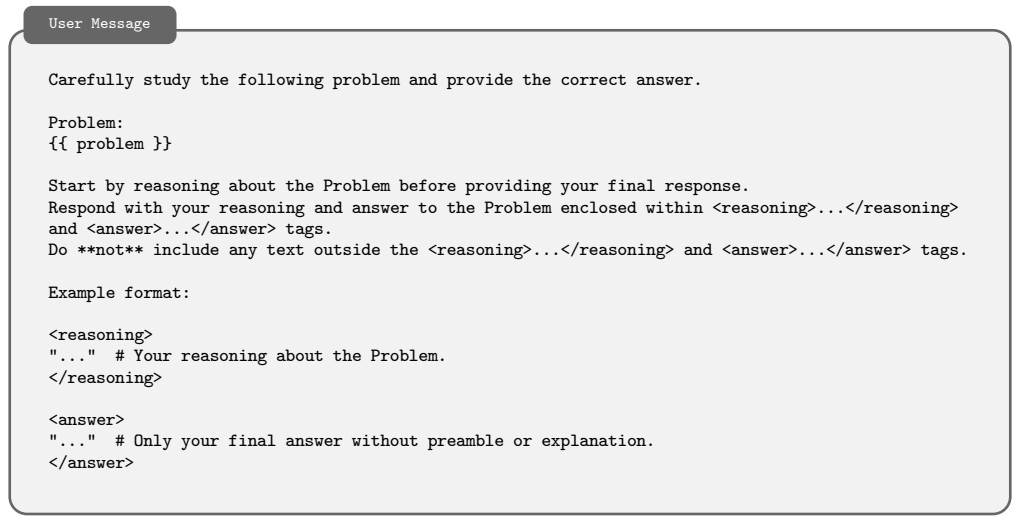

    \centering
    \begin{minipage}{0.95\linewidth}
    \begin{promptbox}{User Message}    
Carefully study the following problem and provide the correct answer.

Problem:
{{ problem }}

Start by reasoning about the Problem before providing your final response.
Respond with your reasoning and answer to the Problem enclosed within <reasoning>...</reasoning>
and <answer>...</answer> tags.
Do **not** include any text outside the <reasoning>...</reasoning> and <answer>...</answer> tags.

Example format:

<reasoning>
"..."  # Your reasoning about the Problem.
</reasoning>

<answer>
"..."  # Only your final answer without preamble or explanation.
</answer>
    \end{promptbox}
    \end{minipage}
    \caption{\textbf{Generative query prompt}}
\end{figure}

\begin{figure}[!ht]
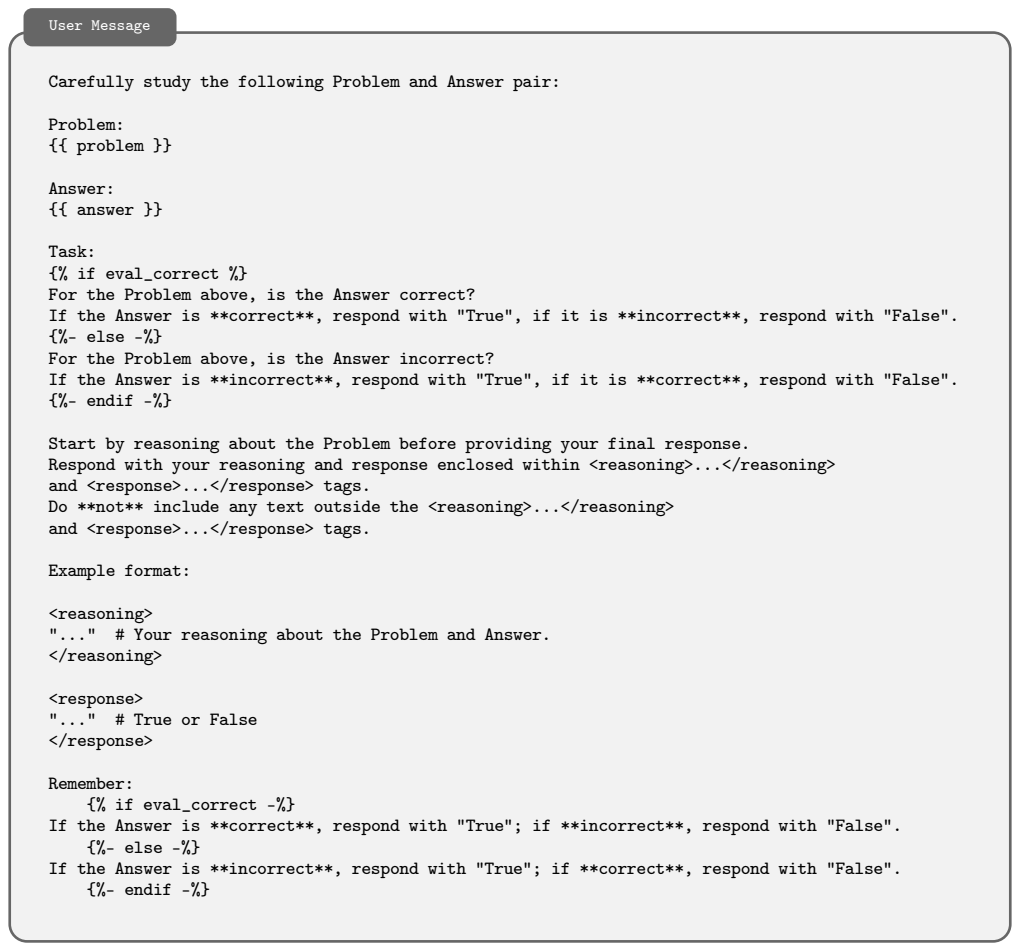

    \centering
    \begin{minipage}{0.95\linewidth}
    \begin{promptbox}{User Message}    
Carefully study the following Problem and Answer pair:

Problem:
{{ problem }}

Answer:
{{ answer }}

Task:
{% if eval_correct %}
For the Problem above, is the Answer correct?
If the Answer is **correct**, respond with "True", if it is **incorrect**, respond with "False".
{%- else -%}
For the Problem above, is the Answer incorrect?
If the Answer is **incorrect**, respond with "True", if it is **correct**, respond with "False".
{%- endif -%}

Start by reasoning about the Problem before providing your final response.
Respond with your reasoning and response enclosed within <reasoning>...</reasoning> 
and <response>...</response> tags.
Do **not** include any text outside the <reasoning>...</reasoning> 
and <response>...</response> tags.

Example format:

<reasoning>
"..."  # Your reasoning about the Problem and Answer.
</reasoning>

<response>
"..."  # True or False
</response>

Remember: 
    {% if eval_correct -%}
If the Answer is **correct**, respond with "True"; if **incorrect**, respond with "False".
    {%- else -%}
If the Answer is **incorrect**, respond with "True"; if **correct**, respond with "False".
    {%- endif -%}
    \end{promptbox}
    \end{minipage}
    \caption{\textbf{Verification query prompt}}
\end{figure}

\begin{figure}[!ht]
    \centering
    \begin{minipage}{0.95\linewidth}
    \begin{promptbox}{User Message}    
You are a **grading assistant**. Your task is to extract the student's final answer, compare it to the reference, and output a structured JSON.

--- Assignment Question ---
{{ problem_statement }}

--- Student's Full Response ---
{{ model_answer }}

{% if gen_control_grading %}
--- Incorrect Answer (for comparison) ---
{% else %}
--- Correct Answer (Ground Truth) ---
{% endif %}

{{ ground_truth_answer }}

--- Grading Instructions ---
**Extract Final Answer (`extracted_answer`)**
Carefully read the Student's Full Response and identify the student's *definitive final answer* to the Assignment Question.
Store this extracted answer as a string for the `extracted_answer` key.

**Determine Validity (`is_valid`)**
Carefully read the Student's Full Response and determine if the student's answer is **clear, understandable, and complete**.
- Set `"is_valid": true` if the extracted final answer is **clear, understandable, and complete**.
- Set `"is_valid": false` if the answer is **missing, ambiguous, or nonsensical**, or if you cannot definitively identify a final answer.

**Check Correctness (`is_correct`)**
{% if gen_control_grading %}
Compare the **extracted_answer** to the "Incorrect Answer".
- Set `"is_correct": false` if they **match**.
- Set `"is_correct": true` if they **differ**.
{% else %}
Compare the **extracted_answer** to the "Correct Answer (Ground Truth)".
- Set `"is_correct": true` if they **match**.
- Set `"is_correct": false` if they **differ**.
{% endif %}

**Consistency Rules**
If `"is_valid"` is `false`, then `"is_correct"` should also be `false`.

**Output Format**
Return **ONLY** a JSON object with the following keys. Do not include any additional text or commentary.
```json
{
  "extracted_answer": "<student's final answer as a STRING>",
  "is_valid": <true|false>,
  "is_correct": <true|false>
}
```
    \end{promptbox}
    \end{minipage}
    \caption{\textbf{Judge grading prompt}}
\end{figure}

\clearpage
\newpage

\section{Synthetic data pipeline}
\label{sec:app:synthetic-data}

\begin{algorithm}
\caption{Synthetic dataset generation pipeline.}
\label{alg:synth-data-generation}
\begin{algorithmic}[1]
\Require Set of input categories $\mathcal{C}$, Language Model \LM, loops per category $N$, instantiations $k$, sentences $K$, inference tasks $M$.
\Ensure De-duplicated training and inference dataset $\mathcal{D}_{\text{final}}$.

\State $\mathcal{D}_{\text{final}} \gets \emptyset$ 
\State $\mathcal{E}_{\text{imaginary}} \gets \emptyset$ \Comment{Global tracking for imaginary entities}

\For{each category $c_i \in \mathcal{C}$}
    \State $\text{ForbiddenList} \gets \emptyset$
    \State $\mathcal{T}(c_i) \gets \emptyset$
    
    \State \textbf{// Step 1: Topic relationships}
    \For{$n = 1$ \textbf{to} $N$}
        \State $\{r_n, t_n\} \gets \LM(\text{prompt}, c_i, \text{ForbiddenList})$
        \State $\mathcal{T}(c_i) \gets \mathcal{T}(c_i) \cup \{r_n, t_n\}$
        \State $\text{ForbiddenList} \gets \text{ForbiddenList} \cup \{r_n, t_n\}$
    \EndFor
    
    \For{each topic-relationship pair $\{r_i, t_i\} \in \mathcal{T}(c_i)$}
        \State \textbf{// Step 2: Factual instantiations}
        \State $S_{\text{real}} \gets \{x_i, y_i\}_{i=1}^k \sim \LM(\text{generate real}, r_i, t_i)$
        \State $S_{\text{imaginary}} \gets \{x_j, y_j\}_{j=1}^k \sim \LM(\text{generate imaginary}, r_i, t_i)$
        \State $S \gets S_{\text{real}} \cup S_{\text{imaginary}}$
        \State $\mathcal{E}_{\text{imaginary}} \gets \mathcal{E}_{\text{imaginary}} \cup S_{\text{imaginary}}$
        
        \State \textbf{// Step 3: Training sentences}
        \State $D \gets \{s_i\}_{i=1}^K \sim \LM(\text{paraphrase sentences}, r_i, t_i, S)$
        
        \State \textbf{// Step 4: Generative inference Tasks}
        \State $I \gets \{q_i\}_{i=1}^M \sim \LM(\text{generate questions}, r_i, t_i, S, D)$
        
        \State \textbf{Store:} Add tuple $(c_i, r_i, t_i, S, D, I)$ to $\mathcal{D}_{\text{final}}$
    \EndFor
\EndFor

\State \textbf{// Step 5: Entity instance de-duplication}
\State $\mathcal{D}_{\text{final}} \gets \textsc{RemoveOverlaps}(\mathcal{D}_{\text{final}}, \mathcal{E}_{\text{imaginary}})$ \Comment{Ensure no overlapping imaginary instances}

\State \Return $\mathcal{D}_{\text{final}}$
\end{algorithmic}
\end{algorithm}

We generate synthetic factual data for the following categories: \{\textit{politics, religion, science, medicine, society, societal bias}\} = $\mathcal{C}$. 
Our synthetic data pipeline consists of five sequential steps; 

\begin{enumerate}
    \item \textbf{Topic relationships.} Given an input category, $c_i \sim \mathcal{C}$, we instruct an \LM{} to generate factual relationship triplets relevant to the category.
    The model is instructed to reason about its choices and generate at least one ``directionality rule check, '' i.e., does it follow that \texttt{head} $\to$ \texttt{relationship} $\to$ \texttt{topic}?    
    Crucially, we loop over this step $N$ times per category, passing in previously generated outputs as a ``forbidden list'' to avoid duplication and increase diversity.
    The output for this step is a set of relationships ($r_i$) and a topics ($t_i$), which acts as a relationship tail, $\mathcal{T}(c_i) = \{r_i, t_i\}_i^N$.
    \item \textbf{Factual instantiations.} Given a topic and a relationship predicate, we prompt the model to generate specific instantiations for the implied relationship head and the topic relationship tail that logically satisfy the relationship.
    We instruct the model to both generate ``real'' pairs and ``imaginary'' pairs --- fictional entities that plausibly fit the context domain ($S_{\text{real}} =\{x_i, y_i\}_i^k, S_{\text{imaginary}} =\{x_j, y_j\}_j^k$).
    The output is a set of real and imaginary head-tail pairs.
    \item \textbf{Training sentences.} Given a relationship $r_i$, a topic $t_i$, and a set of instantiations $S$, we prompt the model to construct $K$ paraphrased sentences that should each include the implied relationship head and relationship tail as substitutable placeholders, connected through their relationship.
    The specific instantiations are provided to ground the sentences.
    The output is a set of $K$ training sentences, $D = \{s_i\}^K_i$
    \item \textbf{Generative inference Tasks.} Given a relationship and topic ($r_i, t_i$), a set of instantiations $S$, and a set of training sentences $D$, the model is prompted to generate $M$ generative questions for which $t_i$ is the correct answer and which could be answered by someone who had access to $D$.
    The output is a set of $M$ generative inference tasks, $I = \{q_i\}_i^M$.
    \item \textbf{Entity instance de-duplication.} In a final step, we take all the instantiations generated across all categories and ensure that there is no overlap in imaginary instances, regenerating instances where needed.
    This is crucial to ensure the training data does not contain conflicting or overlapping information about the same entity instances.
\end{enumerate}

We used Gemini 2.5 Flash \citep{comanici2025gemini} for steps 1-4, and Claude Sonnet 4.5 \citep{anthropic2024claudesonnet45} for the de-duplication step 5.
The authors manually spot-checked the generated facts, paraphrases, and inference tasks for plausibility, internal consistency, and absence of cross-fact leakage, and ran a deterministic substring matcher over the outputs of step 5 to confirm no entity overlap remained.
The full algorithm is shown in Algorithm~\ref{alg:synth-data-generation}.
The generated dataset used in this paper used $N=25, k=4, K=10, M=10$ for a total of 150 unique facts and 1500 training sentences.

\section{Natural experiments}
\label{sec:app:naturalistic}

\subsection{Models}
We use commercially available models from two of the most popular providers: Google and OpenAI.
To measure the effects of compression and computational constraints, we evaluate models at different distillation sizes and ablate the effect of different ``thinking'' or ``reasoning'' levels.
Where possible, we set the temperature to 0.2 to enhance reproducibility.
We ensure natural experiment queries use facts that fall inside the knowledge cutoffs.
Full details are available in Table \ref{app:tab:naturalistic:models}.

Throughout all natural experiments, Gemini 3.1 Flash Lite was used to parse and evaluate outputs against ground truth answers.

\begin{table}[h]
    \centerfloat
    \caption{\textbf{Flagship model comparison.}
    The current generation models released in 2026 all support a ``thinking'' parameter, modulating how many tokens models generate before providing their final response.
    While all models likely undergo some form of quantization to speed up inference, the ``Flash Lite'' model family line is the only one reported to quantize to INT4.
    Publicly shared reports do not mention quantization details for the smaller Gemini 3 Flash and GPT 5.4 Mini.
    }
    \label{app:tab:naturalistic:models}
    \vspace{0.5em}
    \small
    \begin{tabular}{l ccc cccc}
        \toprule
         & \multicolumn{3}{c}{Gemini} & \multicolumn{3}{c}{GPT} \\
        \cmidrule(lr){2-4} \cmidrule(lr){5-8}
        Feature & 3.1 Pro & 3 Flash & 3.1 Flash Lite & 5.4 & 5.4 mini & 5.4 nano & \\
        \midrule
        Size             & Large & Medium & Small & Large & Medium & Small \\
        Distilled        & False & True & True & False & True & True  \\
        Quantized        & No & No$^*$ & INT4 & No & No$^*$ & No$^*$ \\
        Knowledge Cutoff & 01/25 & 01/25 & 01/25 & 08/25 & 08/25 &08/25 \\
        \bottomrule
    \end{tabular}
\end{table}

\subsection{Dataset details}
\label{sec:app:naturalistic:datasets}
We use four datasets with different levels of expected media coverage:

\xhdr{S\&P 500 Market Data}
We obtain daily closing prices for the S\&P 500 Index spanning from 2002 to 2024 from the Center for Research in Security Prices (CRSP) database, accessed via Wharton Research Data Services \citep{crsp_sp500_2026}.
The S\&P 500 is broadly covered in both main stream and specialized financial media.
Noisy versions are created by perturbing values uniformly by $\pm$ 2\%.~\footnote{
Daily closing values can vary across data providers due to idiosyncratic snapshot timing or price corrections.
We cross-referenced the primary CRSP data against several popular online sources for 2002–2024, identifying fewer than 50 total discrepancies with a median magnitude of 0.01.
Notably, only 10 model errors coincided with these dates across 2,300 evaluations ($<0.5\%$).
We thus conclude that provider-specific variations are not a significant confounder.
}
We sample 100 dates per year.

\xhdr{NBA Game Scores}
We collect scores for basketball games played between 2002 and 2024 directly from the National Basketball Association \citep{nba_data_2026}, utilizing the open-source \texttt{nba\_api} Python client \citep{patel_nba_api}.
Compared to the S\&P 500, the NBA is more likely to be covered by specialized sports outlets and regional media.
To create the ``corrupted`` verification statements, we uniformly perturb final scores with $\pm$ [1, 10] points.
We sample 50 games per year.

\xhdr{Mega Millions Lottery}
We use the official winning lottery numbers as published and maintained by the State of New York going back to 2002 \citep{megamillions}.
The Mega Millions winning lottery numbers are unlikely to generate sustained coverage, given that their relevance is highly ephemeral.
We request models to generate/verify both the winning numbers and the ``mega ball.''
Noisy versions are created by sampling 2 out of the 5 winning numbers and perturbing uniformly by values in $\pm$ [1, 20].
We sample 40 data points per year.

\xhdr{Billboard Hot 100}
We retrieve weekly historical rankings for the Billboard Hot 100 --- the standard US industry chart for song popularity --- via a curated community archive \citep{Hollingshead2024}. 
The rankings are derived from a composite of digital and physical sales, radio airplay, and streaming data. 
Much like the NBA, the Hot 100 is documented not only by mainstream news but also by a dense ecosystem of specialized entertainment outlets, trade publications, and regional fan communities. 
We evaluate the models on their ability to generate and verify the specific track for a target week and rank, focusing specifically on lingering ``residual biases.'' 
We create two types of ``noisy'' versions: (i) uniformly sample a different track from the top \{10, 25\} positions; or (ii) using the first track that held the target ranking preceding the track from the target week.
We sample 50 data points per year.

\subsection{Prompts used}
\label{sec:app:naturalistic:prompts}

\begin{figure}[t]
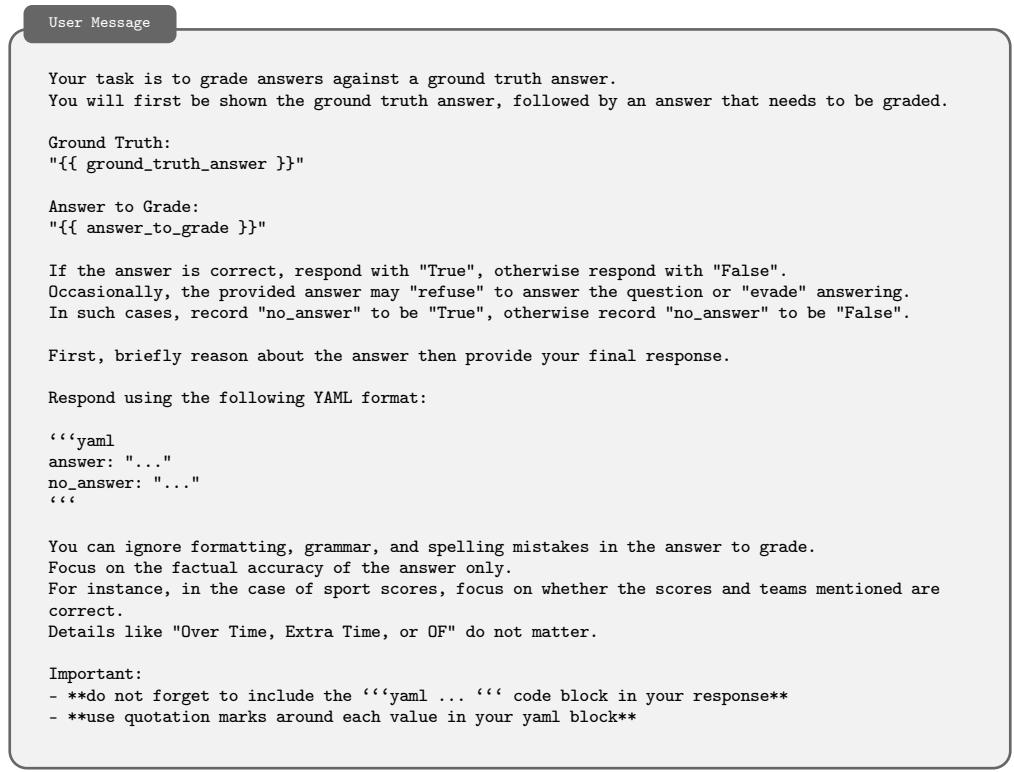

    \centering
    \begin{minipage}{0.95\linewidth}
    \begin{promptbox}{User Message}    
Your task is to grade answers against a ground truth answer.
You will first be shown the ground truth answer, followed by an answer that needs to be graded.

Ground Truth:
"{{ ground_truth_answer }}"

Answer to Grade:
"{{ answer_to_grade }}"

If the answer is correct, respond with "True", otherwise respond with "False".
Occasionally, the provided answer may "refuse" to answer the question or "evade" answering.
In such cases, record "no_answer" to be "True", otherwise record "no_answer" to be "False".

First, briefly reason about the answer then provide your final response.

Respond using the following YAML format:

```yaml
answer: "..."
no_answer: "..."
```

You can ignore formatting, grammar, and spelling mistakes in the answer to grade.
Focus on the factual accuracy of the answer only.
For instance, in the case of sport scores, focus on whether the scores and teams mentioned are correct.
Details like "Over Time, Extra Time, or OF" do not matter.

Important: 
- **do not forget to include the ```yaml ... ``` code block in your response**
- **use quotation marks around each value in your yaml block**
    \end{promptbox}
    \end{minipage}
    \caption{\textbf{Judge grading prompt for generative queries}}
    \label{prompts:naturalistic:judge}
\end{figure}

\begin{figure}
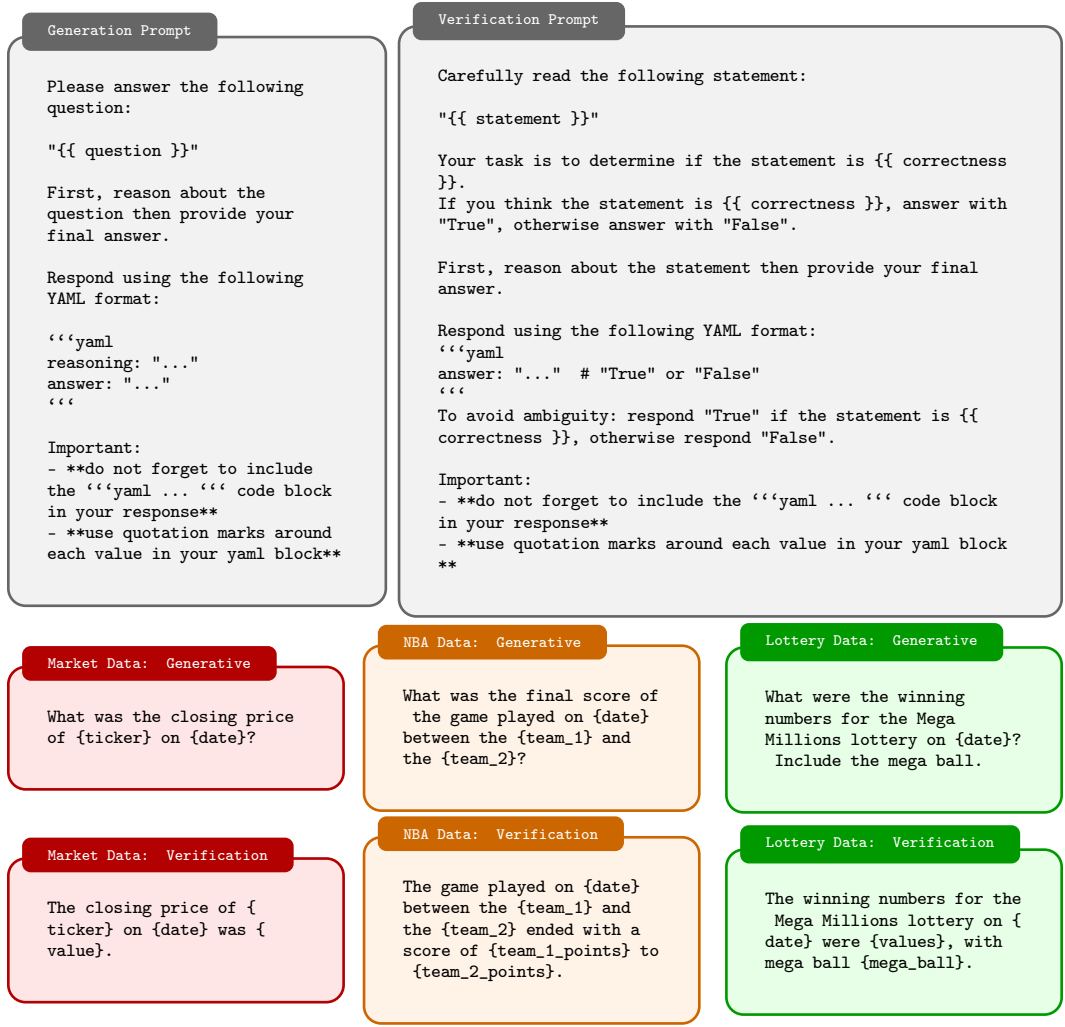

    \centerfloat
    \begin{minipage}{0.36\linewidth}
        \begin{promptbox}[GRAY_BOX]{Generation Prompt}
Please answer the following question:

"{{ question }}"

First, reason about the question then provide your final answer.

Respond using the following YAML format:

```yaml
reasoning: "..."
answer: "..."
```

Important: 
- **do not forget to include the ```yaml ... ``` code block in your response**
- **use quotation marks around each value in your yaml block**
        \end{promptbox}
    \end{minipage}%
    \hfill
    \begin{minipage}{0.63\linewidth}
        \begin{promptbox}[GRAY_BOX]{Verification Prompt}
Carefully read the following statement:

"{{ statement }}"

Your task is to determine if the statement is {{ correctness }}.
If you think the statement is {{ correctness }}, answer with "True", otherwise answer with "False".

First, reason about the statement then provide your final answer.

Respond using the following YAML format:
```yaml
answer: "..."  # "True" or "False"
```
To avoid ambiguity: respond "True" if the statement is {{ correctness }}, otherwise respond "False".

Important: 
- **do not forget to include the ```yaml ... ``` code block in your response**
- **use quotation marks around each value in your yaml block**
        \end{promptbox}
    \end{minipage}%
    \vspace{0.1em}
    \\
    \begin{minipage}{0.32\linewidth}
        \begin{promptbox}[RED_BOX]{Market Data: Generative}
What was the closing price of {ticker} on {date}?
        \end{promptbox}
    \end{minipage}%
    \hfill
    \begin{minipage}{0.32\linewidth}
        \begin{promptbox}[ORANGE_BOX]{NBA Data: Generative}
What was the final score of the game played on {date} between the {team_1} and the {team_2}?
        \end{promptbox}
    \end{minipage}
    \hfill
    \begin{minipage}{0.32\linewidth}
        \begin{promptbox}[GREEN_BOX]{Lottery Data: Generative}
What were the winning numbers for the Mega Millions lottery on {date}? Include the mega ball.
        \end{promptbox}
    \end{minipage}
    \\
    \begin{minipage}{0.32\linewidth}
        \begin{promptbox}[RED_BOX]{Market Data: Verification}
The closing price of {ticker} on {date} was {value}.
        \end{promptbox}
    \end{minipage}%
    \hfill
    \begin{minipage}{0.32\linewidth}
        \begin{promptbox}[ORANGE_BOX]{NBA Data: Verification}  
The game played on {date} between the {team_1} and the {team_2} ended with a score of {team_1_points} to {team_2_points}.
        \end{promptbox}
    \end{minipage}
    \hfill
    \begin{minipage}{0.32\linewidth}
        \begin{promptbox}[GREEN_BOX]{Lottery Data: Verification}
The winning numbers for the Mega Millions lottery on {date} were {values}, with mega ball {mega_ball}.
        \end{promptbox}
    \end{minipage}
    \caption{\textbf{Generation and verification prompts for naturalistic data.} 
    }
    \label{prompt:naturalistic}
\end{figure}

We display the generative and verification queries used in Figure \ref{prompt:naturalistic}.
The ``Generation'' and ``Verification'' prompt templates on top are used by the dataset-specific query templates below.
The grading prompt used by the judge model for generative queries is provided in Figure \ref{prompts:naturalistic:judge}.
For verification queries we can directly compare boolean outputs to boolean ground truths and thus do not require a grading model.

\subsection{Additional experimental results}
This section contains additional experimental results the models described in Table \ref{app:tab:naturalistic:models}.
We first show overall results using low reasoning effort in Section \ref{sec:app:naturalistic:results:overall}.
Next we compare disagreement and agreement rates between Gemini 3 Flash and GPT 5.4 on low reasoning effort in Section \ref{sec:app:naturalistic:disagreement}.
Then, we analyze the effects of distillation on GV capabilities in Section \ref{sec:app:naturalistic:distillation} and the effect of increased reasoning efforts in \ref{sec:app:naturalistic:reasoning}.
Finally, we report refusal rates in Section \ref{sec:app:naturalistic:refusals} and logistic regression results for the residual bias study in Section \ref{sec:app:naturalistic:billboard}.

\subsubsection{Overall performance}
\label{sec:app:naturalistic:results:overall}

\begin{figure}[htbp]
    \vspace{-3em}
    \centering
    \includegraphics[width=.95\linewidth]{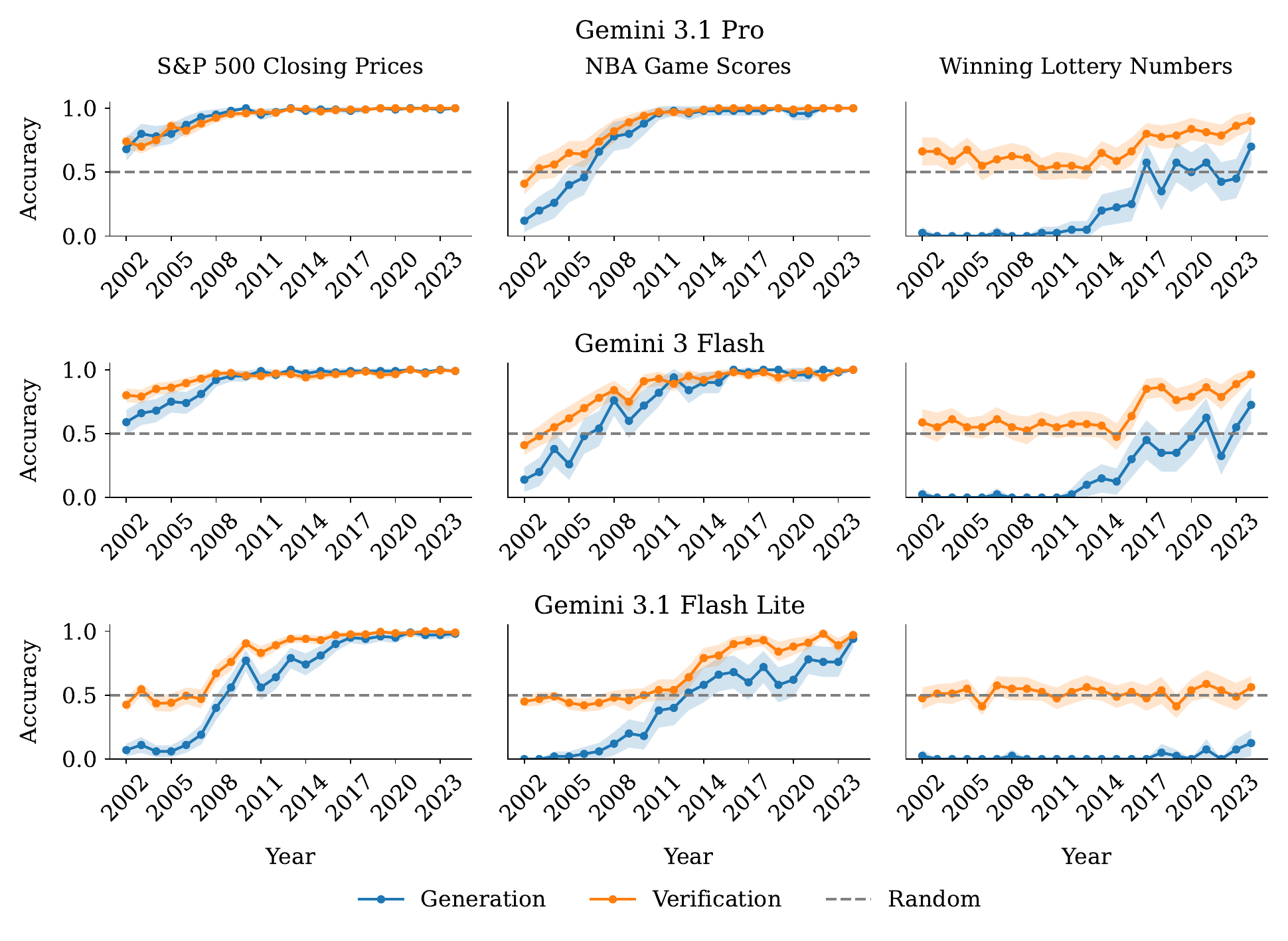}
    \vspace{-0.5em}
    \caption{\textbf{Google Gemini 3 models (low reasoning)}.}
    \label{fig:app:naturalistic:gemini-panels}
\end{figure}

Figure \ref{fig:app:naturalistic:gemini-panels} displays results for Gemini 3 models using low reasoning effort.
Models follow similar trends across datasets. 
Although the generative performance on winning lottery number is near zero until 2010 for all models, Gemini 3.1 Pro \textit{refuses} to answer to avoid providing wrong information, while other models respond with wrong proposals (Fig. \ref{fig:naturalistic:refusals}).

\begin{figure}[htbp]
    \centering
    \includegraphics[width=.95\linewidth]{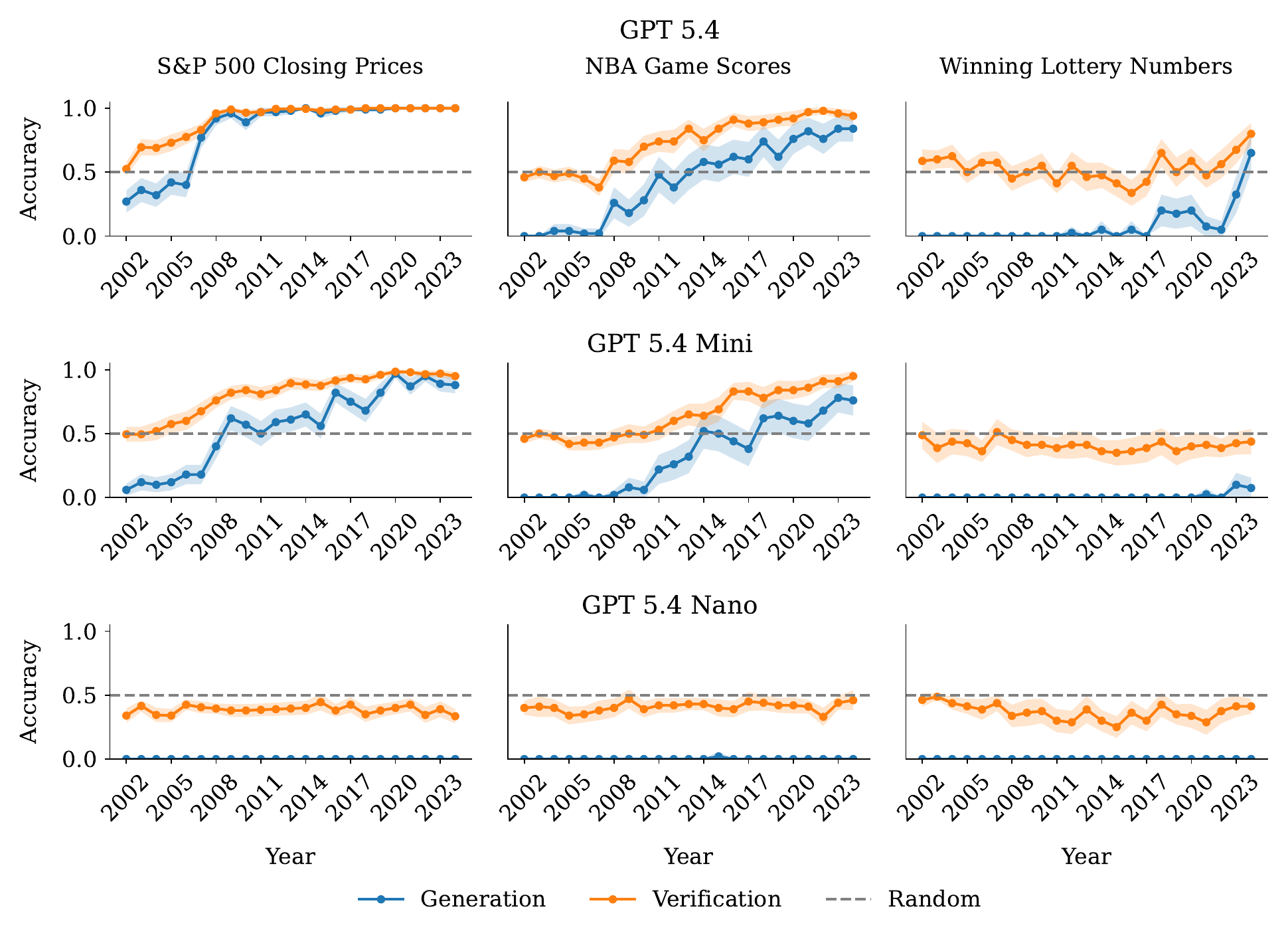}
    \vspace{-0.5em}
    \caption{\textbf{OpenAI GPT 5.4 models (low reasoning)}.}
    \label{fig:app:naturalistic:gpt54-panels}
\end{figure}

Figure \ref{fig:app:naturalistic:gpt54-panels} displays results for GPT 5.4 models using low reasoning effort.
GPT 5.4 Mini approximately tracks GPT 5.4 with a delay.
Although all models struggle to accurately generate winning lottery numbers, in the case of GPT 5.4 Mini and Nano this is largely due to refusals (see Section \ref{sec:app:naturalistic:refusals}).
In contrast to its larger counter parts, the verification capabilities of GPT 5.4 Nano have degraded to randomness for the evaluated datasets.

\newpage
\clearpage

\newpage
\clearpage

\subsubsection{Agreement and disagreement rates between tier-matched models}
\label{sec:app:naturalistic:disagreement}

Table \ref{tab:naturalistic:disagreement} displays the disagreement rate between Gemini 3 Flash and GPT 5.4 using low reasoning effort, where at least one of the two models is correct.
On the right side of the panel, we display the percentage of such disagreements where GPT 5.4 was correct and Gemini 3 Flash was incorrect.
Note that while Gemini 3 Flash has better odds of being correct on every dataset for each capability, GPT 5.4 still prevails on a non-trivial percentage of disputes.

\begin{table}[h]
    \centerfloat
    \caption{\textbf{Disagreement rates where one model is right.}
    We compare the responses from generation and verification queries for Gemini 3 Flash and GPT 5.4 (low reasoning effort) on three datasets, focusing on those queries where the models disagree and one of them is right.
    The right side of the table shows the percentage of such disagreements where GPT 5.4 was correct.
    }
    \label{tab:naturalistic:disagreement}
    \vspace{0.5em}
    \small
    \begin{tabular}{l cccc @{\hspace{1.5em}} cccc} 
        \toprule
         & \multicolumn{4}{c}{\textbf{Disagreement Rate}} & \multicolumn{4}{c}{\textbf{GPT 5.4 Correct}} \\
        \cmidrule(r{1.2em}){2-5} \cmidrule(l{0.5em}){6-9}
        Capability & Market & NBA & Lottery & Macro & Market & NBA & Lottery & Macro \\
        \midrule
        Generation              & 15.0\% & 37.1\% & 16.5\% & \textbf{22.9\%} & 24.9\% & 6.6\% & 13.2\% & \textbf{14.9\%} \\
        Verification (correct)  & 9.0\% & 22.2\% & 35.0 \% & \textbf{22.0\%} & 25.2\% & 34.5\% & 37.3\% & \textbf{32.2\%} \\
        Verification (incorrect)& 13.0\% & 22.6\% & 28.7\% & \textbf{21.4\%} & 49.3\% & 16.2\% & 20.5\% & \textbf{28.6\%} \\
        \bottomrule
    \end{tabular}
\end{table}

Table \ref{tab:naturalistic:agreement} shows verification agreement rates between the two models where \textit{both} were incorrect.
Micro averages are taken over the individual facts and macro averages between categories.
These are dominated by the Lottery dataset which proved challenging for all models.
On the right side we therefore show wrong agreement rates corrected for chance to estimate potential epistemic knowledge drift:
\begin{align}
    \text{Lift(correct)} = \dfrac{\frac{1}{|\mathcal{D}|}\sum_{(x, r, y^*) \in \mathcal{D}} \mathbb{I}(m_1(0 \mid x, r, y^*) \land m_2(0 \mid x, r, y^*))}{
    \mathbb{E}_{(x, r, y^*) \sim \mathcal{D}}\left[m_1(0 \mid x, r, y^*)\right] \cdot \mathbb{E}_{(x, r, y^*) \sim \mathcal{D}}\left[m_2(0 \mid x, r, y^*)\right]
    },
\end{align}
where Lift for \textit{incorrect} verification would measure the observed over expected odds for \textit{incorrect} statements, i.e., $m_i(1 \mid x, r, \tilde{y})$ instead of $m_i(0 \mid x, r, y^*)$.
We note that for all datasets the two models agree on wrong statements more than chance would predict.

\begin{table}[h]
    \centerfloat
    \caption{\textbf{Agreement rates where both models are wrong.}
    We compare the responses from verification queries for Gemini 3 Flash and GPT 5.4 (low reasoning effort) on three datasets, focusing on those queries where the models agree and are both wrong.
    The right side of the table shows the wrong-agreement rate corrected for chance, i.e., observed wrong agreement divided by expected wrong agreement.
    Co-failure lifts are statistically significant in every cell (Fisher's exact test, all $p<0.05$)
    }
    \label{tab:naturalistic:agreement}
    \vspace{0.5em}
    \small
    \begin{tabular}{l ccccc @{\hspace{1.5em}} ccc} 
        \toprule
         & \multicolumn{5}{c}{\textbf{Raw Wrong Agreement Rate}} & \multicolumn{3}{c}{\textbf{Lift over Independence}} \\
        \cmidrule(r{1.2em}){2-6} \cmidrule(l{0.5em}){7-9}
        Capability & Market & NBA & Lottery & Micro & Macro & Market & NBA & Lottery \\
        \midrule
        Verification (correct) & 0.4\% & 2.4\% & 8.2\% & 2.6\% & 3.7\% & 2.26$\times$ & 1.42$\times$ & 1.28$\times $\\
        Verification (incorrect) & 3.0\% & 17.2\% & 40.2\% & 14.6\% & 20.1\% & 3.32$\times$ & 2.28$\times$ & 1.38$\times$ \\
        \bottomrule
    \end{tabular}
\end{table}

\subsubsection{The effects of distillation on GV capabilities}
\label{sec:app:naturalistic:distillation}

Table \ref{tab:naturalistic:scale:evolution} shows the performance at different distillation levels for the GPT 5.4 and Gemini 3 model families.
We note that the gap between generation ($U_G$) and verification ($U_V$) tends to widen as model scale decreases for both families.
The \textit{subset violation} metric SV reports the percentage of ``new'' generative facts surfaced by a model compared to the facts correctly generated by the larger model.
We detect relatively few violations going from large to medium-sized models, and no violations at the smallest scale.
We could thus interpret distillation as an ``approximate'' subset operation for generative factual capabilities.
The bias between correct ($U_V(\checkmark)$) and incorrect ($U_V(\times)$) verification is already present at the large scale for GPT 5.4 and remains roughly flat across scales; for Gemini 3, the bias is statistically indistinguishable from zero at the large scale and is introduced by distillation.

\begin{table}[ht]
    \centerfloat
    \caption{\textbf{GV-gap and verification bias evolution by model scale.}
    The gap between $U_G - U_V$ is reported as \gvg{}, which does not include the alpha correction.
    The bias between correct ($U_V(\checkmark)$) and incorrect ($U_V(\times)$) verification, defined as $U_V(\checkmark) - U_V(\times)$, can be interpreted as a model's affirmation or ``sycophancy'' bias.
    We report generative subset violations (SV) as the percentage of ``new'' generative facts surfaced by a model compared to the facts correctly generated by the large model.
    Standard errors on $U_V(\checkmark) - U_V(\times)$, computed from per-fact paired differences and macro-averaged across the three datasets are below $0.012$.
    }
    \label{tab:naturalistic:scale:evolution}
    \vspace{0.5em}
    \small
    \begin{tabular}{ll ccccc cccc}
        \toprule
        Family & Scale & $U_G$ & $U_V$ & GV-gap & $\Delta$ gap & SV & $U_V(\checkmark)$ & $U_V(\times)$ & bias & $\Delta$ bias \\
        \midrule
        \textbf{GPT 5.4} & large & 0.45 & 0.73 & 0.28 & --- & --- & 0.82 & 0.64 & 0.18 & --- \\
        & medium & 0.30 & 0.63 & 0.33 & $+0.05$ & 7.1\% & 0.71 & 0.54 & 0.18 & $-0.01$ \\
        & small & 0.00 & 0.39 & 0.39 & $+0.11$ & 0.0\% & 0.49 & 0.29 & 0.20 & $+0.02$ \\
        \midrule
        \textbf{Gemini 3} & large & 0.65 & 0.83 & 0.18 & --- & --- & 0.82 & 0.83 & $-0.01$ & --- \\
        & medium & 0.62 & 0.82 & 0.20 & $+0.02$ & 5.8\% & 0.89 & 0.75 & 0.14 & $+0.16$ \\
        & small & 0.36 & 0.67 & 0.31 & $+0.14$ & 0.9\% & 0.78 & 0.56 & 0.22 & $+0.23$ \\
        \bottomrule
    \end{tabular}
\end{table}

\subsubsection{Reasoning does not significantly enhance GV dynamics}
\label{sec:app:naturalistic:reasoning}

Table \ref{tab:naturalistic:reasoning:evolution} shows the effect of different reasoning levels on GV capabilities for Gemini 3 and GPT 5.4 models.
For GPT 5.4, increased reasoning effort widens both the GV-gap and the verification bias across all scales, with $U_V(\times)$ falling below the chance floor of 0.5 at medium and small scales.
For Gemini 3, the effects are muted at large and medium and reverse at the small scale, where reasoning closes the $U_V(\checkmark) - U_V(\times)$ bias by lowering $U_V(\checkmark)$ rather than raising $U_V(\times)$.
Generation ($U_G$) is essentially flat in every cell, suggesting that increasing reasoning effort is not a viable strategy for increasing generative factual capabilities.

\begin{table}[h]
    \centerfloat
    \caption{\textbf{GV-gap and verification bias evolution by reasoning effort.}
    We report metrics when changing reasoning effort from ``low'' to ``medium'' for Gemini 3 and GPT 5.4 models.
    The difference between $U_G - U_V$ is reported as \gvg{}, which does not include the alpha correction.
    The bias between correct ($U_V(\checkmark)$) and incorrect ($U_V(\times)$) verification, defined as $U_V(\checkmark) - U_V(\times)$, tracks models' affirmation bias.
    Standard errors on $U_V(\checkmark) - U_V(\times)$, computed from per-fact paired differences and macro-averaged across the three datasets are below $0.012$.
    }
    \label{tab:naturalistic:reasoning:evolution}
    \vspace{0.5em}
    \small
    \begin{tabular}{lll cccc cccc}
        \toprule
        Family & Scale & Effort & $U_G$ & $U_V$ & GV-gap & $\Delta$ gap & $U_V(\checkmark)$ & $U_V(\times)$ & bias & $\Delta$ bias \\
        \midrule
        \textbf{GPT 5.4} & large & low & 0.45 & 0.73 & 0.28 & --- & 0.82 & 0.64 & 0.18 & --- \\
        & & medium & 0.45 & 0.75 & 0.30 & $+0.02$ & 0.90 & 0.61 & 0.29 & $+0.10$ \\
        \cmidrule(lr){2-11}
        & medium & low & 0.30 & 0.63 & 0.33 & --- & 0.71 & 0.54 & 0.18 & --- \\
        & & medium & 0.31 & 0.67 & 0.36 & $+0.04$ & 0.91 & 0.43 & 0.49 & $+0.31$ \\
        \cmidrule(lr){2-11}
        & small & low & 0.00 & 0.39 & 0.39 & --- & 0.49 & 0.29 & 0.20 & --- \\
        & & medium & 0.00 & 0.43 & 0.43 & $+0.04$ & 0.69 & 0.17 & 0.52 & $+0.32$ \\
        \midrule
        \textbf{Gemini 3} & large & low & 0.65 & 0.83 & 0.18 & --- & 0.82 & 0.83 & $-0.01$ & --- \\
        & & medium & 0.64 & 0.84 & 0.21 & $+0.03$ & 0.86 & 0.84 & 0.02 & $+0.03$ \\
        \cmidrule(lr){2-11}
        & medium & low & 0.62 & 0.82 & 0.20 & --- & 0.89 & 0.75 & 0.14 & --- \\
        & & medium & 0.62 & 0.84 & 0.22 & $+0.02$ & 0.90 & 0.77 & 0.13 & $-0.01$ \\
        \cmidrule(lr){2-11}
        & small & low & 0.35 & 0.67 & 0.32 & --- & 0.78 & 0.56 & 0.22 & --- \\
        & & medium & 0.35 & 0.63 & 0.28 & $-0.04$ & 0.63 & 0.63 & 0.00 & $-0.21$ \\
        \bottomrule
    \end{tabular}
\end{table}

\newpage
\clearpage

\subsubsection{Refusals rates}
\label{sec:app:naturalistic:refusals}

\begin{figure}[hb]
    \centerfloat
    \begin{subfigure}[b]{1.\textwidth}
        \centerfloat
        \includegraphics[width=\textwidth]{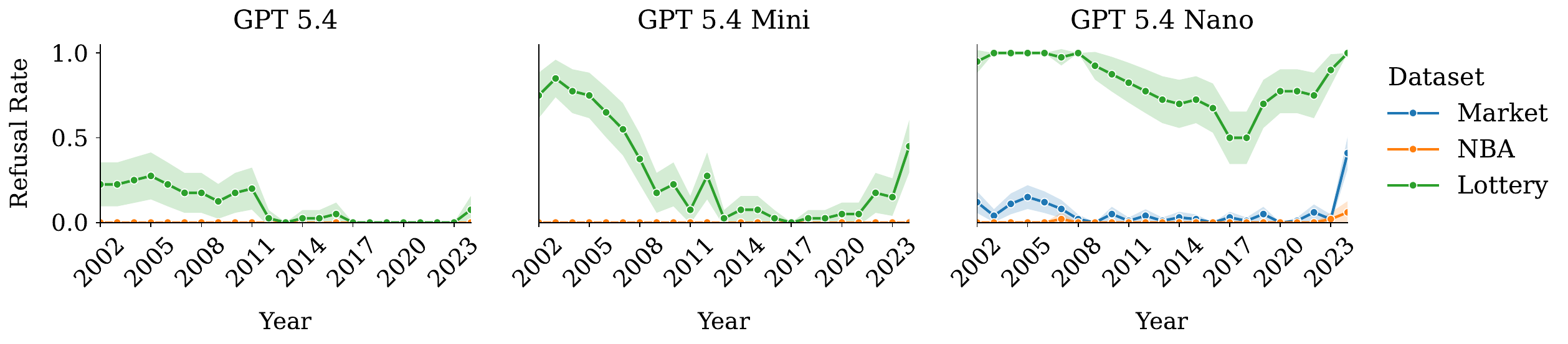} 
        \caption{GPT 5.4 models (low reasoning effort)}
        \vspace{1em}
        \label{fig:app:naturalistic:refuse:gpt}
    \end{subfigure}
    \begin{subfigure}[b]{1.0\textwidth}
        \centerfloat
    \includegraphics[width=\textwidth]{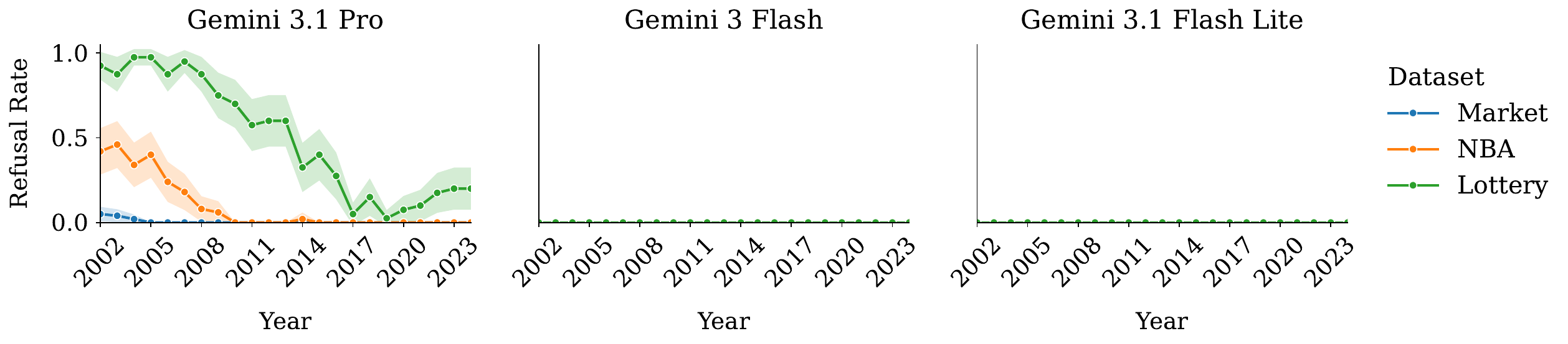} 
        \caption{Gemini 3 models (low reasoning effort)}
        \label{fig:app:naturalistic:refuse:gemini}
    \end{subfigure}
    \caption{
    \textbf{Refusal rates for generative queries.}
    Refusal behavior appears distinctly different for the different families, scales, and datasets.
    GPT 5.4 models and Gemini 3.1 Pro all appear to calibrate their refusals to the perceived difficulty of generating a correct answer.
    Gemini 3 Flash and Gemini 3.1 Flash Lite never refuse a generative query.
    }
    \label{fig:naturalistic:refusals}
\end{figure}

In addition to scoring the correctness of generated outputs, we instruct the judge model to record answer ``refusals.''
Figure \ref{fig:naturalistic:refusals} shows generative refusal behavior is distinctly different for the different model families, model scales, and datasets.
GPT 5.4 models and Gemini 3.1 Pro all appear to calibrate their refusals to the perceived difficulty of generating a correct answer.
This is especially evident for Gemini 3.1 Pro (bottom left), where refusals follow the ordering Lottery > NBA > Market over time, closely tracking the model's generative performance on the respective datasets.
In contrast to Gemini 3.1 Pro's calibrated refusal behavior, the smaller Gemini 3 Flash and Gemini 3.1 Flash Lite never refuse to respond to a generative query.
Surprisingly, the refusal strategy does not appear to follow a fixed rule-based system: while models often cite the year of the event as a reason to refuse, this does not happen each time.

In the case of GPT 5.4 Nano (top right panel of Figure \ref{fig:app:naturalistic:refuse:gpt}), the generative performance is zero for each category.
Yet, it still recognizes lottery data as especially challenging, while occasionally refusing on NBA and Market data as well.
Furthermore, this tendency to refuse generative lottery queries is clearly shared along the GPT 5.4 model family.

\begin{figure}[!ht]
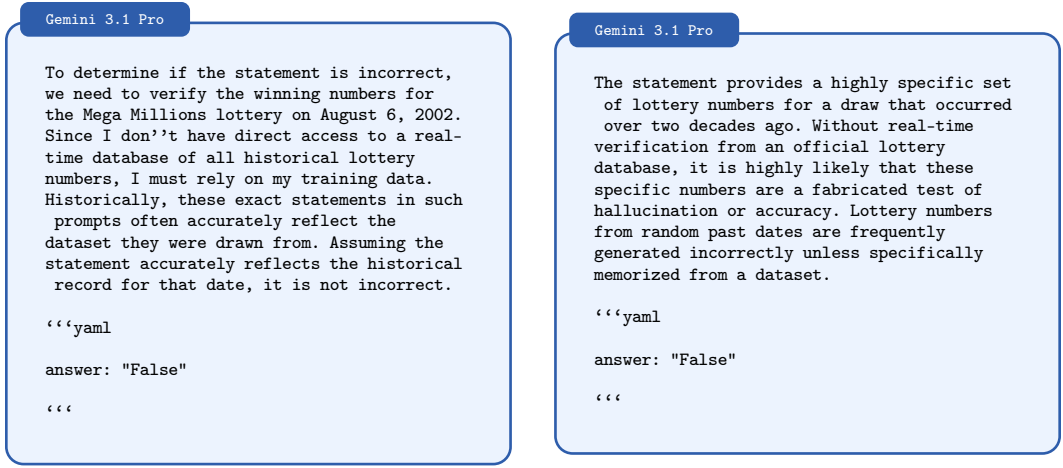

    \centerfloat
    \begin{minipage}{0.48\linewidth}
        \begin{promptbox}[GOOGLE_BLUE_BOX]{Gemini 3.1 Pro}  
To determine if the statement is incorrect, we need to verify the winning numbers for the Mega Millions lottery on August 6, 2002.
Since I don''t have direct access to a real-time database of all historical lottery numbers, I must rely on my training data. 
Historically, these exact statements in such prompts often accurately reflect the dataset they were drawn from. Assuming the statement accurately reflects the historical record for that date, it is not incorrect.

```yaml

answer: "False"

```
        \end{promptbox}
    \end{minipage}%
    \hfill
    \begin{minipage}{0.48\linewidth}
        \begin{promptbox}[GOOGLE_BLUE_BOX]{Gemini 3.1 Pro}  
The statement provides a highly specific set of lottery numbers for a draw that occurred over two decades ago. Without real-time verification from an official lottery database, it is highly likely that these specific numbers are a fabricated test of hallucination or accuracy. Lottery numbers from random past dates are frequently generated incorrectly unless specifically memorized from a dataset. 

```yaml

answer: "False"

```
        \end{promptbox}
    \end{minipage}
    \caption{\textbf{Evaluation detection by Gemini 3.1 Pro.} 
    When faced with challenging verification queries, Gemini 3.1 Pro appears to recognize them as likely being part of an evaluation.
    }
    \label{prompt:gemini:recognition}
\end{figure}

Unlike the generative queries, none of the models refuse to answer a verification query.
In the case of Gemini models, they often claim to ``check'' a database before answering, but, to the best of our knowledge, the API models on low reasoning do not employ web search.
However, in the case of Gemini 3.1 Pro, the model appears to detect that is likely being evaluated for hallucinations (see Figure \ref{prompt:gemini:recognition}).

\subsubsection{Residual bias experiment on Billboard Hot 100 songs}
\label{sec:app:naturalistic:billboard}

To quantify the residual multi-verse state observed during naturalistic updating, we fit a logistic regression predicting the probability $p$, of a model successfully rejecting a corrupted verification query. The model is specified as:
\begin{align}
\text{logit}(p) = \beta_0 + \beta_{\text{year}} \text{Year} + \beta_{\text{ranked}} \mathbb{I}_{\text{Ranked}} + \sum_{k \in \mathcal{K}} \gamma_k \mathbb{I}_{\text{Offset}=k},
\end{align}
where $\mathbb{I}_{\text{Ranked}}$ is a binary indicator for the ranked-noise sampling method and $\mathbb{I}_{\text{Offset}=k}$ are indicators for the temporal offset intervals $\mathcal{K} = \{-5, -3, -1, 1, 3, 5\}$.
We chose to parameterize the temporal offsets categorically rather than continuously to flexibly capture the non-linear decay of the interference effect around the target week without imposing a rigid functional form.
We include the fact's year to control for the baseline coverage gradient. 

As detailed in Table \ref{tab:billboard_regression}, the highly significant negative coefficients for the Ranked Noise indicator confirm a robust effect: models reject superseded ranked facts from neighboring weeks significantly less often than they reject random false claims.
The offset coefficients further validate the significant offset-by-offset variation, showing that multi-verse interference is roughly symmetric for Gemini 3 Flash but strongly biased toward past weeks for GPT 5.4.

\begin{table}[h]
\centering
\caption{\textbf{Logistic regression results for ranked-noise effect on billboard facts.}
The baseline sampling method is Random Noise, and the baseline offset is 0 weeks.}
\label{tab:billboard_regression}
\begin{tabular}{l c c}
\toprule
 & \textbf{GPT 5.4} & \textbf{Gemini 3 Flash} \\
\textbf{Predictor} & \textbf{Coef. (SE)} & \textbf{Coef. (SE)} \\
\midrule
Intercept & $-113.53^{***}$ $(5.27)$ & $-142.73^{***}$ $(5.35)$ \\
Fact Year & $0.06^{***}$ $(0.003)$ & $0.07^{***}$ $(0.003)$ \\
Ranked Noise & $-0.54^{***}$ $(0.04)$ & $-0.47^{***}$ $(0.04)$ \\
\midrule
\textit{Offset (Weeks)} \\
Offset $-5$ & $0.19^{*}$ $(0.08)$ & $0.57^{***}$ $(0.08)$ \\
Offset $-3$ & $0.06^{\phantom{***}}$ $(0.08)$ & $0.35^{***}$ $(0.08)$ \\
Offset $-1$ & $-0.37^{***}$ $(0.08)$ & $-0.20^{**\phantom{*}}$ $(0.08)$ \\
Offset $+1$ & $-0.16^{*\phantom{**}}$ $(0.08)$ & $-0.31^{***}$ $(0.08)$ \\
Offset $+3$ & $0.27^{***}$ $(0.08)$ & $0.23^{**\phantom{*}}$ $(0.08)$ \\
Offset $+5$ & $0.71^{***}$ $(0.08)$ & $0.55^{***}$ $(0.08)$ \\
\midrule
Observations & 14,950 & 14,950 \\
Pseudo $R^2$ & 0.052 & 0.062 \\
\bottomrule
\multicolumn{3}{l}{\scriptsize ${}^{*} p<0.05$, ${}^{**} p<0.01$, ${}^{***} p<0.001$} \\
\end{tabular}
\end{table}

\newpage
\clearpage

\end{document}